\documentclass[twoside,12pt]{Thesis_Class_File}

\usepackage{graphicx}
\usepackage{longtable}
\usepackage{lipsum}
\usepackage{tabularx}
\usepackage{multirow}
\usepackage{cite}
\usepackage{booktabs}
\usepackage{flushend}
\usepackage{subfig}
\usepackage{amsmath}
\usepackage{mathtools}
\usepackage{setspace}
\usepackage{color,soul}
\usepackage{comment}
\usepackage{lscape}
\usepackage{array}
\usepackage{sectsty}
\usepackage[compact]{titlesec}  
\chapterfont{\centering}
\usepackage[inner=2.54cm,outer=2.54cm,bottom=2.54cm,top=2.54cm]{geometry}
\usepackage{layout}
\usepackage{bbding}
\usepackage [autostyle, english = american]{csquotes}
\MakeOuterQuote{"}
\usepackage{commath}
\RequirePackage{amsfonts,amstext,amssymb,amsmath,amsthm,amscd,bm,paralist,color}
\theoremstyle{plain}

\newtheorem{assumption}{Assumption}[chapter]

\theoremstyle{remark}

\usepackage{soul}
\usepackage{frcursive}
\usepackage{inslrmin}
\usepackage{pdfpages}
\usepackage{algorithm}
\usepackage{algpseudocode}
\usepackage{xpatch}
\makeatletter
\xpatchcmd{\algorithmic}{\itemsep\z@}{\itemsep=2ex plus2pt}{}{}
\makeatother
\usepackage{pifont}
\usepackage{tabu}
\usepackage{booktabs}
\usepackage[numbered]{mcode}
\usepackage{siunitx}

\usepackage[toc,page]{appendix}
\usepackage{afterpage}

\newcommand{\forceindent}{\leavevmode{\parindent=2em\indent}}

\ifpdf
    \pdfcatalog { /PageMode (/UseOutlines)
                  /OpenAction (fitbh)  }
\fi

\title{\textbf{\LARGE Controller Design and Implementation of a New Quadrotor Manipulation System}}

\ifpdf
  \author{Ahmed Mohammed Elsayed Khalifa}
  \collegeordept{{Graduate School of Innovative Design Engineering,}}
  \university{Egypt-Japan University of Science and Technology \textbf{(E-JUST)}}
  \crest{\includegraphics[{scale=0.23}]{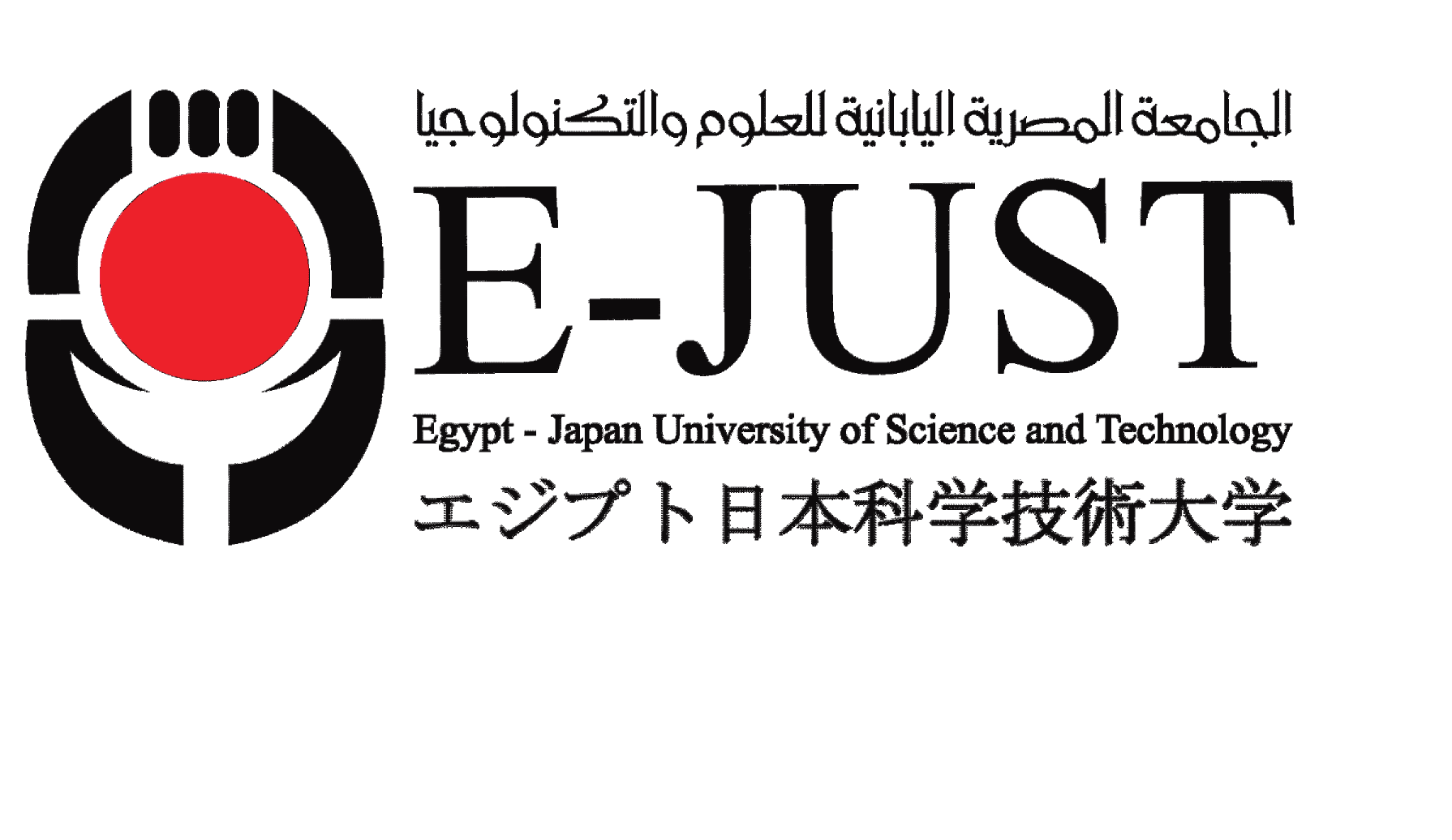}} 

\fi

\degree{In Partial Fulfillment of the Requirements for the Degree of \\\vspace{3mm}\Large{Doctor of Philosophy}\\\vspace{10mm}\normalsize{in}\\\vspace{3mm}\Large{Mechatronics and Robotics Engineering\\}}
\degreedate{September 2016}

\begin{document}

\renewcommand\baselinestretch{1.5}
\baselineskip=18pt plus1pt


\maketitle 
\renewcommand\baselinestretch{1.2}
\baselineskip=18pt plus1pt

\newpage\null\thispagestyle{empty}\newpage
\thispagestyle{empty}

\begin{center}
{\setlength{\baselineskip}{1.5\baselineskip}
\textbf{\LARGE Controller Design and Implementation of a New Quadrotor Manipulation System}
\par}
\vspace{3mm}
By\\
\large{Ahmed Mohammed Elsayed Khalifa}\\ 
\vspace{3mm}
\normalsize{For The Degree}\\
\normalsize{of}\\
\large{Doctor of Philosophy}\\
\vspace{3mm}
\normalsize{in}\\
\large{Mechatronics and Robotics Engineering}\\
\end{center}

\vspace*{5mm}
\begin{center}
	\textbf{\underline{Supervision Committee}}
\end{center}
\begin{tabular}{p{6cm}  p{6cm} p{5cm}}
\textbf{Name, Title} & \textbf{Affiliation} & \textbf{Signature, Date} \\
\\
Prof. Mohamed A. Fanni & Mechatronics and Robotics \newline Engineering, E-JUST &\ldots\ldots\ldots\ldots \\
\\
Prof. Toru Namerikawa & Systems and Control \newline Engineering, Keio University &\ldots\ldots\ldots\ldots \\

\end{tabular}
\vspace*{5mm}
\begin{center}
	\textbf{\underline{Examination Committee}}
\end{center}
\vspace*{5mm}
\begin{tabular}{p{6cm}  p{6cm} p{5cm}}
\textbf{Name, Title} & \textbf{Affiliation} & \textbf{Approved, Date} \\
\\
Prof. Sohair F. Rezeka & Mechanical Engineering, \newline Alexandria University & \ldots\ldots\ldots\ldots \\
\\
Prof. Khaled A. El-Metwally & Automatic Control \newline Engineering, Cairo University & \ldots\ldots\ldots\ldots \\
\\
Prof. Mohamed A. Fanni & Mechatronics and Robotics \newline Engineering, E-JUST & \ldots\ldots\ldots\ldots \\

\end{tabular}

\vspace*{5mm}

\newpage\null\thispagestyle{empty}\newpage

\begin{copyrightKhalifa}  
\begin{flushleft}	
\textbf{Controller Design and Implementation of a New Quadrotor Manipulation System}
  
by Ahmed Khalifa

Copyright \textcopyright \ 2016 Ahmed Mohammed Elsayed Khalifa. All rights reserved.
This document was created with the document preparation system \LaTeX.
\end{flushleft}
\end{copyrightKhalifa}
 
\newpage\null\thispagestyle{empty}\newpage
\phantomsection
\addcontentsline{toc}{chapter}{DECLARATION AND CERTIFICATE OF ORIGINALITY}



\begin{declaration}  
	
I certify that in the preparation of this thesis, I have observed the provisions of E-JUST Code of Ethics dated 8 September 2013. Further; I certify that this work is free of plagiarism and all materials appearing in this thesis have been properly quoted and attributed. 
I certify that all copyrighted material incorporated into this thesis is in compliance with the international copyright law and that I have received written permission from the copyright owners for my use of their work, which is beyond the scope of the law. 
I agree to indemnify and save harmless E-JUST from any and all claims that may be asserted or that may arise from any copyright violation.
I hereby certify that the research work in this thesis is my original work and it does not include any copied parts without the appropriate citation.

\vspace{10mm}

Alexandria, Egypt,

\today

Ahmed Mohammed Elsayed Khalifa,

\end{declaration}

\newpage\null\thispagestyle{empty}\newpage
\phantomsection
\addcontentsline{toc}{chapter}{SUMMARY}

\begin{dedication} 	
This thesis is dedicated to my father 1958 $\textendash$ 2002, and to both Prof. Ahmed Abo-Ismail 1945 $\textendash$ 2015 and Dr. Ahmed Ramadan 1973 $\textendash$ 2015 who never got to read the final draft. You are the determination in every page.

\end{dedication}

\newpage\null\thispagestyle{empty}\newpage
\phantomsection
\addcontentsline{toc}{chapter}{ACKNOWLEDGMENT}

\begin{acknowledgment}      

\begin{center}
\textit{First of all, countless thanks to ALLAH the almighty} 
\end{center}
\vspace{5mm}

\forceindent While my name may be alone on the front cover of this thesis, I am by no means its sole contributor. Rather, there are a number of people behind this piece of work who deserve to be both acknowledged and thanked here: My academic supervisors; Prof. Mohamed Fanni, late Prof. Ahmed Abo-Ismail, late Dr. Ahmed Ramadan, and Prof. Toru Namerikawa.

\forceindent I would like to express my sincere condolences for the loss of Prof. Ahmed Abo-Ismail and Dr. Ahmed Ramadan. Prof. Ahmed Abo-Ismail is the founder of the Mechatronics and Robotics Engineering Department and he is the former Vice President of Education and Academic Affairs and one of E-JUST founders and pillars. Dr. Ahmed Ramadan made an extraordinary effort to build and establish most of the Labs in the Department of Mechatronics and Robotics Engineering at E-JUST. He was really thoroughly enjoyable, impeccably honest, straightforward, positive in his attitude and really has high morals. We pray for them to rest in peace and for their families to pass these difficult times. We ask ALLAH to meet them in the Paradise. 

I am forever indebted to my supervisor, Assoc. Prof. Mohamed Fanni, for his novel ideas, enthusiasm, guidance, and unrelenting support throughout this research. He has routinely gone beyond his duties to fire fight my worries and concerns, and have worked to instill great confidence in both myself and my work. THANK YOU FOR MAKING ME MORE THAN I AM.

I would like to show my sincere appreciation and gratitude to Prof. Abdelfatah Mohamed, the former chairperson of the Department of Mechatronics and Robotics Engineering, for his support, guidance and encouragement.  

I would like to convey my heartfelt thanks to my home university, Menofia University, Faculty of Electronic Engineering, and Department of Industrial Electronics and Control Engineering for their support and encouragement.

I would like to show my sincere appreciation and gratitude to Prof. Toru Namerikawa, Keio University, for accepting me in his research group and providing generous support, insightful advice and kind encouragement throughout my time at Keio University to date. The Namerikawa Lab has always been a place where people are ready to offer keen criticism and advice on any and all topics, academic or otherwise. For this I thank Mr. Okawa, Mr. Mori, Mr. Komagine, Mr. Hayashi, and Mr. Shinohara.  

Without the open-source software generously released by several individuals around the world, much of the implementation and experimental part of the work in this thesis would have been considerably
more difficult. I am therefore thankful to the authors of ROS, the ROS AscTec and Laser drivers, as well as those of Linux and an uncountable number of supporting packages for this wonderful operating system.

Thanks to my beloved family, my parents, my brother, and my sisters, for their self-sacrifice, consistent love, support, understanding and encouragement. 

Most of all, I thank my darling wife, whose love and strength have never flagged through the many ups and downs of this journey; and my children, whose smiles upon my return from a too-long day on campus always lift my spirits.

I would like to gratefully acknowledge the financial support of the Ministry of Higher Education, Government of Egypt, for my postgraduate scholarship.

\vspace{2 cm}  
\hfill Ahmed Khalifa

\hfill September 2016

\end{acknowledgment}


\addcontentsline{toc}{chapter}{TABLE OF CONTENTS}


\renewcommand{\contentsname}{TABLE OF CONTENTS}
{\tableofcontents} 


\renewcommand{\listtablename}{LIST OF TABLES}
\renewcommand{\listfigurename}{LIST OF FIGURES}

\listoftables   
\listoffigures	
\cleardoublepage
\markboth{\nomname}{\nomname}
\printnomenclature

\nomenclature[P]{$z_i$}{$z$- axis of frame $i$; $i= b, 0, 1, 2$}
\nomenclature[S]{$\Sigma_b$, $O_{b}$- $x_b$ $y_b$ $z_b$}{Robot body frame}
\nomenclature[S]{$\Sigma$, $O$- $x$ $y$ $z$}{World-fixed inertial reference frame}
\nomenclature[P]{$p_b$}{Position of body frame with respect to the world-fixed inertial reference frame}
\nomenclature[P]{$R_b$}{Rotation matrix from body frame to world frame}
\nomenclature[S]{$\Phi_b$=$[\psi \ \theta \ \phi]^T$}{ The triple $ZYX$ yaw-pitch-roll angles}
\nomenclature[S]{$\Sigma_e$, $O_{2}$- $x_2$ $y_2$ $z_2$}{ Frame attached to the end-effector of the manipulator}
\nomenclature[S]{$\Theta = [\theta_1 \ \theta_2]^T$}{The  vector of joint angles of the manipulator}
\nomenclature[S]{$d_{m_i}$, $a_{m_i}$, $\alpha_{m_i}$,$\theta_{m_i}$}{DH parameters of link $i$; $i$= $0, 1, 2$}
\nomenclature[P]{$l_{i}$}{Length of link $i$; $i$= $0, 1, 2$}
\nomenclature[P]{$A_i^{i-1}$}{Transformation matrix from frame $i$ to frame $i-1$}
\nomenclature[P]{$R_i^{i-1}$}{Rotation matrix from frame $i$ to frame $i-1$}
\nomenclature[P]{$R_i^{}$}{Rotation matrix from frame $i$ to world frame}
\nomenclature[P]{$R_{}^{i}$}{Rotation matrix from  world frame to frame $i$}
\nomenclature[P]{$O_{n \times m}$}{$n \times m$ null matrix}
\nomenclature[P]{$O_n$}{$n \times n$ null matrix}
\nomenclature[P]{$I_n$}{$n \times n$ identity matrix}
\nomenclature[P]{$p^b_{eb}$}{The position of $\Sigma_e$ with respect to $\Sigma_b$ expressed in $\Sigma_b$}
\nomenclature[P]{$p_e$}{The position of $\Sigma_e$ with respect to $\Sigma$}
\nomenclature[S]{$\chi_e = [x_e \ y_e \ z_e \ \psi_e \ \theta_e \ \phi_e]^T$}{The 6-DOF operational coordinates}
\nomenclature[P]{$q = [x\ y\ z\ \psi\ \theta\ \phi\ \theta_1\ \theta_2]^T$}{The quadrotor/joint space coordinates }
\nomenclature[S]{$\Phi_e$}{Euler angles of the end-effector}
\nomenclature[S]{$\dot{p}_e$}{The linear velocity of $\Sigma_e$ in the world-fixed frame}
\nomenclature[S]{$Skew(.)$}{The $(3 \times 3)$ skew-symmetric matrix operator}
\nomenclature[S]{$\omega_b$}{The angular velocity of the quadrotor expressed in $\Sigma$}
\nomenclature[S]{$\omega_e$}{The angular velocity $\omega_e$ of $\Sigma_e$}
\nomenclature[S]{$\omega^b_{eb}$}{The angular velocity of the end-effector relative to $\Sigma_b$ and is expressed in $\Sigma_b$}
\nomenclature[P]{$v^b_{eb} = [\dot{p}^{bT}_{eb} \ \omega^{bT}_{eb}]^T$}{The generalized velocity of the end-effector with respect to $\Sigma_b$}
\nomenclature[S]{$\dot{\Theta}$}{The joint velocities}
\nomenclature[P]{$J^b_{eb}$}{The manipulator Jacobian}
\nomenclature[P]{$v_e = [\dot{p}^T_e \ \omega^T_e]^T$}{The generalized end-effector velocity}
\nomenclature[P]{$T_b$}{The transformation matrix between the angular velocity $\omega_b$ and the time derivative of Euler angles $\dot{\Phi}_b$}
\nomenclature[P]{$J$}{The system jacobian}
\nomenclature[S]{$\zeta=[x \ y \ z \ \psi \ \theta_1 \ \theta_2]^T$}{Independent coordinate}
\nomenclature[S]{$\sigma_b= [\theta \ \phi]^T$}{The dependent coordinates}
\nomenclature[P]{$Q_b$}{Auxiliary variable to define $Q_b$}
\nomenclature[S]{$\chi_b$}{Quadrotor body coordinates}
\nomenclature[S]{$\eta_b$ = $[p_b \ \psi]^T$}{Auxiliary variable to reformulate $\chi_b$}
\nomenclature[P]{$J_{\eta}$}{Composed by the first 4 columns of $J_b Q_b$}
\nomenclature[P]{$J_{\sigma}$}{Composed by the last 2 columns of $J_b Q_b$}
\nomenclature[S]{$\Phi_e$}{ End-effector Euler angles}
\nomenclature[S]{$\dot{\chi}_e = [\dot{p}^T_e, \dot{\Phi}_e^T]^T$}{End-effector velocities in terms of Euler angles}
\nomenclature[P]{$Q_e$}{The same as $Q_b$ but it is a function of $\Phi_e$ instead of $\Phi_b$}
\nomenclature[S]{$\omega_{i}^{i}$}{The angular velocity of frame $i$ expressed in frame $i$}
\nomenclature[S]{$\dot{\omega}_{i}^{i}$}{The angular acceleration of frame $i$ expressed in frame $i$}
\nomenclature[P]{$v_{i}^{i}$}{The linear velocity of the origin of frame $i$ expressed in frame $i$}
\nomenclature[P]{$\dot{v}^{i}_{c_i}$}{The linear acceleration of the center of mass of link $i$ expressed in frame $i$}
\nomenclature[P]{$\dot{v}_{i}^{i}$}{The linear acceleration of the origin of frame $i$ expressed in frame $i$}
\nomenclature[P]{$r_{i}^{i}$}{The vector from the origin of frame $i-1$ to the origin of link $i$}
\nomenclature[P]{$r_{c_i}^{i}$}{The vector from the origin of frame $i-1$ to the center of mass of link $i$}
\nomenclature[P]{$g^{}$}{The vector of gravity expressed in $\Sigma$}
\nomenclature[P]{$z_{i-1}^{i-1}$}{A unit vector pointing along the $i^{th}$ joint axis and expressed in the $(i-1)^{th}$ link coordinate system}
\nomenclature[P]{$I_{i}^{i}$}{The inertia matrix of link $i$ about its center of mass coordinate frame}
\nomenclature[P]{$f_{i,i-1}^{i}$ / $n_{i,i-1}^{i}$}{The resulting force/moment exerted on link $i$ by link $i-1$ at point $O_{i-1}$}
\nomenclature[P]{$m_{p}$}{mass at the end-effector}
\nomenclature[P]{$CG_{2}$}{ The point of center of gravity of link 2}
\nomenclature[S]{$(\grave{.})$}{ Refers to the value of the parameter after adding the payload}
\nomenclature[P]{$g^{i}$}{The gravity vector expressed in frame $i$ is}
\nomenclature[P]{$g_r$}{The gravity acceleration}
\nomenclature[S]{$\tau_{m_i}$}{The torques acting on joint $i$}
\nomenclature[P]{$b_i$}{The friction coefficient on joints $i$}
\nomenclature[P]{$F_{m,q}^{b}$, $M_{m,q}^{b}$}{The interaction forces and moments of the manipulator acting on the quadrotor expressed in $\Sigma_b$}
\nomenclature[P]{$F_{m,q}^{}$}{The interaction forces expressed in the inertial frame}
\nomenclature[P]{$M_1(q)$, $M_2(q)$, $N_1(q,\dot{q},\ddot{q})$, $N_2(q,\dot{q},\ddot{q})$}{Nonlinear terms of the manipulator dynamics}
\nomenclature[P]{$F_{m,q_x}$, $F_{m,q_y}$, $F_{m,q_z}$}{The interaction forces resulted from the manipulator and affected the quadrotor in $x$, $y$, and $z$ directions expressed in the inertial frame}
\nomenclature[P]{$M_{m,q_\phi}^b$, $M_{m,q_\theta}^b$, and $M_{m,q_\psi}^b$}{The interaction moments from the manipulator to the quadrotor around $x_b$, $y_b$, and $z_b$ directions}
\nomenclature[P]{$m$}{The mass of the quadrotor}
\nomenclature[S]{$\tau_{a_1}$, $\tau_{a_2}$, $\tau_{a_3}$}{The three input moments about the three body axes, $x_b$, $y_b$, $z_b$}
\nomenclature[S]{$\Omega_j$}{Angular velocity of rotor $j$; $j$ = 1, 2, 3, 4}
\nomenclature[P]{$F_j$}{Thrust force of rotor $j$}
\nomenclature[P]{$M_j$}{Drag moment of rotor $j$}
\nomenclature[P]{$K_{f_j}$}{Thrust coefficient of rotor $j$}
\nomenclature[P]{$K_{m_j}$}{Drag coefficient of rotor $j$}
\nomenclature[P]{$T$}{The total thrust applied to the quadrotor from all four rotors}
\nomenclature[P]{$d_q$}{The distance between the quadrotor center of mass and rotor rotational axis}
\nomenclature[S]{$\overline{\Omega}$}{Collective speed of the rotational speed of the four rotors}
\nomenclature[P]{$I_r$}{The rotor inertia}
\nomenclature[P]{$I_x, I_y, I_z$}{The inertia of the vehicle around its body-frame about $x_b-$, $y_b-$, and $z_b-$axis respectively}
\nomenclature[P]{$M$}{The symmetric and positive definite inertia matrix of the combined system}
\nomenclature[P]{$C$}{The matrix of Coriolis and centrifugal terms}
\nomenclature[P]{$G$}{The vector of gravity terms}
\nomenclature[P]{$d_{ex}$}{The vector of external disturbances}
\nomenclature[P]{$u = [F_1 \ F_2 \ F_3 \ F_4 \ \tau_{m_1} \ \tau_{m_2}]^T$}{Vector of the actuator inputs}
\nomenclature[P]{$B_m$}{Input matrix which is used to generate the body forces and moments from the actuator inputs}
\nomenclature[P]{$N_m$}{Control matrix}
\nomenclature[P]{$K_{{\tau}_1}$, $K_{{\tau}_2}$}{Motor's constants of joints $1$ and $2$}
\nomenclature[P]{$H_m$}{Matrix that transforms the body input forces to be expressed in $\Sigma$}
\nomenclature[P]{$\ddot{x}_f, \ddot{y}_f, \ddot{z}_f$}{Auxiliary variables to simplify the equitations of the nonholonomic constraints}
\nomenclature[P]{$\ddot{p}^b$}{Body accelerations in the world frame}
\nomenclature[P]{$s_z$}{The height of the sonar}
\nomenclature[P]{$\ddot{p}_b^b$}{The body-fixed acceleration}
\nomenclature[P]{$p_{b,x}$}{Body position in the world frame along $x$-axis}
\nomenclature[P]{$\dot{p}_{b,x}$}{Body linear velocity in the world frame along $x$-axis}
\nomenclature[P]{$b_x$}{Acceleration sensor bias}
\nomenclature[P]{$X_{obs}$}{State variable of the Luenberger observer}
\nomenclature[P]{$U_{obs}$}{Input of the Luenberger observer}
\nomenclature[P]{$Y_{obs}$}{Measurement of the Luenberger observer}
\nomenclature[P]{$\hat{X}_{obs}$}{Estimated state variable of the Luenberger observer}
\nomenclature[P]{$A_{obs}$}{The A matrix of the state equation of the Luenberger observer}
\nomenclature[P]{$B_{obs}$}{The B matrix of the state equation of the Luenberger observer}
\nomenclature[P]{$C_{obs}$}{ The C matrix of the state equation of the Luenberger observer}
\nomenclature[P]{$L_{obs}$}{The gain matrix of the state equation of the Luenberger observer}
\nomenclature[P]{$G_p$}{Gravity term of the translational part in the dynamic equation of motion}
\nomenclature[P]{$K_p$, $K_d$, $K_i$}{PID tuning parameters}
\nomenclature[P]{$p_{b_r}$}{Reference position}
\nomenclature[S]{$\dot{p}_{b_r}$}{Reference velocity}
\nomenclature[S]{$\theta_r$}{Desired pitch}
\nomenclature[S]{$\phi_r$}{Desired roll}
\nomenclature[S]{$\tau_p$}{Control signal of the translational motion}
\nomenclature[S]{$\chi_{e,r}(t)$}{Reference trajectory for the end-effector pose}
\nomenclature[P]{$p_{e,r}(t)$}{Reference trajectory for the end-effector position}
\nomenclature[S]{$\Phi_{e,r}(t)$}{Reference trajectory for the end-effector orientation}
\nomenclature[S]{$\Delta t$}{Very small positive number of time}
\nomenclature[S]{$\zeta_r(t_i)$}{Reference independent quadrotor/joint space coordinates}
\nomenclature[P]{$r_{ij}$}{The $ij^{th}$ element of the rotation matrix, $R_e$}
\nomenclature[P]{$a_i, b_i, c_i, d_i$}{Auxiliary variables used during solving the inverse kinematics problem, $i$ = }
\nomenclature[S]{$\tau$}{Vector of the input generalized forces}
\nomenclature[P]{$M_n$}{System nominal inertia matrix}
\nomenclature[S]{$\tau^{des}$}{Desired input to the robot}
\nomenclature[P]{$g_i$}{Bandwidth of the low pass filter of the $i^{th}$ variable of $q$; $i$ = 1, 2, \ldots, 8}
\nomenclature[P]{$P$}{Vector of $g_i$s}
\nomenclature[P]{$Q(s)$}{Matrix of the low pass filter of DOb}
\nomenclature[S]{$\tau^{dis}$}{System disturbances including the Coriolis, centrifugal and gravitational terms, and external disturbances}
\nomenclature[S]{$\hat{\tau}^{dis}$}{System estimated disturbances}
\nomenclature[P]{$Q_v$}{Matrix of low pass filter for the estimated velocity}
\nomenclature[P]{$P_v$}{Vector of the cut-off frequency}
\nomenclature[P]{$M_{n_a}$}{Nominal inertia for the translational coordinates}
\nomenclature[P]{$M_{n_v}$}{Nominal inertia for the rotational coordinates}
\nomenclature[S]{$\ddot{q}^{des}$}{Output from the outer loop controller}
\nomenclature[P]{$e_v$}{Error of the DOb internal loop}
\nomenclature[P]{$K_v$}{Auxiliary variable to define the error dynamics of the DOb loop}
\nomenclature[S]{$\delta$}{Auxiliary variable to define the error dynamics of the DOb loop}
\nomenclature[P]{$V$}{Lyapunov function}
\nomenclature[P]{$L_p$}{$L_p$ space; $p$ = [1,$\infty$)}
\nomenclature[S]{$\varUpsilon_1$, $\varUpsilon_2$}{Auxiliary variables to define Lemma \ref{lm:1}}
\nomenclature[S]{$V_c$, $\gamma$}{Auxiliary variables used for stability proof of PD-DOb control}
\nomenclature[P]{$H_{PD}$}{Strictly proper and exponentially stable transfer function for $Lemma$ \ref{lm:2}}
\nomenclature[S]{$\sigma_{b,r}(t)$}{Reference trajectories for the roll and pitch angles }
\nomenclature[S]{$\tau_{\zeta}$}{Control signal for the independent coordinates}
\nomenclature[S]{$\tau_{\sigma}$}{Control signal for the dependent coordinates}
\nomenclature[S]{$\ddot{\zeta}^{des}$}{Desired acceleration for the independent coordinates}
\nomenclature[S]{$\ddot{\sigma}^{des}$}{Desired acceleration for the dependent coordinates}
\nomenclature[S]{$\tau_{\zeta}^{des}$}{Desired torque for the independent coordinates}
\nomenclature[S]{$\tau{\sigma}^{des}$}{Desired torque for the dependent coordinates}
\nomenclature[P]{$u_{max}, u_{min}$}{Maximum and minimum limits of the actuators inputs}
\nomenclature[P]{$x_p$}{State vector of the plant SISO state space model used to design the MPC}
\nomenclature[P]{$u_{mpc}$}{Input of the plant SISO state space model used to design the MPC}
\nomenclature[P]{$A_p, B_p, C_p$}{The $A, B, C$ matrices of the plant SISO state space model used to design the MPC}
\nomenclature[P]{$t_i$}{Initial time of the prediction window of MPC}
\nomenclature[P]{$T_p$}{Length of the prediction window of MPC}
\nomenclature[P]{$x_a$}{State vector of the plant SISO augmented state space model used to design the MPC}
\nomenclature[P]{$\dot{u}_{mpc}$}{Input of the plant SISO augmented state space model used to design the MPC}
\nomenclature[P]{$A_a, B_a, C_a$}{The $A, B, C$ matrices of the plant SISO augmented state space model used to design the MPC}
\nomenclature[P]{$X(t)$}{Set of the Laguerre functions}
\nomenclature[P]{$A_u$}{Matrix used to define Laguerre functions}
\nomenclature[P]{$n_u$}{Number of terms of Laguerre functions}
\nomenclature[P]{$s_u$}{Time scaling factor of Laguerre functions}
\nomenclature[S]{$\iota$}{Time variable within the prediction window of MPC}
\nomenclature[P]{$J_{mpc}$}{Cost function of he MPC}
\nomenclature[S]{$\Phi_{mpc}$}{Auxiliary variable to simplify the prediction state equation of MPC}
\nomenclature[P]{$h$}{Step variable to solve the state equation in recursive manner}
\nomenclature[S]{$\Omega_{mpc}$, $\Psi_{mpc}$}{Auxiliary variables to define the cost function}
\nomenclature[P]{$Q_{mpc}, R_{mpc}$}{Scaling matrices used in the cost function of MPC}
\nomenclature[S]{$\eta$}{Variable to parametrize the optimal control signal}
\nomenclature[P]{$K_{mpc}$}{Equivalent state feedback gain of MPC}
\nomenclature[S]{$C_u$, $M_{mpc}$, $\gamma_{mpc}$}{Auxiliary variables to define the inequality constraints of MPC}
\nomenclature[S]{$\Lambda_i^{k+1}$}{The $i^{th}$ Lagrange multiplier at iteration $k+1$}
\nomenclature[P]{$W_i^{k+1}$}{The weight for $i^{th}$ Lagrange multiplier at iteration $k+1$}
\nomenclature[P]{$h_{ij}$}{The $ij^{th}$ element of the matrix $H_{mpc}$}
\nomenclature[P]{$e_i$}{The $i^{th}$ element of the vector $E_{mpc}$}
\nomenclature[P]{$H_{mpc}, E_{mpc}$}{Matrices used to solve the optimization problem}
\nomenclature[S]{(.)$^*$}{The optimal value of the variable (.)}
\nomenclature[P]{$\hat{F}_e$}{Estimated end-effector contact generalized force}
\nomenclature[S]{$\tau_l$}{End-effector contact generalized force expressed in the Quadrotor/joint space }
\nomenclature[P]{$S_c$}{Environment stiffness}
\nomenclature[P]{$D_c$}{Environment damping}
\nomenclature[S]{$\tau_w$}{Force represents the wind effect}
\nomenclature[P]{$V_{wz}$}{Wind velocity at altitude $z$}
\nomenclature[P]{$V_{w_{z_0}}$}{Specified (measured) wind velocity at altitude $z_0$}
\nomenclature[P]{$A_e$}{Influence effective area of wind on the quadrotor}
\nomenclature[P]{$f_{wx_1}, f_{wx_2}, f_{wy_1}, f_{wy_2}$}{Auxiliary variables to parametrize the wind effect}
\nomenclature[S]{$\psi_w$}{The angle of wind direction}
\nomenclature[S]{$\tau_{int}$}{The system internal dynamics}
\nomenclature[P]{$Y_i(q,\dot{q},\ddot{q})$}{Data regressor of the internal dynamics}
\nomenclature[P]{$h_i$}{Vector of the platform parameters}
\nomenclature[S]{$\tau_l$}{The environment dynamics}
\nomenclature[P]{$Y_l(q,\dot{q},\ddot{q},\chi_e,\dot{\chi}_e)$}{Data regressor of the environment dynamics}
\nomenclature[P]{$h_l$}{Vector of the environment parameters}
\nomenclature[P]{$Y_e$}{Function of the end effector states}
\nomenclature[P]{$Y_w(z,\theta,\phi)$}{Data regressor of the wind effect}
\nomenclature[P]{$h_w$}{Vector of the wind parameters}
\nomenclature[P]{$Y_s$}{The data regressor of the whole system dynamics}
\nomenclature[P]{$h_s$}{Vector of the whole system dynamics parameters}
\nomenclature[S]{$\ddot{\chi}^{des}_e$}{The desired acceleration in the task space}
\nomenclature[P]{$S_{c,d}$, $D_{c,d}$}{Desired values of $S_{c}$ and $D_{c}$ respectively}
\nomenclature[S]{$\ddot{\zeta}^{des}$}{The desired acceleration in the quadrotor/joint space}
\nomenclature[S]{$\tilde{h}_s(t)$}{The parameter estimation error}
\nomenclature[S]{$\tilde{\tau}(t)$}{Dynamics estimation error}
\nomenclature[P]{$R(t)$}{The parameters' covariance matrix}
\nomenclature[S]{$\eta_h$}{The forgetting factor}
\nomenclature[S]{$\eta_h^{min}$}{Constant representing the minimum forgetting factor}
\nomenclature[S]{$\gamma_g$}{Design constant}
\nomenclature[P]{$k_{max}$}{Maximum number of iteration of the MPC}

%


\nomenclature[A]{VTOL}{Vertical Take Off and Landing}
\nomenclature[A]{UAV}{Unmanned Aerial Vehicle}
\nomenclature[A]{DOF}{Degrees Of Freedom}
\nomenclature[A]{MPC}{Model Predictive Control}
\nomenclature[A]{DOb}{Disturbance Observer}
\nomenclature[A]{SISO}{Single Input Single Output}
\nomenclature[A]{IMU}{Inertial Measurement Unit}
\nomenclature[A]{RLS}{Recursive Least Squares}
\nomenclature[A]{FTRLS}{Fast Tracking Recursive Least Squares}
\nomenclature[A]{PD}{Proportional Derivative}
\nomenclature[A]{DH}{Denavit-Hartenberg}
\nomenclature[A]{Max}{Maximum}
\nomenclature[A]{Min}{Minimum}
\nomenclature[A]{$C_{.}$}{Short notation for $\cos(.)$}
\nomenclature[A]{$S_{.}$}{Short notation for $\sin(.)$}
\nomenclature[A]{AscTec}{Ascending Technologies}
\nomenclature[A]{FCU}{Flight Control Unit }
\nomenclature[A]{LLP}{Low Level Processor }
\nomenclature[A]{HLP}{High Level Processor }
\nomenclature[A]{CAM}{Camera}
\nomenclature[A]{GPS}{Global Positioning System}
\nomenclature[A]{WiFi}{Technology that allows electronic devices to connect to a wireless LAN (WLAN) network}
\nomenclature[A]{miniPCI}{ Internal PCI slot for the onboard computer}
\nomenclature[A]{PCI}{Peripheral Component Interconnect}
\nomenclature[A]{SD}{Secure Digital}
\nomenclature[A]{WPS2}{Wireless Play Station 2}
\nomenclature[A]{SSC}{Serial Servo Controller}
\nomenclature[A]{Li-Po}{Lithium-Polymer}
\nomenclature[A]{R/C}{Radio Controlled}
\nomenclature[A]{ROS}{Robot Operating System}
\nomenclature[A]{PC}{Personal Computer}
\nomenclature[A]{SDK}{Software Development Kit}
\nomenclature[A]{3D}{3 Dimensional}
\nomenclature[A]{CAD}{Computer-Aided Design}
\nomenclature[A]{NI}{National Instruments}
\nomenclature[A]{DAC}{Data Acquisition Card}
\nomenclature[A]{RMSE}{Root Mean Squared Error}
\nomenclature[A]{PID}{Proportional Integral Derivative}
\nomenclature[A]{MPC$_{\sigma}$}{The MPC for the dependent coordinates}
\nomenclature[A]{MPC$_{\zeta}$}{The MPC for the independent coordinates}
\nomenclature[A]{w.r.t}{with respect to }
\nomenclature[A]{NINT}{Nearest INTeger (round-off operator)}
\nomenclature[A]{$mtr$}{Motor}
\nomenclature[A]{MCU}{MicroController Unit}

\cleardoublepage

\titlespacing{\section}{0pt}{10pt}{5pt}
\titlespacing{\subsection}{0pt}{10pt}{5pt}

\begin{spacing}{1.5}
	 

\mainmatter




\chapter{SYSTEM DESIGN AND MODELING} 
\label{ch:SystemDesignandModeling}


\ifpdf
    \graphicspath{{2_SystemDesignandModeling/figures/}}
\else
    \graphicspath{{2_SystemDesignandModeling/figures/}}
\fi

This chapter initially describes the proposed flying robot. After that, the design of this system is presented. Then, the kinematic model is described. Finally, the system dynamics are reviewed.

\section{System Description}
\begin{figure}
	\centering
	\includegraphics[width=0.6\textwidth]{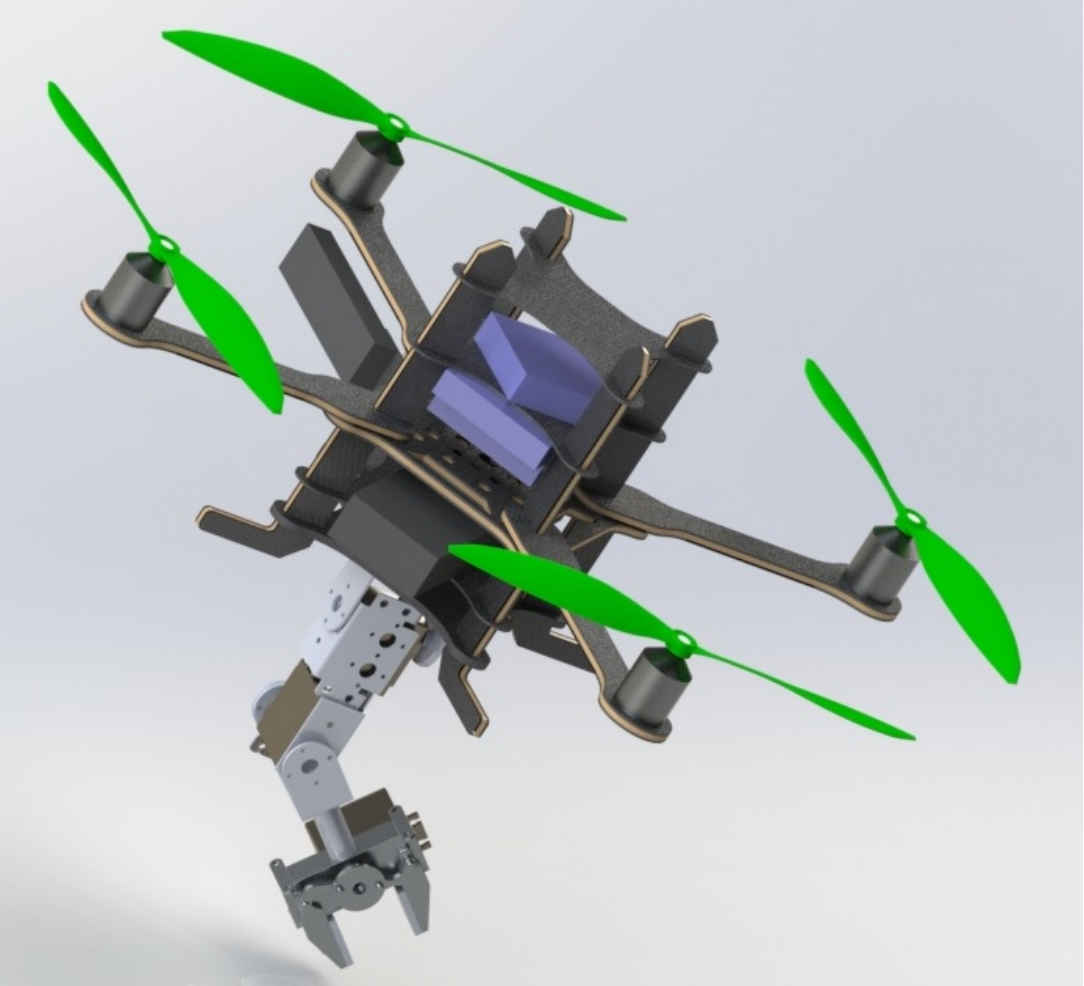}
	\caption{$3$D CAD model of the proposed system}
	\label{fig:3d_exp_setup}
\end{figure}

\begin{figure}[h]
	\centering
	\includegraphics[width=0.95\columnwidth]{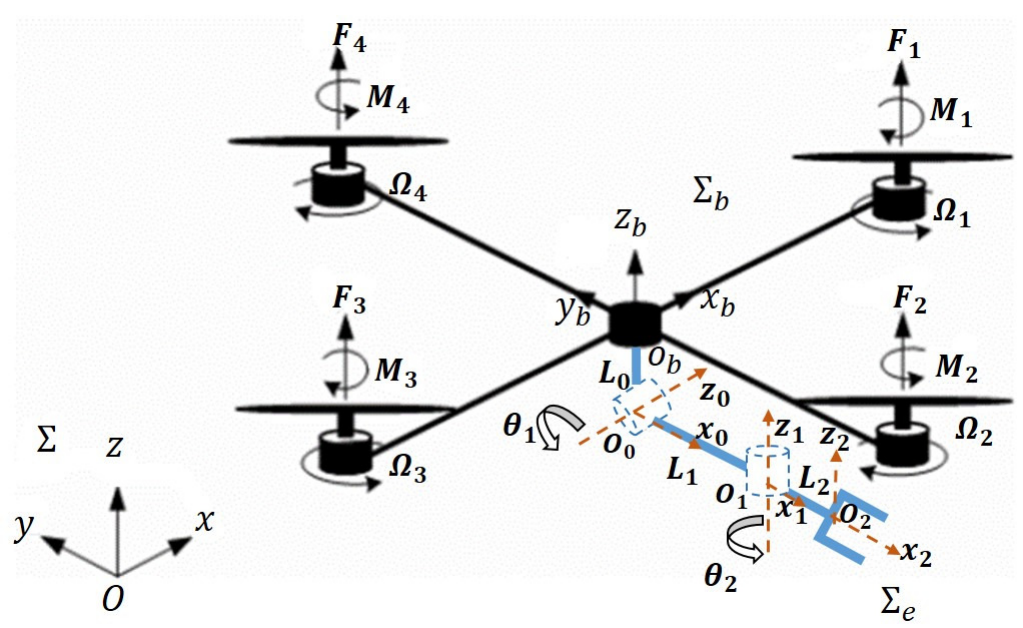}
	\caption{Schematic of Quadrotor Manipulation System with relevant frames}
	\label{frames}
\end{figure}

A 3D CAD model of the proposed system is shown in Fig. \ref{fig:3d_exp_setup}. The system consists mainly of two parts; the quadrotor and the manipulator.
Fig. \ref{frames} presents a sketch of the proposed system with the relevant frames, which indicates the unique topology that permits the end-effector to achieve arbitrary 6-DOF trajectory. The frames are assumed to satisfy the Denavit-Hartenberg (DH) convention \cite{spong2006robot}. The manipulator has two revolute joints. The axis of the first revolute joint ($z_0$), which is fixed to the quadrotor, is parallel to the body $x$-axis of the quadrotor (see Fig. \ref{frames}). The axis of the second joint ($z_1$) is perpendicular to the axis of the first joint and  will be parallel to the body $y$-axis of the quadrotor at home (extended) configuration. Thus, the pitching and rolling rotation of the end-effector is now possible independently from the horizontal motion of the quadrotor. Hence, with this new system, the capability of manipulating objects with arbitrary location and orientation is achieved. By this non-redundant system, the end-effector can achieve 6-DOF motion with minimum number of actuators/links, which is an important factor in flight. The proposed system is distinguished from all other previous systems in the literature by having maximum mobility with minimum weight. The resulted complexity of the inverse kinematics and control are handled latter to prove the capability of the end-effector to track the reference 6-DOF trajectory.
\section{System Design}

\subsection{Quadrotor}
\begin{figure}[h]
	\centering
	\begin{tabular}{ccc}
		\subfloat[Pelican - Perspective]{\includegraphics[width=0.33\columnwidth]{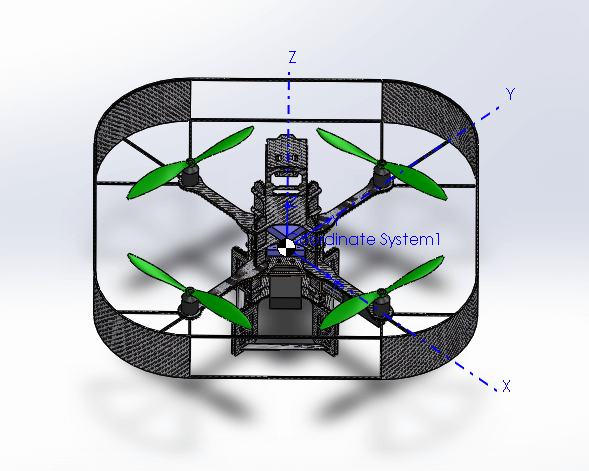}}&
		\subfloat[Pelican - Front]{\includegraphics[width=0.33\columnwidth]{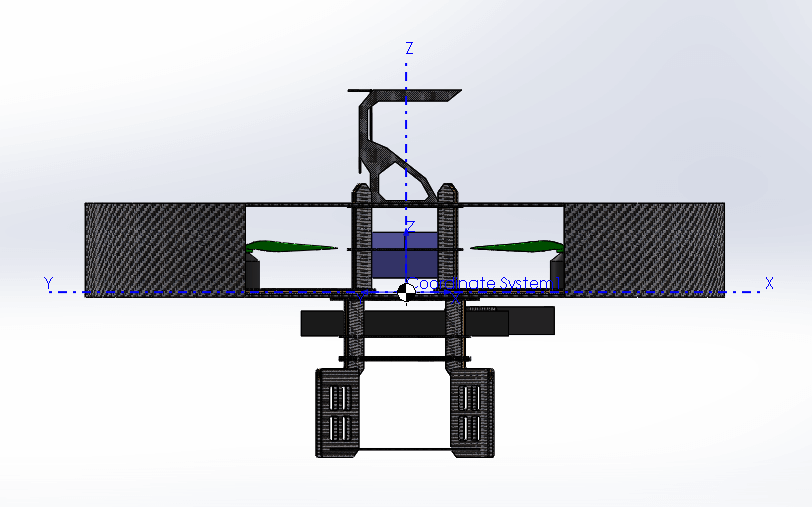}}&
		\subfloat[Pelican - Top]{\includegraphics[width=0.33\columnwidth]{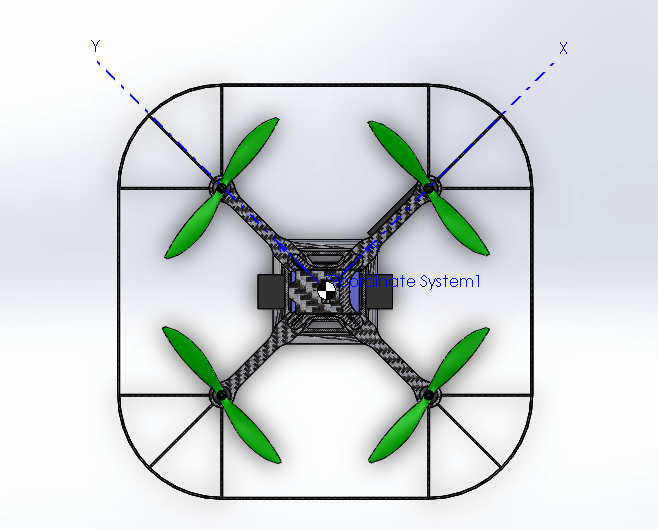}}
	\end{tabular}
	\caption{Different views for the quadrotor platform extracted from a SOLIDWORKS model}
	\label{fig:pelican_cad}
\end{figure}

The quadrotor components are selected such that it can carry an additional weight equals $650$  g (larger than the total arm weight and the maximum payload). AscTec pelican quadrotor \cite{asctec} form Ascending Technologies is used as the quadrotor platform. Fig. \ref{fig:pelican_cad} presents the different views,  which are extracted from a SOLIDWORKS model, for the quadrotor platform. 

The characteristics of this quadrotor can be listed as following:

\begin{itemize}
	 \item Lightweight and customizable tower structure that lets one  mount diverse payloads (e.g., sensors, avionics, and manipulator).
    \item Carbon fiber material with mass = 1.3 kg, so it combines between the rigidity and lightweight.
	\item Maximum payload = 650 g. This should include the additional sensors, avionics, manipulator and the payload to be handled.
	\item Maximum thrust = 36 N.
	\item Maximum airspeed = 16 m/s, and maximum climb rate = 8 m/s.
\end{itemize}
See Appendix \ref{app:experimentalsystem-quad} for more technical details about the AscTec quadrotor.
 
\subsection{Manipulator}

A lightweight manipulator that can carry a payload of 200 g and has a maximum reach of 22 cm is deigned. Its weight  is 200 g. The arm components are selected, purchased and assembled.

The arm components are: 
\begin{itemize}
	\item Three DC motors; HS-422 (Max torque = 0.4  N.m) for the gripper, HS-5485HB  (Max torque = 0.7  N.m) for joint 1, and HS-422 (Max torque = 0.4  N.m) for joint 2 \cite{lynxmotion}.
	\item The manipulator structure accessories are; Aluminum Tubing - 1.50  in diameter, Aluminum Multi-Purpose Servo Bracket, Aluminum Tubing Connector Hub, and Aluminum Long 'C' Servo Bracket with Ball Bearings \cite{lynxmotion}. 
	\item Lightweight grasping mechanism \cite{lynxmotion}.   
\end{itemize}	
See Appendix \ref{app:experimentalsystem-manp} for details about the manipulator's dimensions and design.

\begin{figure}[!h]
	\centering
	\begin{tabular}{c}
		\subfloat[Deflection analysis]{\includegraphics[width=0.7\textwidth]{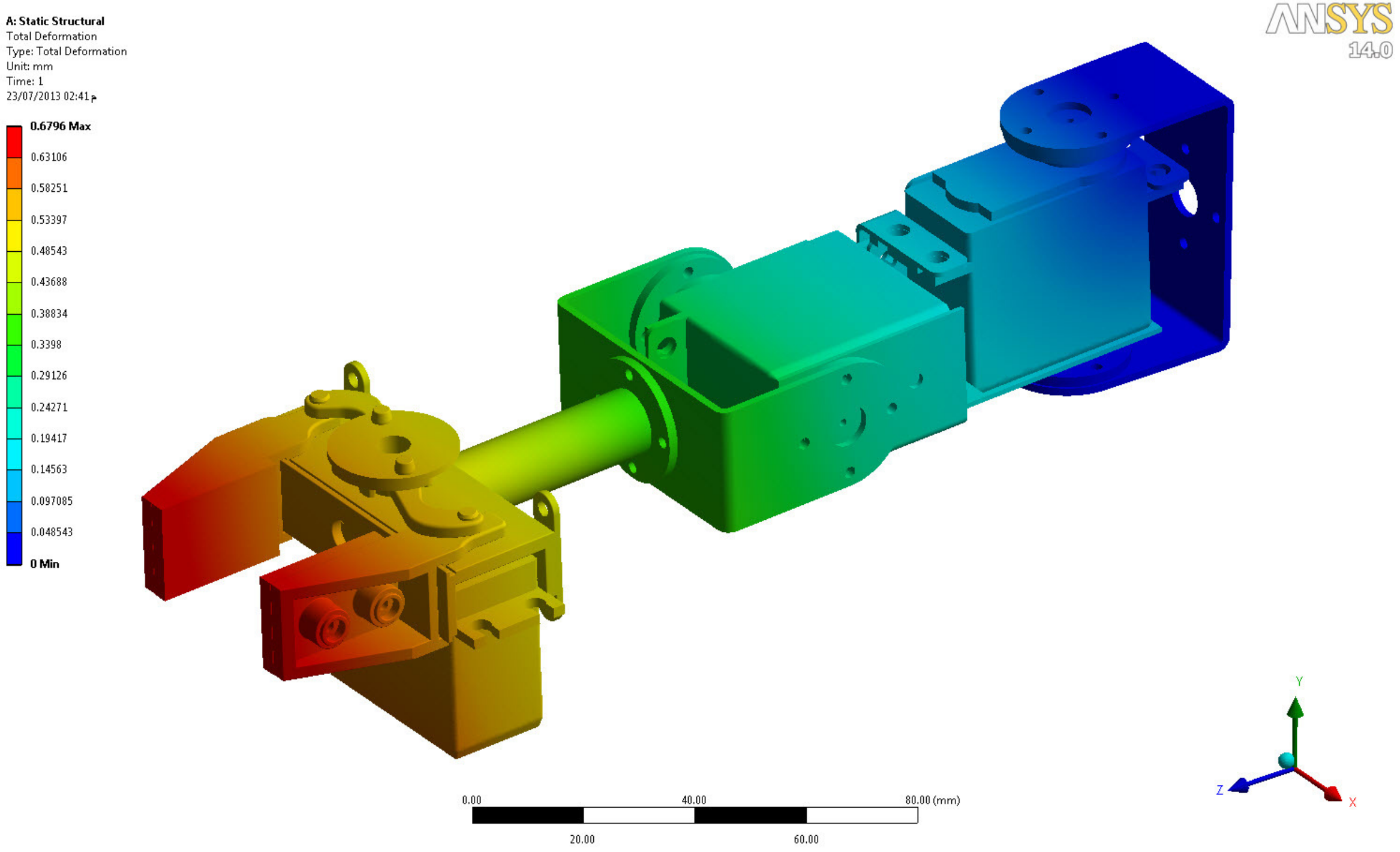}}\\
		\subfloat[Stress analysis]{\includegraphics[width=0.7\textwidth]{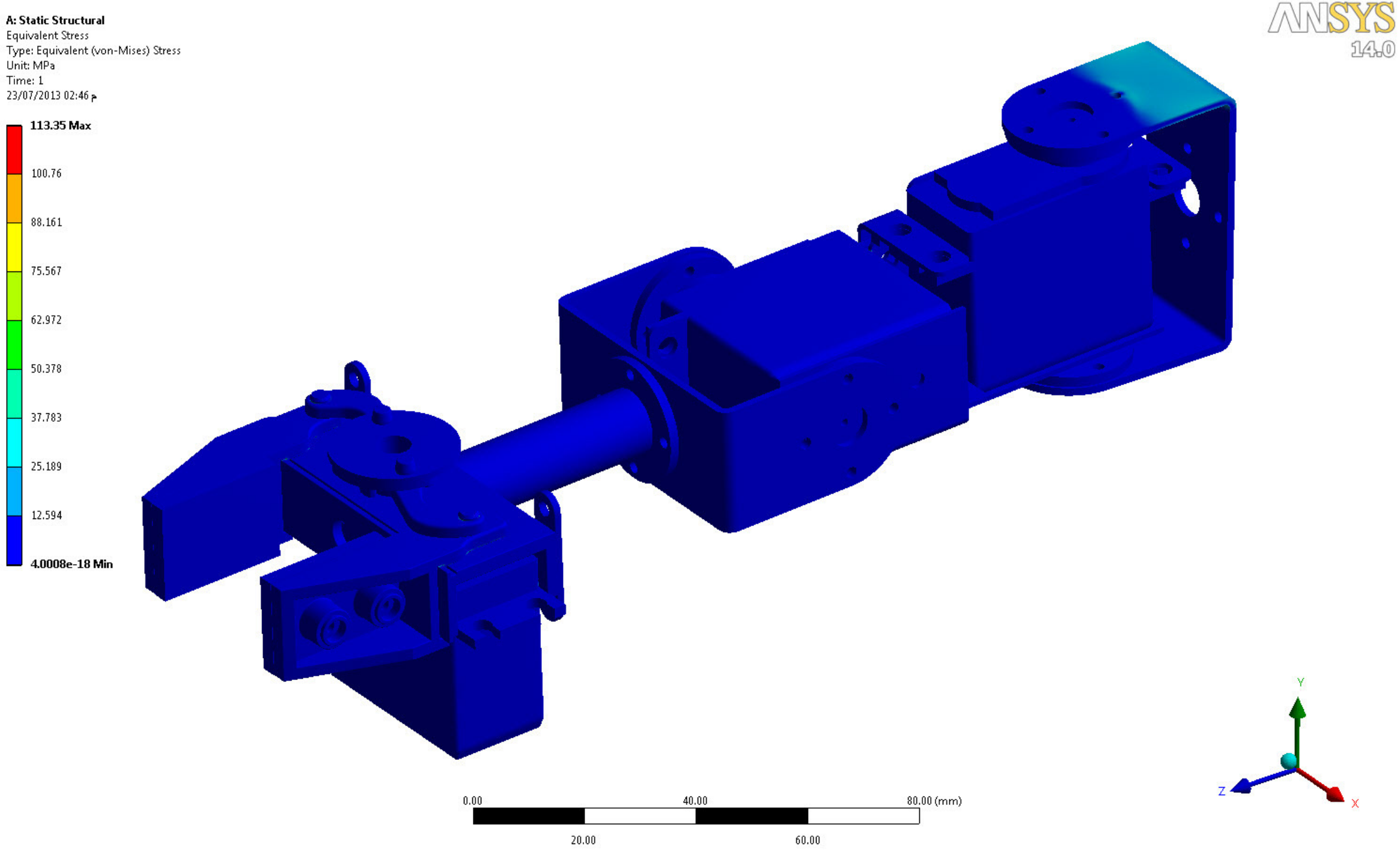}}
	\end{tabular}
	\caption{Manipulator's structure analysis}
	\label {FEA_ANSYS}
\end{figure}

The safety of the manipulator design, with respect to the strength and rigidity, is checked through finite element analysis using ANSYS software (see Fig. \ref{FEA_ANSYS}). 
From these figures, the maximum deflection is about 0.6 mm, which is smaller than the allowable value which equals 1  mm. In addition, the maximum stress of the structure is 113 MPa, which is smaller than the yield strength of aluminum alloy that is 270 MPa. Also, the bearings and gripper are selected to sustain the loads. Therefore, this design is safe.

\begin{figure}[!h]
	\centering
	\includegraphics[width=0.8\columnwidth]{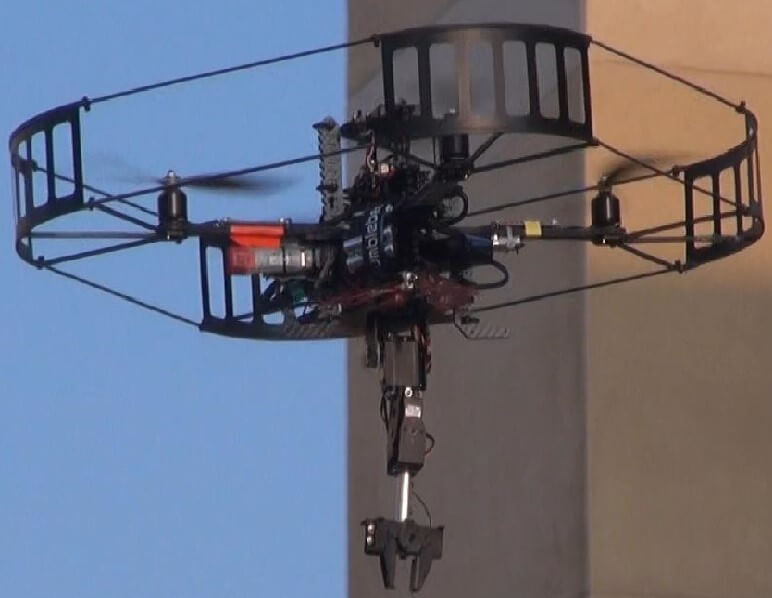}
	\caption{Experimental Validation of the system design}
	\label{fig:desgn_exp_val}
\end{figure}

The design of the proposed system is validated via experimental test as shown in Fig. \ref{fig:desgn_exp_val}.
\section{Forward Kinematics}\label{se:for_kin}
In this section, the position and velocity kinematic analysis are presented. Let $\Sigma_b$, $O_{b}$- $x_b$ $y_b$ $z_b$, denotes the vehicle
body-fixed reference frame with its origin at the quadrotor center of mass, see Fig. \ref{frames}. Its position with respect to the 
world-fixed inertial reference frame, $\Sigma$, $O$- $x$ $y$ $z$, is given by the vector $p_b=[x \ y \ z]^T$ $\in \mathbb{R}^{3}$, while its 
orientation is given by the rotation matrix $R_b$ as
\begin{equation}
R_b= \begin{bmatrix}
C_{\psi} C_{\theta} &  S_{\phi} S_{\theta} C_{\psi}-S_{\psi} C_{\phi} & S_{\psi} S_{\phi}+C_{\psi} S_{\theta} C_{\phi} \\
S_{\psi} C_{\theta} & C_{\psi} C_{\phi}+ S_{\psi} S_{\theta} S_{\phi} & S_{\psi} S_{\theta} C_{\phi}-C_{\psi} S_{\phi} \\
-S_{\theta} & C_{\theta} S_{\phi} & C_{\theta} C_{\phi} \\
\end{bmatrix},
\label{eq:Rb}
\end{equation}
where $\Phi_b$=$[\psi \ \theta \ \phi]^T$ $\in \mathbb{R}^{3}$ is the triple $ZYX$ yaw-pitch-roll angles. Note that $C_{.}$ and $S_{.}$ are short notations for $\cos(.)$ and $\sin(.)$, respectively.
Let us consider the frame  $\Sigma_e$, $O_{2}$- $x_2$ $y_2$ $z_2$, attached to the end-effector of the manipulator, see Fig. \ref{frames}. Let us define $\Theta = [\theta_1 \ \theta_2]^T$ $\in \mathbb{R}^{2}$ to be the  vector of joint angles of the manipulator. Table \ref{D_H} presents the DH parameters for the two link manipulator.
\setlength{\extrarowheight}{3pt}
\begin{table}
	\caption{DH parameters for the manipulator}
	\label{D_H}
	\begin{center}
		\begin{tabular}{|c|c|c|c|c|}
			\hline
			Link($i$) & $d_{m_i}$ & $a_{m_i}$ & $\alpha_{m_i}$ & $\theta_{m_i}$ \\
			\hline
			$0$	& $-l_{0}$ & $0$  & $-\pi/2$ & $-\pi/2$ \\
			\hline
			$1$	& $0$ & $l_{1}$  & $\pi/2$ & $\theta_{1}$ \\
			\hline
			$2$	& $0$ & $l_{2}$ & $0$ & $\theta_{2}$ \\
			\hline
		\end{tabular}
	\end{center}
\end{table}
Transformation from $\Sigma_e$ to $\Sigma$ is given by
\begin{equation}
A_e = A_b A_0^b A_1^0 A_2^1. 
\label{eq:Te}
\end{equation}

Transformation from $\Sigma_e$ to $\Sigma_b$ is given by
\begin{gather}
A_e^b = A_2^b= A_0^b A_1^0 A_2^1 \\ \nonumber
\qquad \qquad  = \begin{bmatrix}
R^b_e & p^b_{eb}\\
O_{1 \times 3} & 1
\end{bmatrix},
\label{eq:Teb}
\end{gather}
where $O_{1 \times 3}$ represents a $1 \times 3$ zeros vector, $p^b_{eb}$ describes the position of $\Sigma_e$ with respect to $\Sigma_b$ expressed in $\Sigma_b$, and $R^b_e$ describes the orientation of $\Sigma_e$ w.r.t $\Sigma_b$. The matrices $A^{b}_{0}$, $A^{0}_{1}$, and $A^{1}_{2}$ can be obtained by
\begin{equation}
A^{b}_{0}= \begin{bmatrix}
0 & 0 & 1 & 0 \\
-1 & 0 & 0 & 0 \\
0 & -1 & 0 & -l_{0} \\
0 & 0 & 0 & 1 \\
\end{bmatrix},
\label{Ab_0}
\end{equation}
\begin{equation}
A^{0}_{1}= \begin{bmatrix}
 C_{\theta_1} & 0 &  S_{\theta_1} & l_{1}  C_{\theta_1} \\
 S_{\theta_1} & 0 & - C_{\theta_1} & l_{1}  S_{\theta_1}\\
0 & 1 & 0 & 0\\
0 & 0 & 0 & 1 \\
\end{bmatrix},
\label{A0_1}
\end{equation}
\begin{equation}
A^{1}_{2}= \begin{bmatrix}
 C_{\theta_2} & - S_{\theta_2} & 0 & l_{2}  C_{\theta_2} \\
 S_{\theta_2} &  C_{\theta_2} & 0 & l_{2}  S_{\theta_2}\\
0 & 0 & 1 & 0\\
0 & 0 & 0 & 1 \\
\end{bmatrix}.
\label{A1_2}
\end{equation}

Thus, the position of $\Sigma_e$ with respect to (w.r.t) $\Sigma$ is given by
\begin{equation}
p_e = p_b + R_b p^b_{eb}.
\label{eq:pe}
\end{equation}

The orientation of $\Sigma_e$ w.r.t $\Sigma$ can be defined by the rotation matrix
\begin{equation}
R_e = R_b R^b_e,
\label{eq:Re_cpct}
\end{equation}

The  forward  kinematics  problem  consists  of  determining the operational coordinates $\chi_e = [x_e \ y_e \ z_e \ \psi_e \ \theta_e \ \phi_e]^T$ $\in \mathbb{R}^{6}$, as a function of the quadrotor/joint space coordinates $q = [x\ y\ z\ \psi\ \theta\ \phi\ \theta_1\ \theta_2]^T$ $\in \mathbb{R}^{8}$. For solving the forward kinematics, the inputs are $8$ variables, $q$, and the output are $6$ variables, $\chi_e$, obtained from $6$ algebraic equations. The end-effector position can be found from (\ref{eq:pe}). Euler angles of the end-effector $\Phi_e$ can be computed from the rotation matrix of $R_e$, from (\ref{eq:Re_cpct}), see Appendix \ref{app:forwardkinematics} for the coding of the forward kinematics.

In the reset of this section, the velocity kinematic analysis is discussed. 

The linear velocity of $\Sigma_e$ in the world-fixed frame,  $\dot{p}_e$, is obtained by the differentiation of (\ref{eq:pe}) as
\begin{equation}
\dot{p}_e = \dot{p}_b - Skew(R_b p^b_{eb}) \omega_b + R_b \dot{p}^b_{eb},
\label{eq:pde}
\end{equation}
where $Skew(.)$ is the $(3 \times 3)$ skew-symmetric matrix operator \cite{spong2006robot}, while $\omega_b$ is the angular velocity of the quadrotor expressed in $\Sigma$.
The angular velocity $\omega_e$ of $\Sigma_e$ is expressed as
\begin{equation}
\omega_e = \omega_b + R_b \omega^b_{eb},
\label{eq:we}
\end{equation}
where $\omega^b_{eb}$ is the angular velocity of the end-effector relative to $\Sigma_b$ and is expressed in $\Sigma_b$.

The generalized velocity of the end-effector with respect to $\Sigma_b$, $v^b_{eb} = [\dot{p}^{bT}_{eb} \ \omega^{bT}_{eb}]^T$ $\in \mathbb{R}^{6}$, can be expressed in terms of the joint velocities $\dot{\Theta}$ via the manipulator Jacobian, $J^b_{eb}$, \cite{Tsai} as
\begin{equation}
v^b_{eb} = J^b_{eb} \dot{\Theta},
\label{eq:vbeb}
\end{equation}
where $J^b_{eb}$ is given by
\begin{equation}
J^b_{eb}= \begin{bmatrix}
0 & l_2  C_{\theta_2}\\
 S_{\theta_1}(l_1 + l_2  C_{\theta_2}) & l_2  C_{\theta_1}  S_{\theta_2}\\
- C_{\theta_1}(l_1 + l_2  C_{\theta_2}) & l_2  S_{\theta_1}  S_{\theta_2}\\
1 & 0\\
0 & - S_{\theta_1} \\
0 &  C_{\theta_1}
\end{bmatrix}.
\label{eq:Jbeb}
\end{equation}

From (\ref{eq:pde}) and (\ref{eq:we}), the generalized end-effector velocity, $v_e = [\dot{p}^T_e \ \omega^T_e]^T$, can be expressed as
 
\begin{equation}
v_e = J_b v_b + J_{eb} \dot{\Theta},
\label{eq:ve}
\end{equation}
where $v_b = [\dot{p}^T_b \ \omega^T_b]^T$,
$J_b= \begin{bmatrix}
I_3 & -Skew(R_b p^b_{eb})\\
O_3 & I_3
\end{bmatrix},$
$J_{eb}= \begin{bmatrix}
R_b & O_3\\
O_3 & R_b
\end{bmatrix} J^b_{eb}$,\\ 
where $I_m$ and $O_m$ denote $(m \times m)$ identity and $(m \times m)$ null matrices, respectively.
If the attitude of the vehicle is expressed in terms of yaw-pitch-roll angles, then (\ref{eq:ve}) will be
\begin{equation}
v_e = J_b Q_b \dot{\chi}_b + J_{eb} \dot{\Theta},
\label{eq:veph}
\end{equation}
with 
$\chi_b= \begin{bmatrix}
p_b \\
\Phi_b 
\end{bmatrix},$ 
$Q_b= \begin{bmatrix}
I_3 & O_3\\
O_3 & T_b
\end{bmatrix},$ 
where $T_b$ describes the transformation matrix between the angular velocity $\omega_b$ and the time derivative of Euler angles $\dot{\Phi}_b$, and it is given as
\begin{equation}
T_b(\Phi_b)= \begin{bmatrix}
0 &  -S_{\psi} & C_{\psi} C_{\theta} \\
0 & C_{\psi} & S_{\psi} C_{\theta} \\
1 & 0 & -S_{\theta} \\
\end{bmatrix}.
\label{eq:Tb}
\end{equation}

Thus, the system Jacobian is
\begin{equation}
v_e = J \dot{q},
\label{eq:veph_tot}
\end{equation}
where $\dot{q} = [\dot{\chi}_b^T \ \dot{\Theta}^T]^T$, and $J = [J_b Q_b \ J_{eb}]$ and it is given as
\begin{equation}
J= \begin{bmatrix}

1 &  0 &  0 &  J_{14} & J_{15}  &  J_{16} &  J_{17} &  J_{18} \\
0 &  1 &  0 &  J_{24} &  J_{25} &  J_{26} &  J_{27}&  J_{28}\\
0 &  0 &  1 & J_{34} & J_{35} & 0 & -C_{(\phi + \theta_1)} C_{\theta} (l_1 + l_2 C_{\theta_2}) & J_{38}\\                                                                                                                             
0 &  0 &  0 &                                                                                                                                                                                                                                                                                                                                          C_{\psi} C_{\theta} &                                                                                                                                                                            -S_{\psi} &                                                                                                                                                                                              0 &                                                                                                                                                        C_{\psi} C_{\theta} &  J_{48} \\
0 &  0 &  0 &                                                                                                                                                                                                                                                                                                                                            C_{\theta} S_{\psi} &                                                                                                                                                                             C_{\psi} &                                                                                                                                                                                              0 &                                                                                                                                                        C_{\theta} S_{\psi} &   J_{58}\\
0 &  0 &  0 &                                                                                                                                                                                                                                                                                                                                                   -S_{\theta} &                                                                                                                                                                                   0 &                                                                                                                                                                                              1 &                                                                                                                                                               -S_{\theta} &                                                                                                                                              C_{(\phi + \theta_1)} C_{\theta} \\

\end{bmatrix},
\label{eq:jacob}
\end{equation}
with 

\begin{gather}
J_{14} = S_{\theta} ((C_{\psi} S_{\phi} - C_{\phi} S_{\psi} S_{\theta}) (l_0 + l_1 S_{\theta_1} + l_2 C_{\theta_2} S_{\theta_1}) - C_{\theta_1} (l_1 + l_2 C_{\theta_2}) (C_{\psi} C_{\phi} + S_{\psi} S_{\phi} S_{\theta}) \nonumber \\ + l_2 C_{\theta} S_{\psi} S_{\theta_2}) - C_{\theta} S_{\psi} (l_2 S_{\theta} S_{\theta_2} + C_{\phi} C_{\theta} (l_0 + l_1 S_{\theta_1} + l_2 C_{\theta_2} S_{\theta_1}) + C_{\theta} C_{\theta_1} S_{\phi} (l_1 + l_2 C_{\theta_2})),
\end{gather}

\begin{equation}
J_{15} = -C_{\psi} (l_2 S_{\theta} S_{\theta_2} + C_{\phi} C_{\theta} (l_0 + l_1 S_{\theta_1} + l_2 C_{\theta_2} S_{\theta_1}) + C_{\theta} C_{\theta_1} S_{\phi} (l_1 + l_2 C_{\theta_2})),
\end{equation}

\begin{equation}
J_{16} = C_{\theta_1} (l_1 + l_2 C_{\theta_2}) (C_{\psi} C_{\phi} + S_{\psi} S_{\phi} S_{\theta}) - (C_{\psi} S_{\phi} - C_{\phi} S_{\psi} S_{\theta}) (l_0 + l_1 S_{\theta_1} + l_2 C_{\theta_2} S_{\theta_1}) - l_2 C_{\theta} S_{\psi} S_{\theta_2},
\end{equation}

\begin{equation}
J_{17} = - (l_1 S_{\theta_1} + l_2 C_{\theta_2} S_{\theta_1}) (C_{\phi} S_{\psi} - C_{\psi} S_{\phi} S_{\theta}) - (l_1 C_{\theta_1} + l_2 C_{\theta_1} C_{\theta_2}) (S_{\psi} S_{\phi} + C_{\psi} C_{\phi} S_{\theta}),
\end{equation}

\begin{equation}
J_{18} = l_2 C_{\psi} C_{\theta} C_{\theta_2} - l_2 C_{\theta_1} S_{\theta_2} (C_{\phi} S_{\psi} - C_{\psi} S_{\phi} S_{\theta}) + l_2 S_{\theta_1} S_{\theta_2} (S_{\psi} S_{\phi} + C_{\psi} C_{\phi} S_{\theta}),
\end{equation}

\begin{gather}
J_{24} = C_{\psi} C_{\theta} (l_2 S_{\theta} S_{\theta_2} + C_{\phi} C_{\theta} (l_0 + l_1 S_{\theta_1} + l_2 C_{\theta_2} S_{\theta_1}) + C_{\theta} C_{\theta_1} S_{\phi}  (l_1 + l_2 C_{\theta_2})) - S_{\theta} (C_{\theta_1} \nonumber \\ (l_1 + l_2 C_{\theta_2}) (C_{\phi} S_{\psi} - C_{\psi} S_{\phi} S_{\theta}) - (S_{\psi} S_{\phi} + C_{\psi} C_{\phi} S_{\theta})  (l_0 + l_1 S_{\theta_1} + l_2 C_{\theta_2} S_{\theta_1}) + l_2 C_{\psi} C_{\theta} S_{\theta_2}),
\end{gather}

\begin{equation}
J_{25} = -S_{\psi} (l_2 S_{\theta} S_{\theta_2} + C_{\phi} C_{\theta} (l_0 + l_1 S_{\theta_1} + l_2 C_{\theta_2} S_{\theta_1}) + C_{\theta} C_{\theta_1} S_{\phi} (l_1 + l_2 C_{\theta_2})),
\end{equation}

\begin{equation}
J_{26} = C_{\theta_1} (l_1 + l_2 C_{\theta_2}) (C_{\phi} S_{\psi} - C_{\psi} S_{\phi} S_{\theta}) - (S_{\psi} S_{\phi} + C_{\psi} C_{\phi} S_{\theta}) (l_0 + l_1 S_{\theta_1} + l_2 C_{\theta_2} S_{\theta_1}) + l_2 C_{\psi} C_{\theta} S_{\theta_2},
\end{equation}

\begin{equation}
J_{27} =  (l_1 S_{\theta_1} + l_2 C_{\theta_2} S_{\theta_1}) (C_{\psi} C_{\phi} + S_{\psi} S_{\phi} S_{\theta}) + (l_1 C_{\theta_1} + l_2 C_{\theta_1} C_{\theta_2}) (C_{\psi} S_{\phi} - C_{\phi} S_{\psi} S_{\theta}),
\end{equation}

\begin{equation}
J_{28} = l_2 C_{\theta} C_{\theta_2} S_{\psi} + l_2 C_{\theta_1} S_{\theta_2} (C_{\psi} C_{\phi} + S_{\psi} S_{\phi} S_{\theta}) - l_2 S_{\theta_1} S_{\theta_2} (C_{\psi} S_{\phi} - C_{\phi} S_{\psi} S_{\theta}), 
\end{equation}

\begin{equation}
J_{34}= -C_{\theta} ((l_2 C_{(\phi + \theta_1 - \theta_2)})/2 + l_1 C_{(\phi + \theta_1)} - l_0 S_{\phi} + (l_2 C_{(\phi + \theta_1 + \theta_2)})/2), 
\end{equation}

\begin{equation}
J_{35} =  l_0 C_{\phi} S_{\theta} - l_2 C_{\theta} S_{\theta_2} + l_1 C_{\phi} S_{\theta} S_{\theta_1} + l_1 C_{\theta_1} S_{\phi} S_{\theta} + l_2 C_{\phi} C_{\theta_2} S_{\theta} S_{\theta_1} + l_2 C_{\theta_1} C_{\theta_2} S_{\phi} S_{\theta}, 
\end{equation}

\begin{equation}
J_{38} = l_2 C_{\phi} C_{\theta} S_{\theta_1} S_{\theta_2} - l_2 C_{\theta_2} S_{\theta} + l_2 C_{\theta} C_{\theta_1} S_{\phi} S_{\theta_2},
\end{equation}

\begin{equation}
J_{48} = C_{\theta_1} (S_{\psi} S_{\phi} + C_{\psi} C_{\phi} S_{\theta}) + S_{\theta_1} (C_{\phi} S_{\psi} - C_{\psi} S_{\phi} S_{\theta}),
\end{equation}

\begin{equation}
J_{58} = - C_{\theta_1} (C_{\psi} S_{\phi} - C_{\phi} S_{\psi} S_{\theta}) - S_{\theta_1} (C_{\psi} C_{\phi} + S_{\psi} S_{\phi} S_{\theta}), 
\end{equation}

See Appendix \ref{app:system_Jacobian} for the programming of the system Jacobian.

Since the quadrotor is an under-actuated system, i.e., only $4$ independent control inputs are available for the 6-DOF system, the position and the yaw angle are usually the controlled variables. Hence, it is worth to define the independent coordinate, $\zeta=[x \ y \ z \ \psi \ \theta_1 \ \theta_2]^T$, as the controlled variables, and the dependent coordinates, $\sigma_b= [\theta \ \phi]^T$, as the intermediate variables. Hence, it is worth rewriting the vector $\chi_b$ as
$
\chi_b= \begin{bmatrix}
\eta_b \\
\sigma_b 
\end{bmatrix}, 
$
$
\eta_b= \begin{bmatrix}
p_b \\
\psi 
\end{bmatrix}, 
$  
$
\sigma_b= \begin{bmatrix}
\theta \\
\phi 
\end{bmatrix}. 
$ 

Thus, the differential kinematics (\ref{eq:veph}) will be

\begin{equation}
\begin{aligned}
v_e &= J_{\eta} \dot{\eta}_b + J_{\sigma} \dot{\sigma}_b + J_{eb} \dot{\Theta}\\
&=J_{\zeta} \dot{\zeta} + J_{\sigma} \dot{\sigma}_b,
\end{aligned} 
\label{eq:vediv}
\end{equation}
where $\zeta = [\eta_b^T \ \Theta^T]^T$ is the vector of the controlled variables, $J_{\eta}$ is composed by the first 4 columns of $J_b Q_b$, $J_{\sigma}$ is composed by the last 2 columns of $J_b Q_b$, and $J_{\zeta} = [J_{\eta} \ J_{eb}]$.

If the end-effector orientation is expressed via a triple of Euler angles, $ZYX$, $\Phi_e$, the differential kinematics (\ref{eq:vediv}) can be rewritten in terms of the vector $\dot{\chi}_e = [\dot{p}^T_e, \dot{\Phi}_e^T]^T$ as follows
\begin{equation}
\begin{aligned}
\dot{\chi}_e & = Q_e^{-1}(\Phi_e) v_e\\
& = Q_e^{-1}(\Phi_e) [J_{\zeta} \dot{\zeta} + J_{\sigma} \dot{\sigma}_b],
\end{aligned} 
\label{eq:xedot}
\end{equation}
where $Q_e$ is the same as $Q_b$ but it is a function of $\Phi_e$ instead of $\Phi_b$.

\section{Dynamic Model} \label{se:dyn}

For the manipulator dynamics, Recursive Newton Euler method \cite{Tsai} is used to derive the equations of motion. Since the quadrotor is considered to be the base of the manipulator, the initial linear and angular velocities and accelerations, used in Newton Euler algorithm, are that of the quadrotor expressed in body frame. Applying the Newton Euler algorithm to the manipulator considering that the link (with length $l_{0}$) that is fixed to the quadrotor is the base link, manipulator's equations of motion can be obtained, in addition to, the forces and moments, from manipulator, that affect the quadrotor.

For the system structure, we assume:
\begin{assumption}
	The quadrotor body is rigid and symmetric. The manipulator links are rigid.
\end{assumption}
For each link $i$, ($i$ = $0, 1, 2$), let us define 
\begin{itemize}
	\item $\omega_{i}^{i}$ is the angular velocity of frame $i$ expressed in frame $i$.
	\item $\dot{\omega}_{i}^{i}$ is the angular acceleration of frame $i$ expressed in frame $i$.
	\item $v_{i}^{i}$ is the linear velocity of the origin of frame $i$ expressed in frame $i$.
	\item $\dot{v}^{i}_{c_i}$ is the linear acceleration of the center of mass of link $i$ expressed in frame $i$.
	\item $\dot{v}_{i}^{i}$ is the linear acceleration of the origin of frame $i$ expressed in frame $i$.
	\item $r_{i}^{i}$ is the position vector from the origin of frame $i-1$ to the origin of link $i$.
	\item $r_{c_i}^{i}$ is the position vector from the origin of frame $i-1$ to the center of mass of link $i$.
	\item $g^{}$ is the vector of gravity expressed in $\Sigma$.
	\item $z_{i-1}^{i-1}$ is a unit vector pointing along the $i^{th}$ joint axis and expressed in the $(i-1)^{th}$ link coordinate system.
	\item $R_{i}^{i-1}$ is the rotation matrix from frame $i$ to frame $i-1$.
	\item  $I_{i}^{i} = I_i \begin{bmatrix} 0 & 0 & 0\\ 0 & 1 & 0\\ 0 & 0 & 1
	\end{bmatrix}$ is the inertia matrix of link $i$ about its center of mass coordinate frame.
	\item $f_{i,i-1}^{i}$ / $n_{i,i-1}^{i}$ are the resulting force/moment exerted on link $i$ by link $i-1$ at point $O_{i-1}$. 
	\item $R_{}^{i}$ is the rotation matrix from frame $\Sigma$ to frame $i$.
\end{itemize}

Applying a payload of mass equal to $m_{p}$ at the end-effector will change link 2's parameters such as mass moments of inertia, total mass, and center of gravity as

\begin{equation}
\grave{I}_{2}^{2} = I_{2}^{2} + m_{2} (\grave {d}_{CG_2} - d_{CG_2})^{2}+ m_{p} (l_{2} - \grave {d}_{CG_2})^{2},
\label{I22new}
\end{equation}
\begin{equation}
\grave{d}_{CG_2} = \frac{m_{2} d_{CG_2} + m_{p} l_{2}}{m_{2} + m_{p}},
\label{CGnew}
\end{equation}
\begin{equation}
\grave{m}_{2} = m_{2} + m_{p},
\label{m2new}
\end{equation}

where $CG_{2}$ is the point of center of gravity of link 2, ${d}_{CG_2} = l_2/2$, and $(\grave{.})$ refers to the value of the parameter after adding the payload. Changing the point of center of gravity of link 2 will make the $r_{c_2}^{2}$ to be
\begin{equation}
r_{c_2}^{2} = [-(l_2 - \grave{d_{CG_2}}) \  0 \ 0]^{T}.
\label{rc22new}
\end{equation}

For the link $0$, one calculates the following:

\begin{equation}
\label{Wb0}
\omega_{b}^{0} = R_{b}^{0} \omega_{b}^b,
\end{equation}
\begin{equation}
\label{vb0}
v_{b}^{0} = R_{b}^{0} v_{b}^b,
\end{equation}
\begin{equation}
\label{Wdb0}
\dot{\omega}_{b}^{0} = R_{b}^{0} \dot{\omega}_{b}^b,
\end{equation}
\begin{equation}
\label{vdb0}
\dot{v}_{b}^{0} = R_{b}^{0} \dot{v}_{b}^b,
\end{equation}
\begin{equation}
r_{b_0}^{0} = [0 \ l_0 \ 0]^{T},
\label{rboo}
\end{equation}
\begin{equation}
r_{b_0}^{} = R_{0}^{} r_{b_0}^{0},
\label{rboi}
\end{equation}
\begin{equation}
\label{W00}
\omega_{0}^{0} = \omega_{b}^{0},
\end{equation}
\begin{equation}
\label{Wd00}
\dot{\omega}_{0}^{0} = \dot{\omega}_{b}^{0},
\end{equation}
\begin{equation}
\label{v00}
v_{0}^{0} = v_{b}^0 + \omega_{b}^{0} \times r_{b_0}^{0},
\end{equation}
\begin{equation}
\label{vd00}
\dot{v}_{0}^{0} = \dot{v}_{b}^0 + \dot{\omega}_{b}^{0} \times r_{b_0}^{0} + \omega_{b}^{0} \times (\omega_{b}^{0} \times r_{b_0}^{0}),
\end{equation}
\begin{equation}
\label{Wd0i}
\dot{\omega}_{0}^{} = R_{0}^{} \dot{\omega}_{0}^0,
\end{equation}
\begin{equation}
\label{W0i}
\omega_{0}^{} = R_{0}^{} \omega_{0}^0,
\end{equation}
\begin{equation}
\label{vd0i}
\dot{v}_{0}^{} = \dot{v}_{b}^{} + \dot{\omega}_{0}^{} \times r_{b_0}^{} + \omega_{0}^{} \times (\omega_{0}^{} \times r_{b_0}^{}).
\end{equation}

For link 1, one can calculate the following:
\begin{equation}
\label{w11}
\omega_{1}^{1} = R_{}^{1} \omega_{0}^{} + R_{0}^{1}  \dot{\theta}_{1} z_{0}^{0},
\end{equation}
\begin{equation}
\label{wd11}
\dot{\omega}_{1}^{1} =  R_{}^{1} \dot{\omega}_{0}^{} + R_{0}^{1} \ddot{\theta}_{1} z_{0}^{0} +  R_{}^{1} \omega_{0}^{} \times ( R_{0}^{1} \dot{\theta}_{1} z_{0}^{0}),
\end{equation}

\begin{equation}
\label{vd11}
\dot{v}_{1}^{1} = R_{}^{1} \dot{v}_{0}^{} + \dot{\omega}_{1}^{1} \times r_{1}^{1} + \omega_{1}^{1} \times (\omega_{1}^{1} \times r_{1}^{1}),
\end{equation}
\begin{equation}
\label{vcd11}
\dot{v}_{c_1}^{1} =\dot{v}_{1}^{1} + \dot{\omega}_{1}^{1} \times r_{c_1}^{1} + \omega_{1}^{1} \times (\omega_{1}^{1} \times r_{c_1}^{1}).
\end{equation}

For link 2, one can calculate the following:
\begin{equation}
\label{w22}
\omega_{2}^{2} = R_{1}^{2} (\omega_{1}^{1} + \dot{\theta}_{2} z_{1}^{1}),
\end{equation}
\begin{equation}
\label{wd22}
\dot{\omega}_{2}^{2} =  R_{1}^{2} (\dot{\omega}_{1}^{1} + \ddot{\theta}_{2} z_{1}^{1} +  \omega_{1}^{1} \times (\dot{\theta}_{2} z_{1}^{1})),
\end{equation}
\begin{equation}
\label{vd22}
\dot{v}_{2}^{2} = R_{1}^{2} (\dot{v}_{1}^{1} + \dot{\omega}_{2}^{2} \times r_{2}^{2} + \omega_{2}^{2} \times (\omega_{2}^{2} \times r_{2}^{2})),
\end{equation}
\begin{equation}
\label{vcd22}
\dot{v}_{c_2}^{2} =\dot{v}_{2}^{2} + \dot{\omega}_{2}^{2} \times r_{c_2}^{2} + \omega_{2}^{2} \times (\omega_{2}^{2} \times r_{c_2}^{2}).
\end{equation}

The inertial forces and moments acting on link $i$ are given by
\begin{equation}
\label{Fii}
F_{i}^{i} = - m_{i} \dot{v}_{c_i}^{i},
\end{equation}
\begin{equation}
\label{Nii}
N_{i}^{i} = - I_{i}^{i} \dot{\omega}_{i}^{i} - \omega_{i}^{i} \times I_{i}^{i} \omega_{i}^{i}.
\end{equation}

The total forces and moments acting on link $i$ are given by
\begin{equation}
\label{fi_i-1}
f_{i,i-1}^{i} = f_{i+1,i}^{i} - m_{i} g^{i} - F_{i}^{i},
\end{equation}
\begin{equation}
\label{ni_i-1}
n_{i,i-1}^{i} = n_{i+1,i}^{i} +(f_{i}^{i} + r_{c_i}^{i}) \times f_{i,i-1}^{i} - r_{c_i}^{i} \times f_{i+1,i}^{i} - N_{i}^{i},
\end{equation}
\begin{equation}
\label{fi-1_i-1}
f_{i,i-1}^{i-1} = R_{i}^{i-1} f_{i,i-1}^{i},
\end{equation}
\begin{equation}
\label{ni-1_i-1}
n_{i,i-1}^{i-1} = R_{i}^{i-1} n_{i,i-1}^{i},
\end{equation}
where
\begin{equation}
z_{0}^{0} = [0 \ 0 \ 1]^T,
\end{equation}
\begin{equation}
r_{1}^{1} = [a_1 \ d_1 \sin(\alpha_1) \ d_1 \cos(\alpha_1)]^T,
\end{equation}
\begin{equation}
r_{c_1}^{1} = [-l_{1}/2 \ 0 \ 0]^T,
\end{equation}
\begin{equation}
z_{1}^{1} = [0 \ 0 \ 1]^T,
\end{equation}
\begin{equation}
r_{2}^{2} = [a_2 \ d_2 \sin(\alpha_2) \ d_2 \cos(\alpha_2)]^T.
\end{equation}

The gravity vector expressed in frame $i$ is
\begin{equation}
\label{g2}
g^{i} = R_{}^{i} g^{},
\end{equation}
where
\begin{equation}
\label{RI1}
R_{}^{1} = R_{0}^{1} R_{b}^{0} R_{}^{b},
\end{equation}
\begin{equation}
\label{RI2}
R_{}^{2} = R_{1}^{2} R_{}^{1},
\end{equation}
and
\begin{equation}
g^{} = [0 \ 0 \ -g_r]^T,
\end{equation}
where $g_r$ is the gravity acceleration.

The torques acting on joints 1 and 2 are finally given by
\begin{equation}
\label{Tm1details}
\tau_{m_1} = (n_{1,0}^{0})^{T}  z_{0}^{0} + b_{1} \dot{\theta}_{1},
\end{equation}
\begin{equation}
\label{Tm2details}
\tau_{m_2} = (n_{2,1}^{1})^{T}  z_{1}^{1} + b_{2} \dot{\theta}_{2},
\end{equation}
where $b_1$ and $b_2$ are  the friction coefficients.

The interaction forces and moments of the manipulator acting on the quadrotor expressed in $\Sigma_b$, $F_{m,q}^{b}$ and $M_{m,q}^{b}$ are given as
\begin{equation}
F_{m,q}^{b} =   R_{0}^{b} (- f_{1,0}^{0}),
\label{interact_f}
\end{equation}
\begin{equation}
M_{m,q}^{b} =   R_{0}^{b} (-n_{1,0}^{0} + r_{b_0}^{0} \times (- f_{1,0}^{0})).
\label{interact_M}
\end{equation}

The interaction forces expressed in the inertial frame can be obtained by
\begin{equation}
\label{FmqI}
F_{m,q}^{} = R_{b}^{} F_{m,q}^{b}.
\end{equation}

The equations of motion of the manipulator can be reformulated as
\begin{equation}
\label{arm1}
M_{1}(q)\ddot{\theta_1} = \tau_{m_1} + N_1(q,\dot{q},\ddot{q}),
\end{equation}
\begin{equation}
\label{arm2}
M_{2}(q)\ddot{\theta_2} = \tau_{m_2} + N_2(q,\dot{q},\ddot{q}),
\end{equation}
where $M_1(q)$, $M_2(q)$, $N_1(q,\dot{q},\ddot{q})$, and $N_2(q,\dot{q},\ddot{q})$ are  nonlinear terms, and they are functions of the system states.

The Newton Euler method is used to find the equations of motion of the quadrotor after adding the forces/moments from the manipulator. They are given by (\ref{Xdd}-\ref{epdd}).
\begin{equation}
\label{Xdd}
m\ddot{x} = T(C_{\psi} S_{\theta} C_{\phi} + S_{\psi} S_{\phi}) + F_{m,q_x}
\end{equation}
\begin{equation}
\label{Ydd}
m\ddot{y} = T(S_{\psi} S_{\theta} C_{\phi} - C_{\psi} S_{\phi}) + F_{m,q_y}
\end{equation}
\begin{equation}
\label{Zdd}
m\ddot{z} = -m g_r + T C_{\theta} C_{\phi} + F_{m,q_z}
\end{equation}
\begin{equation}
\label{Phdd}
I_{x}\ddot{\phi} = \dot{\theta}\dot{\phi}(I_{y}-I_{z}) - I_{r}\dot{\theta}\overline{\Omega} + \tau_{a_1} + M_{m,q_\phi}^b
\end{equation}
\begin{equation}
\label{thdd}
I_{y}\ddot{\theta} = \dot{\psi}\dot{\phi}(I_{z}-I_{x}) + I_{r}\dot{\phi}\overline{\Omega} + \tau_{a_2} + M_{m,q_\theta}^b
\end{equation}
\begin{equation}
\label{epdd}
I_{z}\ddot{\psi} = \dot{\theta}\dot{\phi}(I_{x}-I_{y}) + \tau_{a_3} + M_{m,q_\psi}^b
\end{equation}
where $F_{m,q_x}$, $F_{m,q_y}$, and $F_{m,q_z}$ are the interaction forces resulted from the manipulator and affected the quadrotor in $x$, $y$, and $z$ directions expressed in the inertial frame and $M_{m,q_\phi}^b$, $M_{m,q_\theta}^b$, and $M_{m,q_\psi}^b$ are the interaction moments from the manipulator to the quadrotor around $x_b$, $y_b$, and $z_b$ directions.

The variables in (\ref{Xdd}-\ref{epdd}) are defined as follows:
$m$ is the mass of the quadrotor. Each rotor $j$ has angular velocity $\Omega_j$ and it produces thrust force $F_j$ and drag moment $M_j$ which are given by
\begin{equation}
F_j = K_{f_j} \Omega_j^2,
\label{eq:thrust}
\end{equation}
\begin{equation}
M_j = K_{m_j} \Omega_j^2,
\label{eq:dragmoment}
\end{equation}
where $K_{f_j}$ and $K_{m_j}$ are the thrust and drag coefficients, respectively.

$T$ is the total thrust applied to the quadrotor from all four rotors, and it is given by
\begin{equation}
T = \sum\limits_{j=1}^{4} F_j.
\label{thrust_sum}
\end{equation}
$\tau_{a_1}$, $\tau_{a_2}$, and $\tau_{a_3}$  are the three input moments about the three body axes, $x_b$, $y_b$, $z_b$, and they are given as
\begin{equation}
\tau_{a_1} = d_q(F_4 - F_2),
\label{Ta1}
\end{equation}
\begin{equation}
\tau_{a_2} = d_q(F_3 - F_1),
\label{Ta2}
\end{equation}
\begin{equation}
\tau_{a_3} = -M_1 + M_2 - M_3 + M_4.
\label{Ta3}
\end{equation}
$d_q$ is the distance between the quadrotor center of mass and rotor rotational axis. 
$\overline{\Omega}$ is given by
\begin{equation}
\overline{\Omega} = \Omega_1 - \Omega_2 + \Omega_3 - \Omega_4.
\label{omega_bar}
\end{equation}
$I_r$ is the rotor inertia. $I_f$ is the inertia matrix of the vehicle around its body-frame , and it is given as
\begin{equation}
I_f = I_i \begin{bmatrix} I_x & 0 & 0\\ 0 & I_y & 0\\ 0 & 0 & I_z
\end{bmatrix}. 
\end{equation}
The dynamical model of the quadrotor-manipulator system can be rewritten in a matrix form as follows
\begin{gather}
M(q) \ddot{q} + C(q,\dot{q}) \dot{q} + G(q) + d_{ex} = \tau, \nonumber \\
 \tau = B_m u,
\label{eq:dyn_gen}
\end{gather}
where $M$ $\in \mathbb{R}^{8 \times 8}$ represents the symmetric and positive definite inertia matrix of the combined system, $C$ $\in \mathbb{R}^{8 \times 8}$ is the matrix of Coriolis and centrifugal terms, $G$ $\in \mathbb{R}^{8}$ is the vector of gravity terms, $d_{ex}$ $\in \mathbb{R}^{8}$ is the vector of external disturbances, $\tau$ $\in \mathbb{R}^{8}$ is the vector of the input generalized forces, $u = [F_1 \ F_2 \ F_3 \ F_4 \ \tau_{m_1} \ \tau_{m_2}]^T$ $\in \mathbb{R}^{6}$ is vector of the actuator inputs, and $B_m= H_m(q) \ N_m(K_{f_j},K_{m_j},d_q)$ $\in \mathbb{R}^{8 \times 6}$ is the input matrix which is used to generate the body forces and moments from the actuator inputs. The control matrix,
$N_m$ $\in \mathbb{R}^{8 \times 6}$, is given by
\begin{equation}
N_m= \begin{bmatrix}
0 & 0 & 0 & 0 & 0 & 0 \\
0 & 0 & 0 & 0 & 0 & 0 \\
1 & 1 & 1 & 1 & 0 & 0 \\
\gamma_1 & -\gamma_2 & \gamma_3 & -\gamma_4 & 0 & 0 \\
-d_q & 0 & d_q & 0 & 0 & 0 \\
0 & -d_q & 0 & d_q & 0 & 0 \\
0 & 0 & 0 & 0 & K_{{\tau}_1} & 0 \\
0 & 0 & 0 & 0 & 0 & K_{{\tau}_2} 
\end{bmatrix},
\label{eq:N}
\end{equation}

where $\gamma_j=K_{m_j}/K_{f_j}$. $K_{{\tau}_1}$ and $K_{{\tau}_2}$ are coefficients that are used to emulate the occurring of faults in the motor's constants of joints $1$ and $2$, respectively. $H_m$ $ \in \mathbb{R}^{8 \times 8}$ is a matrix that transforms the body input forces to be expressed in $\Sigma$ and is given by
\begin{equation}
H_m= \begin{bmatrix}
R_b & O_3 & O_2 \\
O_3 & T_b^T R_b & O_2 \\
O_{2 \times 3} & O_{2 \times 3} & I_2 
\end{bmatrix}.
\label{eq:H}
\end{equation}

From the equations of the translation dynamics part of (\ref{Xdd}-\ref{Zdd}), one can extract the expressions of the second order nonholonomic constraints as

\begin{equation}
\sin(\phi) - \frac{\ddot{x}_f S_{\psi} - \ddot{y}_f C_{\psi}}{\sqrt{\ddot{x}_f^{2} + \ddot{y}_f^{2} + \ddot{z}_f^{2}}} =0,
\label{sin_ph}
\end{equation}

\begin{equation}
\tan(\theta) -\frac{\ddot{x}_f C_{\psi} + \ddot{y}_f S_{\psi}}{\ddot{z}_f} =0,
\label{tan_th}
\end{equation}
where $\ddot{x}_f$ = $\ddot{x} - \frac{F_{m,q_x}}{m}$, $\ddot{y}_f$ = $\ddot{y} - \frac{F_{m,q_y}}{m}$, and $\ddot{z}_f$ = $\ddot{z} + g_r - \frac{F_{m,q_z}}{m}$.

It is to be noted that the force terms in the above equations are also function of the system states and their derivatives. Equations (\ref{sin_ph}) and (\ref{tan_th}) can be solved for the desired trajectories of $\phi$ and $\theta$ through substituting by the desired trajectories of the other variables. These nonholonomic constrains will be utilized later to solve both the inverse kinematics and control problems.

The next step is to build the experimental system to test the system feasibility.

%


\chapter{\uppercase{System Construction and Experiments}} 
\label{ch:Implementation}


\ifpdf
    \graphicspath{{3_Implementation/figures/PNG/}{3_Implementation/figures/PDF/}{3_Implementation/figures/}}
\else
    \graphicspath{{3_Implementation/figures/EPS/}{3_Implementation/figures/}}
\fi


In this chapter, the whole system is built and its parameters are identified. Furthermore, experimental setup with a solution to the problem of the state estimation of the quadrotor manipulation system is introduced taking into consideration the position of the manipulator with respect to the sensors. This solution is based on a combination of several sensors and data fusion. Quadrotor/joint space control is designed, with a formulation which enables the system to work in either teleoperation or autonomous mode, based on PID with gravity compensation technique. After that, this controller is implemented in both simulation and real time environments. An experimental test is implemented to demonstrate the feasibility and the effectiveness of the proposed system in the presence of holding and releasing a payload.
 
\section{Hardware}
\begin{figure}
	\centering
	\includegraphics[width=1\textwidth]{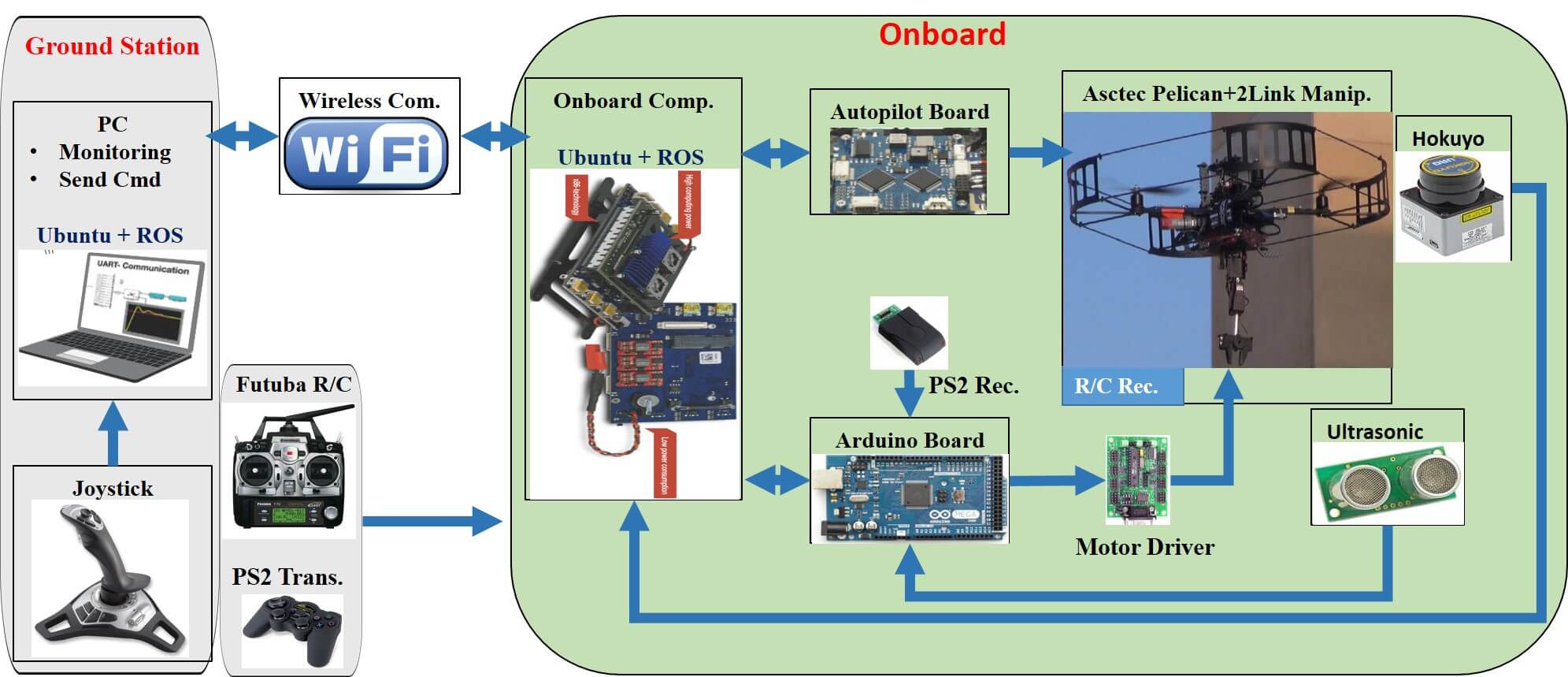}
	\caption{Aerial manipulation functional block diagram}
	\label{sys-conn}
\end{figure}

Fig. \ref{sys-conn} shows the proposed experimental implementation of the whole connected system with the user interface and measurement system.

The quadrotor is equipped with $4$ rotors with $10$ inch diameter, which allow to carry an additional weight of about $650$ g. Depending on the battery size and payload, flight times between $10$ and $20$ minutes can be achieved. In addition, the quadrotor has a Flight Control Unit (FCU) "AscTec Autopilot" as well as a suitable structure enabling one to easily mount different payloads like computer boards, onboard sensors,  and the robotic arm with its avionics. The FCU has an Inertial Measurement Unit (IMU) as well as two $32$-Bit $60$ MHz ARM-7 microcontrollers used for data fusion and flight control. One of these microcontrollers, the Low Level Processor (LLP), is responsible for the hardware communication, an emergency controller, and IMU sensor data fusion. An attitude and GPS-based position controller is implemented also on this processor. The other microcontroller, High Level Processor (HLP), is dedicated for custom code. All relevant and fused IMU data is provided at an update rate of $1$ kHz via a high speed serial interface. In particular, this comprises body accelerations, body angular velocities, magnetic compass, height measurement by an air pressure sensor, and the estimated attitude of the vehicle. For the onboard expensive computations, a $1.6$ GHz Intel Atom-based embedded computer is used. This computer is equipped with 1 GB RAM, a Micro SD card slot for the operating system, a 802.11n based miniPCI Express WiFi card. See Appendix \ref{app:experimentalsystem-quad} for more technical details.

State estimation is one of the key issue to implement quadrotor manipulation system. The available onboard sensor, IMU can only provide the quadrotor's orientation, angular rates, and linear Accelerations. However, one can not depend on the accelerations to get linear position due to large integration drift for long term motion. Thus, there is a need to an external positioning system which may be GPS, 3D motion capture system, onboard camera, or range sensors (laser or sonar). However, GPS is not a reliable service (accuracy 3 $\textendash$ 15 m) as its availability can be limited by urban canyons and is completely unavailable in indoor environments. Onboard camera is tried but the available one (Web CAM) has low resolutions and gives bad results. As a result, finally, the range sensors are used. 

\begin{figure}
	\centering
	\includegraphics[width=0.8\columnwidth]{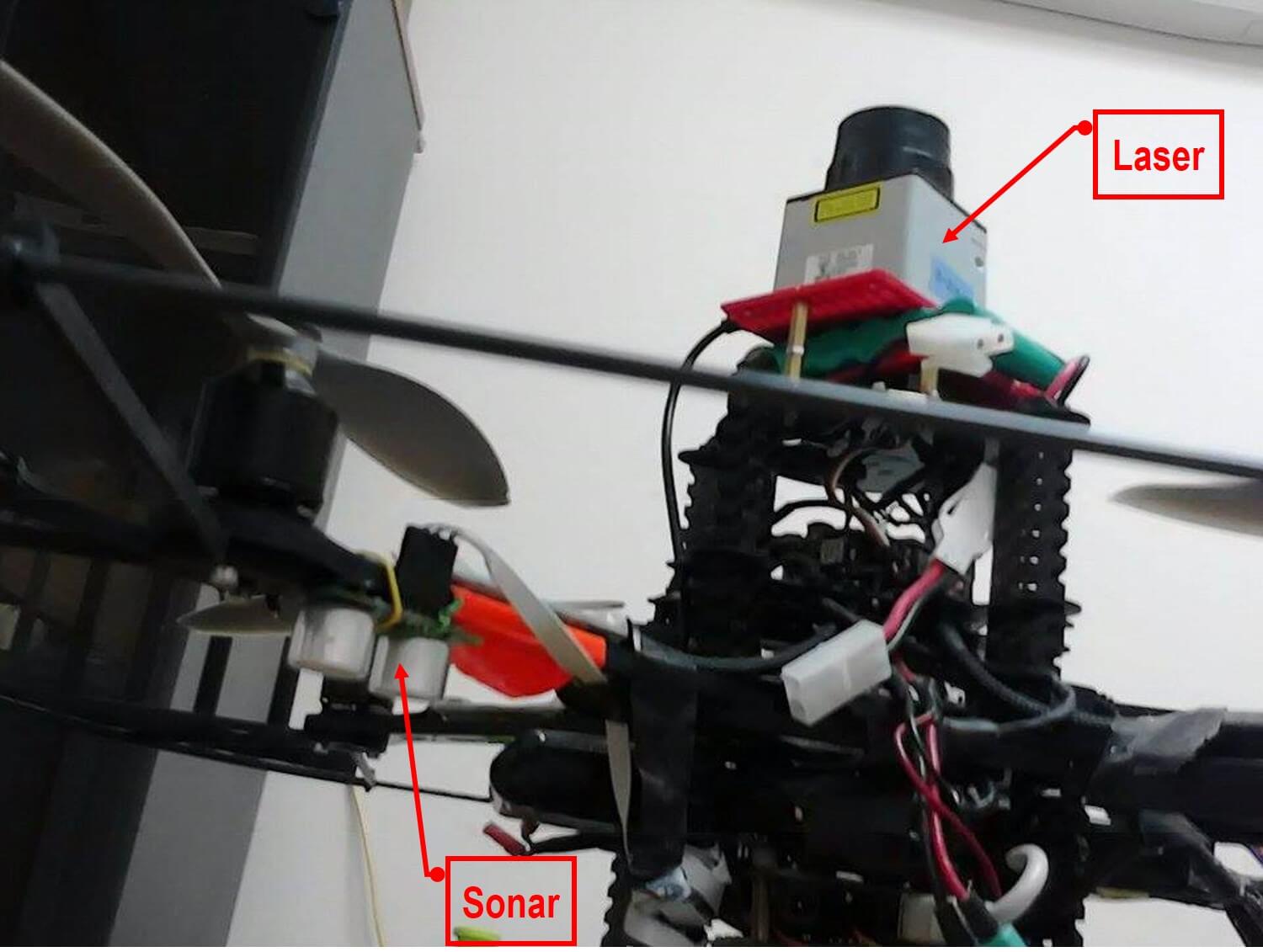}
	\caption{Close sight on position sensors connections}
	\label{fig:hokuyo_ulrasonic}
\end{figure}

For measuring the horizontal position, $x$ and $y$, an Hokuyo URG-04LX Laser Range Finder is used \cite{hokuyo}, see Fig. \ref{fig:hokuyo_ulrasonic}. It is a 2D laser range finder with a range of $4$ m and a field of view of $240$ degrees. It has an update rate of $30$ Hz with resolution of 1 mm. It has weight of 160 g. It is connected to the onboard computer through USB connection such that the high computation processing of laser data can be carried out onboard. This sensor is placed on the top center of the quadrotor. See Appendix \ref{app:experimentalsystem-laser} for more technical details.

The vertical position, $z$, is measured using Ultrasonic Ranging module SRF04 \cite {sonar_ref}, see Fig. \ref{fig:hokuyo_ulrasonic}. It is a 1D sonar range finder with a range of $3$ cm to $3$ m and a resolution of $3$ cm. It has an update rate of $40$ Hz. It is mounted downward under one of the rotors such that it is not affected by the movement of the manipulator, so one can avoid the sensor misreadings. See Appendix \ref{app:experimentalsystem-sonar} for more technical details. The data from this sensor is acquired and processed by using an Arduino board \cite{arduino} which is connected to the onboard computer by USB connection. 

The 2-link manipulator's components are: Three DC motors; HS-422 (Max torque = $0.4$  N.m) for the gripper, HS-5485HB  (Max torque = $0.7$  N.m) for joint $1$, and HS-422 (Max torque = $0.4$  N.m) for joint $2$. Motor's Driver (SSC32) which is used as an interface between the main control unit and the  motors. Wireless Play Station 2 (WPS2) R/C is used to send commands to manipulator's motors remotely. The manipulator structure components are; Aluminum Tubing - $1.50$  in diameter, Aluminum Multi-Purpose Servo Bracket, Aluminum Tubing Connector Hub, and Aluminum Long "C" Servo Bracket with Ball Bearings. The Arduino board (Mega 2560) is utilized as an interface between the low level peripherals (such as ultrasonic sensor, WPS2 receiver, and motor driver (SSC32)), and the onboard computer. See Appendix \ref{app:experimentalsystem-manp} for more technical details.

The Power management system is setup as following:
A $11.1$ V Li-Po battery is used to power the motors and the electronics on board. A voltage regulator circuit is used to convert this voltage to $5$ V DC to power the manipulator motors and its avionics. The Hokuyo sensor, Ultrasonic sensor, and the Arduino board are powered by the $5$ V from the USB port of the onboard computer.  A low voltage detection is implemented in the quadrotor making it lands once the battery voltage drops below $8$ V. See Appendix \ref{app:experimentalsystem-power} for power connections.

The ground station consists of a computer, joystick, and Futuba R/C. It allows a user to tele-operate the system such as visualization, sending commands, and emergency recovery.

In this work, a position controller and the position data fusion algorithms are implemented on the HLP, based on the Hokuyo and Ultrasonic sensors input from the onboard computer and the inertial data provided by the LLP. On the LLP, the attitude controller is used as inner loop.

\section{Software} 

An Ubuntu Linux 12.04, with Robot Operating System (ROS) framework (fuerte version) \cite{ros}, is installed on both the onboard computer and the ground station computer. 

Since multiple different computation subsystems (PC, Onboard computer, autopilot board, and Arduino board) are used, which need to communicate among each other, the ROS is used as a middle-ware, see Appendix \ref{app:experimentalsystem-ros} for more details about ROS. This also enable us to communicate with the ground station over the WiFi data link for monitoring and control purposes. In addition, the ROS framework has the essential drivers and software for processing the data from the Hokuyo sensor to find the horizontal position.

The algorithm on the HLP is implemented based on a Software Development Kit (SDK) that provides all the communication routines to allow us to send/receive the low level commands/measurements to/from the LLP. This algorithm is used to implement the data fusion algorithm and the position control. A C++ software (i.e., packages and nodes) is developed under ROS to manage and process data among the the computers systems.

A program is implemented on Arduino board based on a C++ Arduino IDE to acquire and process data from the sonar to get the vertical position. In addition, it receives the manipulator's commands from the WPS2 and send the control signal to the motor drivers. 

\section{Identification}

\begin{figure}[t]
	\centering
	\includegraphics[width=1\columnwidth]{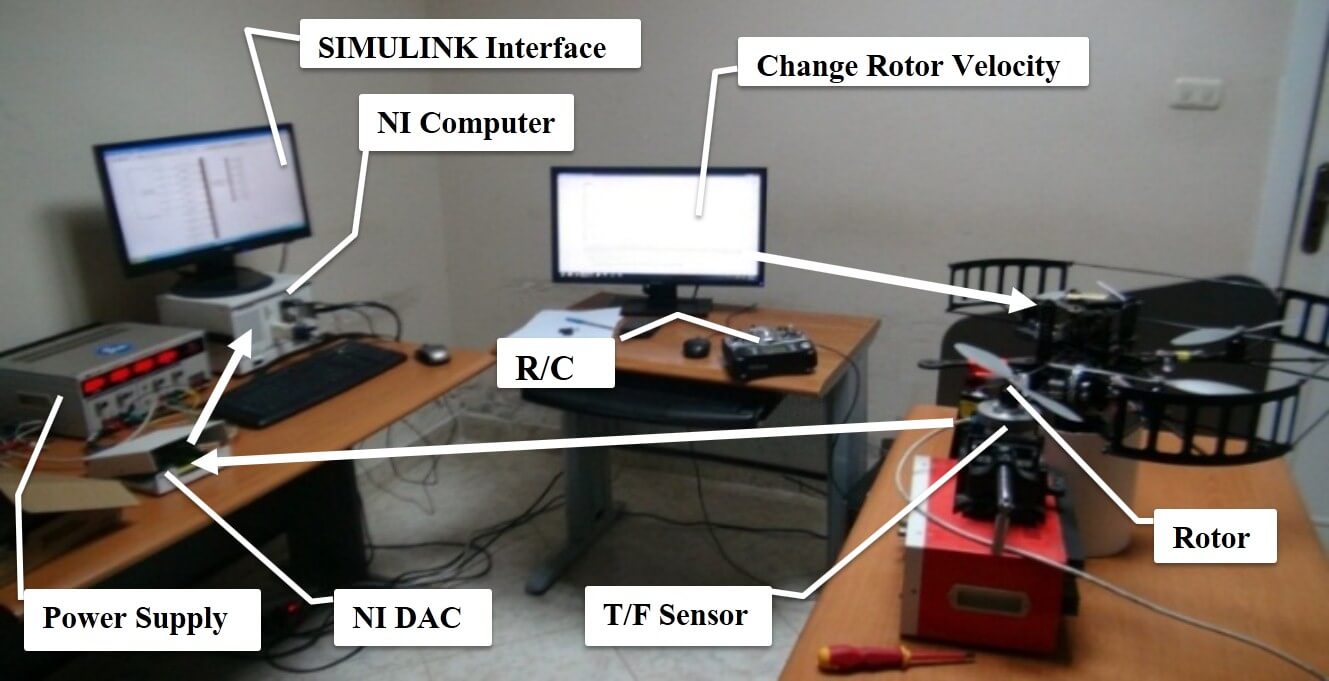}
	\caption{Experiment to estimate the rotor coefficients}
	\label{iden-exp}
\end{figure}

The aim of this section is to find the real parameters of the system to be used in the control design and to emulate a quite realistic setup during the simulation study. The identified parameters include the structure parameters  (e.g., mass, geometrical parameters, and mass moment of inertia) and rotor parameters (i.e., $K_{f_j}$ and $K_{m_j}$). To calculate the structure parameters, a 3D CAD model is developed using SOLIDWORKS. To estimate the rotor parameters, an experimental setup is carried out, see Fig. \ref{iden-exp}. 
In this experiment, the rotor is mounted on a $6$-DOF torque/force sensor (see Appendix \ref{app:experimentalsystem-tfsensor} for more technical details) that is connected to a NI Data Acquisition Card (NI DAC). Then, the DAC is connected to a PC, running SIMULINK program as an interface, to read data from DAC. The velocity of rotor is changed gradually, and in each time, the generated thrust and drag moment is measured and recorded using SIMULINK program. By using MATLAB Curve Fitting toolbox the acquired data of thrust and moment is fitted to be in the form of (\ref{eq:thrust} and \ref{eq:dragmoment}). Thus, the thrust and moment coefficients can be obtained. 

\begin{figure}[!h]
	\centering
	\includegraphics[width=0.5\columnwidth]{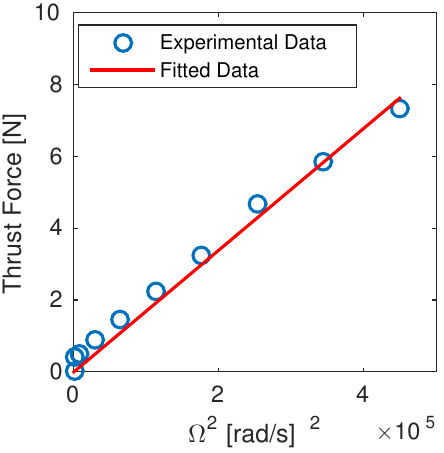}
	\caption{Relation between the generated thrust force and the quadrotor's squared motor speed}
	\label{thurst-iden}
\end{figure}

\begin{figure}[!h]
	\centering
	\includegraphics[width=0.5\columnwidth]{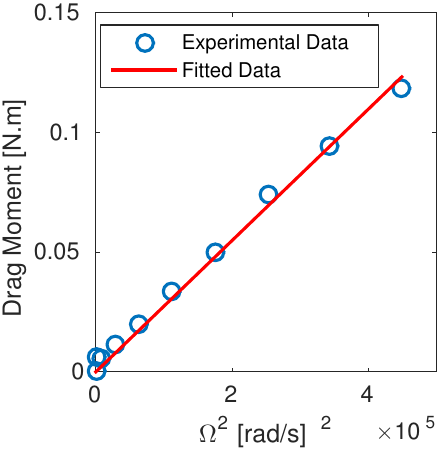}
	\caption{Relation between the generated drag moment and the quadrotor's squared motor speed}
	\label{moment-iden}
\end{figure}

Figure \ref{thurst-iden} (\ref{moment-iden}) shows the relationship between generated thrust force (drag moment) and the squared rotor speed. 

To measure the goodness of data fitness, the Root Mean Squared Error (RMSE) of the curve fitting is calculated, which is 0.3378 for the thrust relation and 0.003742 for the moment relation. 

The identified parameters are collected in Table \ref{tab:sys_par}. It is noted that any error in the estimated parameters will be compensated by using a robust control technique. 
\setlength{\extrarowheight}{3pt}
\begin{table}[h]
	\caption{System Parameters}
	\label{tab:sys_par}
	\begin{center}
		\setlength{\tabcolsep}{1pt}
		\begin{tabular}{|c|c|c|c|c|c|}
			\hline
			Par.	&Value	&Unit&	Par. &	Value&	Unit \\
			\hline
			$m$&	$1$ &	$kg$ &	$l_2$&	$85\times10^{-3}$&	$m$ \\
			\hline
			$d_q$ &	$223.5\times10^{-3}$ &	$m$ &$m_0$&	$30\times10^{-3}$&	$kg$\\
			\hline
			$I_x$&	$13.215\times10^{-3}$&	$N.m.s^2$& $m_1$	&$55\times10^{-3}$	&  $kg$\\
			\hline
			$I_y$	& $12.522\times10^{-3}$ &	$N.m.s^2$ &	$m_2$ &	$112\times10^{-3}$& $kg$\\
			\hline
			$I_z$&	$23.527\times10^{-3}$&	$N.m.s^2$	& $I_r$	& $33.216\times 10^{-6}$& $N.m.s^2$\\
			\hline
			$l_0$	& $30\times10^{-3}$ & $m$ &	$l_1$& $70\times10^{-3}$&$m$ \\
			\hline
			$K_{F_1}$ & $1.667\times10^{-5}$ & $kg.m.rad^{-2}$ & $K_{F_2}$&$1.285\times10^{-5}$&$kg.m.rad^{-2}$\\
			\hline
			$K_{F_3}$ &$1.711\times10^{-5}$&$kg.m.rad^{-2}$&$K_{F_4}$& $1.556\times10^{-5}$ &$kg.m.rad^{-2}$\\
			\hline
			$K_{M_1}$ & $3.965\times10^{-7}$ &$kg.m^{2}.rad^{-2}$&$K_{M_2}$ & $2.847\times10^{-7}$ &$kg.m^{2}.rad^{-2}$\\
			\hline
			$K_{M_3}$ &$4.404\times10^{-7}$& $kg.m^{2}.rad^{-2}$ & $K_{M_4}$&$3.170\times10^{-7}$ & $kg.m^{2}.rad^{-2}$ \\
			\hline
		\end{tabular}
	\end{center}
\end{table}
\section{6-DOF State Estimation} \label{se:pose_est}

\begin{figure}[!h]
\centering
\includegraphics[width=1\columnwidth]{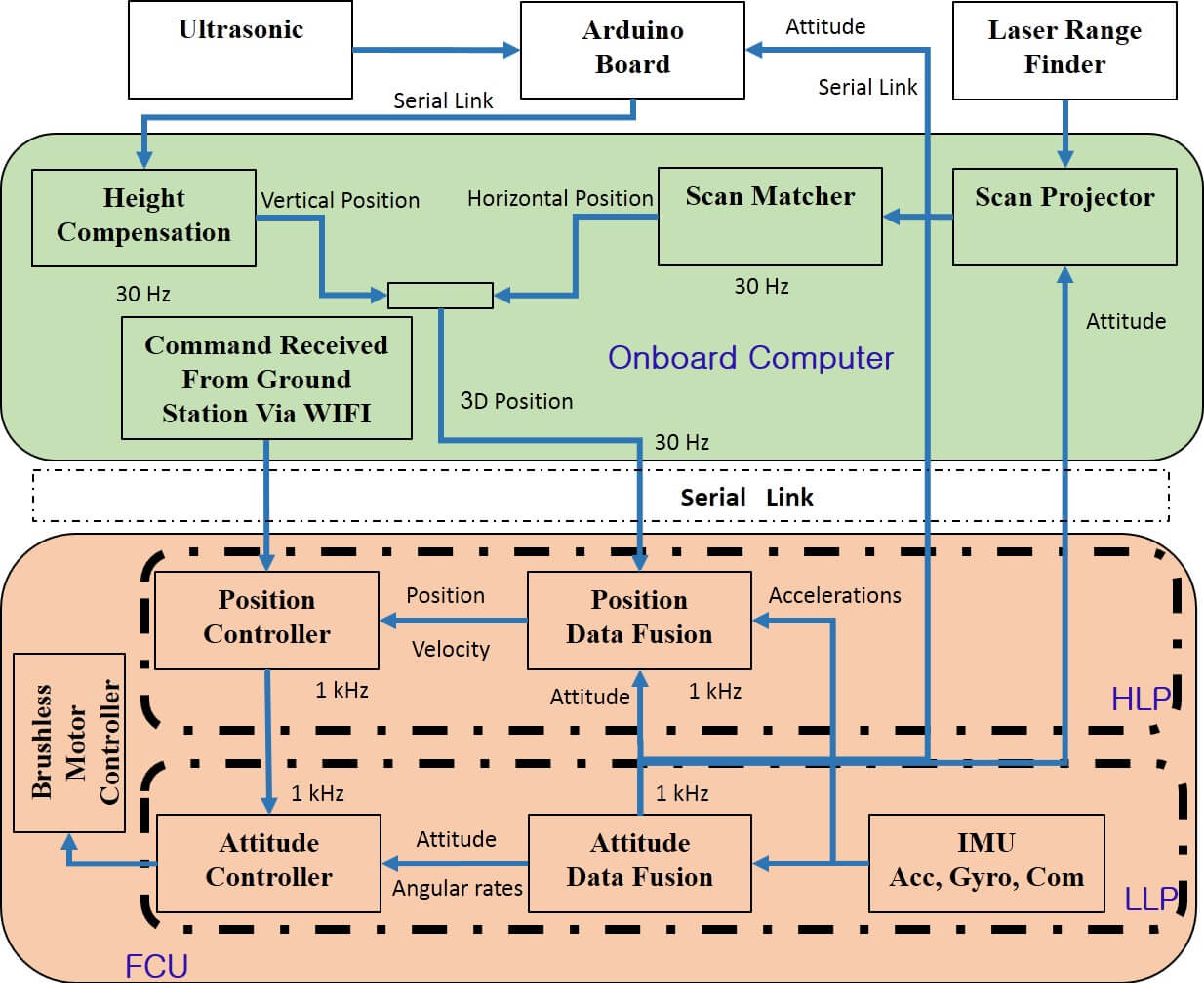}
\caption{Block diagram of the state estimation scheme}
\label{fig:pose_est_blk}
\end{figure}

Position holding is one of the most important factors to achieve the accurate aerial manipulation. The accurate measurements are crucial for implementing the position holding.
In this section, the proposed state measurement/estimation scheme, which will be used to find the system states, is described. The state estimation system is presented in details in Fig. \ref{fig:pose_est_blk}.  

On the LLP, there is a data fusion algorithm to find the quadrotor orientation at rate of 1 kHz. In addition, the IMU itself can provide the angular rates directly. Thus, both the angular position and velocity of the quadrotor are available. The position measurements are obtained by using the laser and ultrasonic sensors. Since this process is slow (about $30$ Hz) compared to the motion of the quadrotor (quadrotor dynamics have high bandwidth 1kHz),
this information is fused with inertial sensor data (body accelerations $\ddot{p}_b^b$) provided by the IMU at a rate of $1$ kHz. The outputs of that filter are finally fed into a position controller. The expensive computation processing of the laser data  is run on the onboard computer at approximately $30$ Hz, while the fusion filter and the position controller are executed on the HLP at $1$ kHz. Note that, this ensures the minimum possible delays, and allows us to handle the fast movements of the quadrotor.

The horizontal position from the laser range finder sensor can be obtained as following.

First, the laser scan is projected orthogonally onto the horizontal $xy$-plane by using the roll and pitch angles of the last state. The projected data is then passed to a scan matcher \cite{dryanovski2013open}. The scan matcher uses the yaw angle reading from the IMU as an initial guess for the yaw angle of the quadrotor. The output of the scan matcher includes $x$, $y$, and yaw angle pose components.

The vertical position from the ultrasonic sensor is obtained as following.
The ultrasonic sensor is connected to the Arduino board. The sonar signal is transmitted and the echo of it is received. By measuring the time of signal traveling, one can calculate the distance ($s_z$) that is the vertical height of the sensor. However, the calculated distance must be compensated due to the roll and pitch rotations of quadrotor by
\begin{equation}
z = s_z \cos(\phi) \cos(\theta).
\end{equation} 
Therefore, the measured horizontal position and vertical position (at rate of $30$ Hz), and the angles and linear accelerations (at rate of $1$ kHz) are available now. However, the $6$-DOF measurements rate are required to be high enough to enable a high update rate in the position control loop to match the quadrotor's system dynamics. Thus, these measurements must be fused to achieve this high rate. The filter is designed and decoupled for all the three axes $x$, $y$, $z$ in the world inertial frame. The body-fixed accelerations, $\ddot{p}_b^b$, are rotated in the global frame by the rotational matrix, $R_b$, based on the attitude angles provided by the LLP, as
\begin{equation}
\ddot{p}_b = R_b \ddot{p}_b^b - [0 \ 0 \ g_r]^T, 
\end{equation} 
where $g_r$ is the acceleration due to the gravity.
The filter is based on a Luenberger observer \cite{achtelik2011onboard}. In the following, only the filter for the $x$-axis is described and and thus it can be applied for the other axes accordingly. The filter uses the position, $p_{b,x}$, speed, $\dot{p}_{b,x}$, and the acceleration sensor bias, $b_x$, as the states. The acceleration expressed in the world frame is the system input. The
measurement is the position sensors' readings, $p_{s,x}$, from the Laser (for $x$ and $y$) or Sonar (for $z$). 

Let us define the state variable as $X_{obs} = [p_{b,x} \ \dot{p}_{b,x} \ b_x]^T$, $ U_{obs} = \ddot{p}_x$, and $Y_{obs}= p_{s,x}$. Then, the observer state equations can be written as
\begin{equation}
\begin{split}
\dot{\hat{X}}_{obs} = A_{obs} \hat{X}_{obs} + L_{obs}(Y-\hat{Y}) +B_{obs} U_{obs}, \\
Y = C_{obs}  \hat{X}_{obs},
\end{split}
\end{equation}
where 
$
A_{obs}= \begin{bmatrix}
0 & 1 & 0 \\
0 & 0 & 1 \\
0 & 0 & 0 
\end{bmatrix}
$, 
$
B_{obs}= \begin{bmatrix}
0 \\
1 \\
0 
\end{bmatrix} 
$,
$
C_{obs}= \begin{bmatrix}
1 & 0 & 0  
\end{bmatrix} 
$, and 
$
L_{obs}= \begin{bmatrix}
L_{obs_1} \\
L_{obs_2} \\
L_{obs_3} 
\end{bmatrix} 
$.

The output of this position filter is the linear position and velocity of the vehicle at update rate of $1$ kHz. 

Unlike the currently developed state measurement system for aerial manipulation system, with this proposed scheme, one can  measure/estimate the system states in different places indoors. In addition, one can utilize this scheme for being used outdoors by using GPS-based position holding for wide operation (quadrotor is far from the target) and using these sensors when the quadrotor is near the objects. If a high performance onboard camera is available, then the data from the laser can be fused with the camera data to obtain more accurate posing indoors and outdoors.
\section{Control Design for the Experiments} 

\begin{figure}[h]
\centering
\includegraphics[width=1\columnwidth]{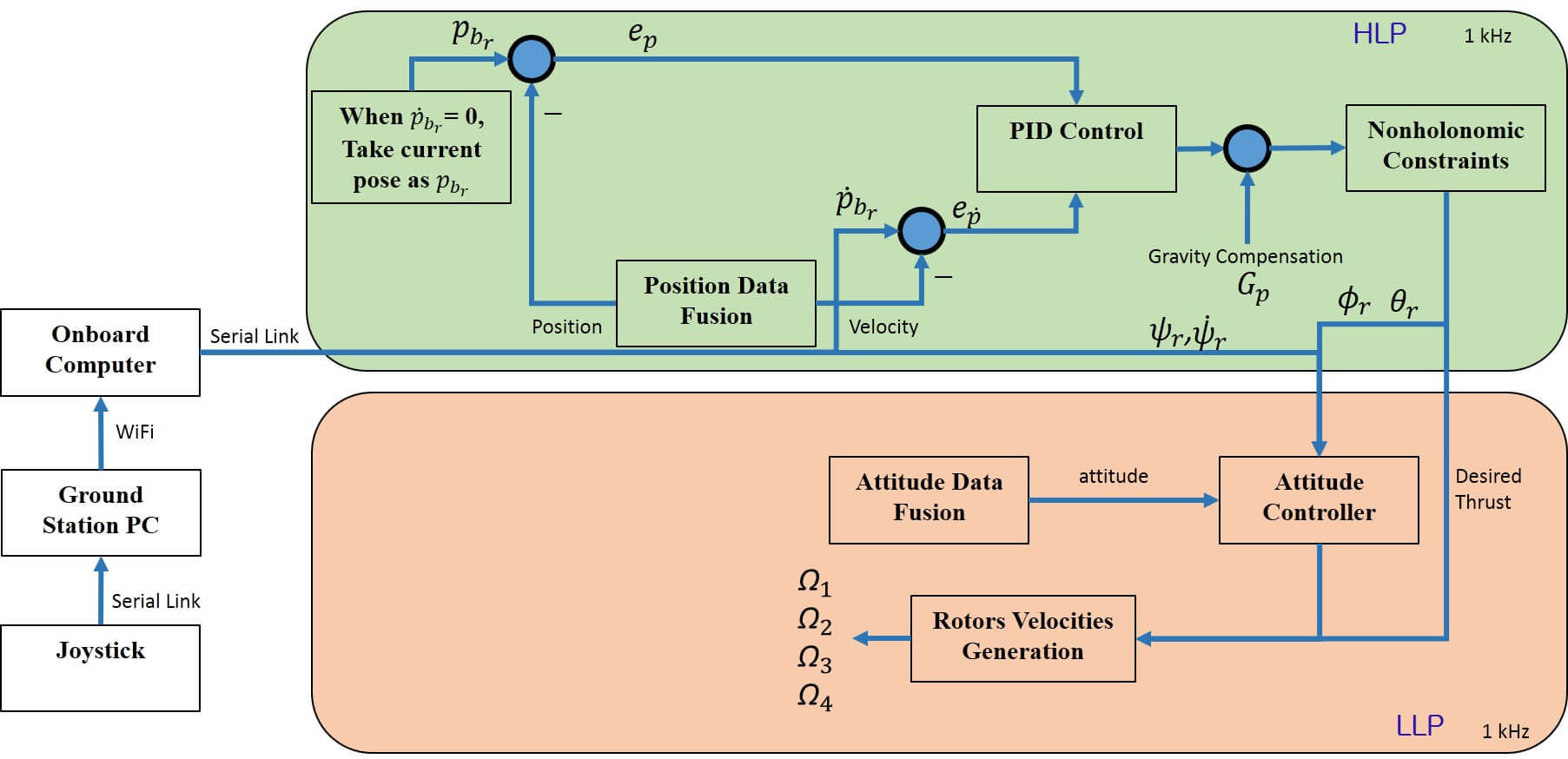}
\caption{Block diagram of the control structure; Note that the circle block does the summation function, and the input with minus signal (-) means+ subtraction while the input with no signal means addition}
\label{fig:ctrl_blk}
\end{figure}

In this section, the real time control system is presented. Fig. \ref{fig:ctrl_blk} shows the control structure.   
Since the quadrotor is an under-actuated system, i.e., only $4$ independent control inputs are available against the $6$ DOF, the position and the yaw angle are usually the controlled variables, while the pitch and roll angles are used as intermediate control inputs for horizontal position control. For position control, a cascade structure is used. As inner loop, the well tested attitude loop provided by the LLP of the FCU is used. The outer loop is the position loop, and is implemented on the HLP based on the concept of PID control with gravity compensation as following
\begin{equation}
\tau_p = [\tau_x \ \tau_y \ \tau_z]^T = K_p (p_{b_r} - p_b) + K_d (\dot{p}_{b_r} - \dot{p}_b) + K_i \int_{0}^{t} (p_{b_r} - p_b) dt + G_p,
\label{eq:tp}
\end{equation} 
where $G_p$ is the gravity term of the translational part in the dynamic equation of motion (\ref{eq:dyn_gen}) and it can be obtained as $G(1:3)$. $K_p$, $K_d$, and $K_i$ are PID tuning parameters, and $p_{b_r}$ and $\dot{p}_{b_r}$ are the reference position and velocity, respectively. 

Roll and pitch commands are generated from the outer loop as reference to the inner attitude controller in order to reach and maintain desired $y$ and $x$ position, respectively. Thrust commands are generated by the outer controller to control the height by sending desired thrust values. The desired values for the intermediate controller, $\theta_r$ and $\phi_r$, are obtained from the output of position controller through the following relation
\begin{equation}
\begin{bmatrix}
\theta_r \\
\phi_r
\end{bmatrix} = \frac{1}{\tau_p(3)}
\begin{bmatrix}
C_{\psi} & S_{\psi} \\
S_{\psi} & -C_{\psi} 
\end{bmatrix}
\begin{bmatrix}
\tau_p(1) \\
\tau_p(2) 
\end{bmatrix}. 
\label{eq:sigmad_imp}
\end{equation} 
This relation is derived from (\ref{eq:Rb}) based on the small angle approximation of the roll and pitch angles.

Since it is aimed to teleoperate  the system, as a first step to autonomous mode, the manipulation task can be achieved by sending velocity commands to the controller. Then, for position holding, one will send a zero velocity command and make the current position as the desired position (at the instant of sending zero command velocity). The velocity commands can be sent using the joystick, where the quadrotor does not need to reach a target position but just follows the desired velocity, allowing a smooth flight. Moreover, this control scheme can be utilized to send position commands to work in autonomous mode.

For the manipulator joints' control, the built-in position sensor with a PID controller is used. Since the dynamics of the manipulator is simple, one use this built-in controller as a first test. If it does not provide required performance, design another advanced controller will be designed. The simplicity in the manipulator dynamics comes from the lightweight and slow motion of the  manipulator parts.

The desired position can be obtained from the command velocity and measured position as in algorithm \ref{alg:pos_des}.
\begin{algorithm}
\caption{Algorithm to generate the desired position}
\label{alg:pos_des}
\begin{algorithmic}[1]
	\State $Initailization:$ $k$ = $1$ and $p_{b_r}(0)$ = $0$
	\While()
	\If{$\dot{p}_{b_r}$ = $0$}
	\State $p_{b_r}(k)$ = $p_{b_r}(k-1)$
	\Else
	\State $p_{b_r}(k)$ = $p_{b}(k)$
	\EndIf
	\EndWhile
\end{algorithmic}
\end{algorithm} 

\section{Simulation Results}

This proposed control strategy is applied in simulation MATLAB/SIMULINK program to the model of the considered aerial manipulation robot.  In order to make the simulation environment to be quite realistic, a normally distributed measurement noise, with mean of $10^{-3}$ and standard deviation of $5 \times 10^{-3}$, has been added to the measured signals. In addition, one can use the parameter obtained from the experimental identification that is presented previously. The controller parameters are presented in Table \ref{tab:pid_par_sim}. These parameters are tuned to get a satisfactory response.

\begin{figure}[h]
\centering
\begin{tabular}{cc}
	\subfloat[Motion in $x$ axis]{\includegraphics[width=0.4\columnwidth]{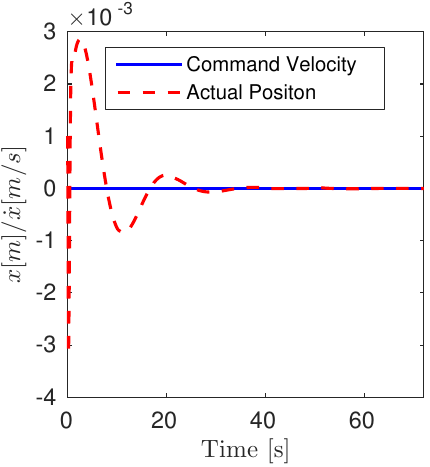}\label{fig:sim_results_x}}&
	\subfloat[Motion in $y$ axis]{\includegraphics [width=0.4\columnwidth]{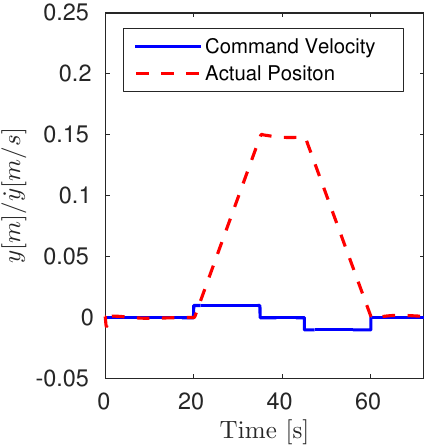}\label{fig:sim_results_y}}\\
	\subfloat[Motion in $z$ axis]{\includegraphics[width=0.4\columnwidth]{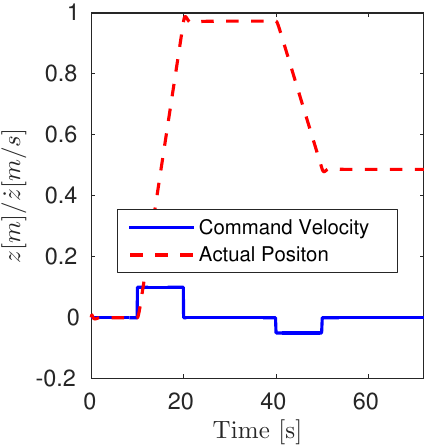}\label{fig:sim_results_z}}&
	\subfloat[Motion around $z$ axis]{\includegraphics [width=0.4\columnwidth]{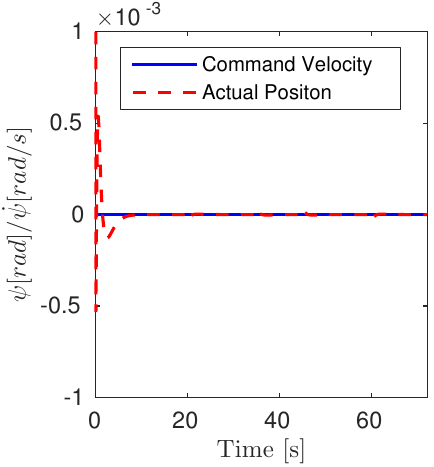}}\\
	\subfloat[Motion of joint 1]{\includegraphics[width=0.4\columnwidth]{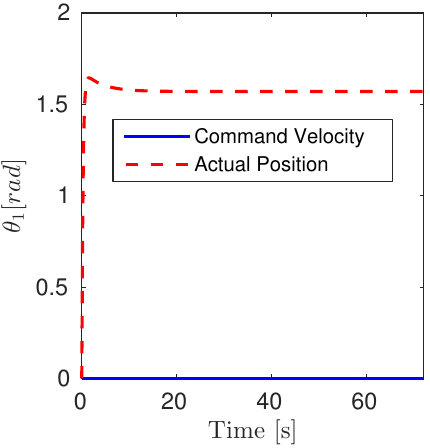}\label{fig:sim_results_th1}}&
	\subfloat[Motion of joint 2]{\includegraphics [width=0.4\columnwidth]{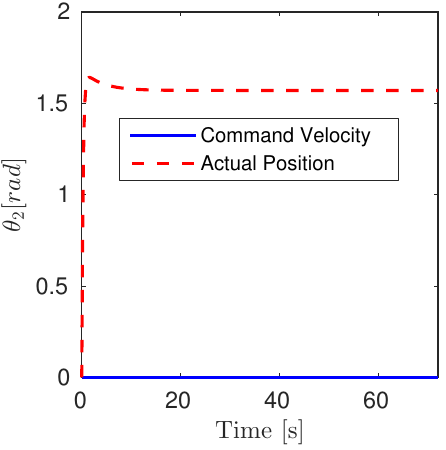}\label{fig:sim_results_th2}}\\
\end{tabular}
\caption{Simulation results: actual position/command velocity history of the system}
\label{fig:sim_results}
\end{figure}

The simulation results are presented in Fig. \ref{fig:sim_results}. In this figure, the target position is assumed to be in the direction of the $y$-axis of quadrotor, and the object (with weight of $50$ g) is attached to the end-effector gripper before starting the operation. During the vehicle taking off, by sending a positive velocity command in the $z$ direction until it is reached at suitable height, see Fig. \ref{fig:sim_results_z}, a very small drift in some coordinates occurs, see Fig. \ref{fig:sim_results}, which is then quickly compensated. At this moment, a zero velocity command is given such that the vehicle is hold in its initial position due to the position holding control. To move towards the target position, in which one have to place the carried object, a positive velocity command is sent in the $y$ direction, see Fig. \ref{fig:sim_results_y}. As soon as the target position is reached, a zero $y$ velocity and a negative $z$ velocity are sent to put the end-effector near to the destination point and then a zero $z$ velocity is given to hold the vehicle at this point, and a command is sent to open the gripper to release the object. After object releasing, a negative velocity is sent in the $y$ direction to go back to the starting point again. Once it is reached at this point, a zero velocities are sent to all coordinates to finish the task. Moreover, a zero velocity command to the manipulator joints is always sent , see Figs. \ref{fig:sim_results_th1} and \ref{fig:sim_results_th2}. These results show the feasibility and efficiency of the proposed system. These results will be verified experimentally in the next section.
\begin{table}
\caption{Controller Parameters in Simulation}
\label{tab:pid_par_sim}
\begin{center}
	\begin{tabular}{|c|c|c|c|}
		\hline
		Parameter & Value & Parameter & Value\\
		\hline
		$K_{p_{x,y}}$ & $2$ & $K_{d_{x,y}}$ & $7$\\
		\hline
		$K_{i_{x,y}}$ & $0.5$ & $K_{p_{z}}$ & $10$\\
		\hline
		$K_{d_{z}}$ & $10$ & $K_{i_{z}}$ & $5$\\
		\hline
	\end{tabular}
\end{center}
\end{table}
\section{Experimental Results} \label{se:exp_result}

\begin{figure}
	\centering
	\begin{tabular}{cc}
		\subfloat[Taking off, with the object, from home position]{\includegraphics[width=0.5\columnwidth]{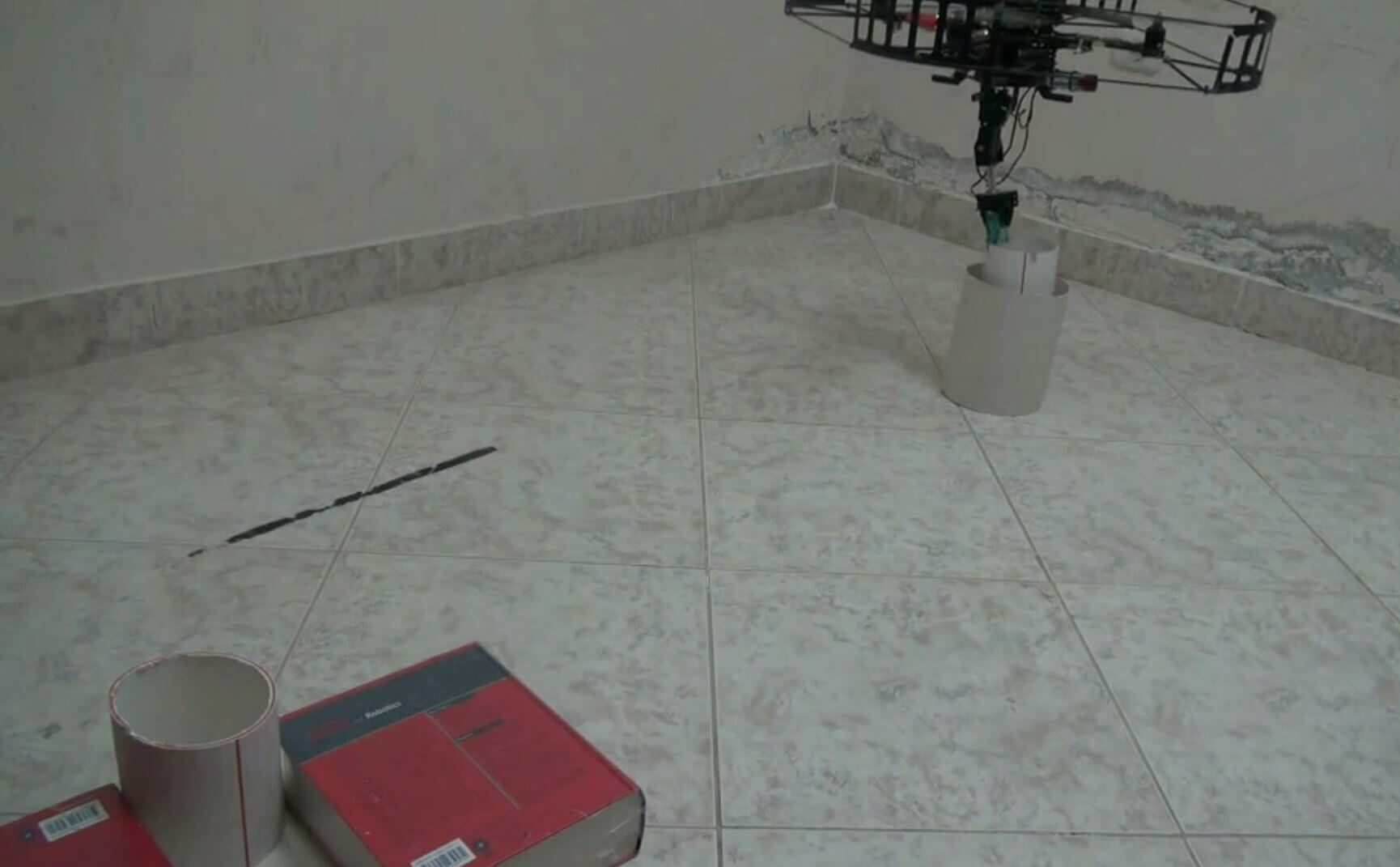}}&
		\subfloat[Moving to the target position]{\includegraphics [width=0.5\columnwidth]{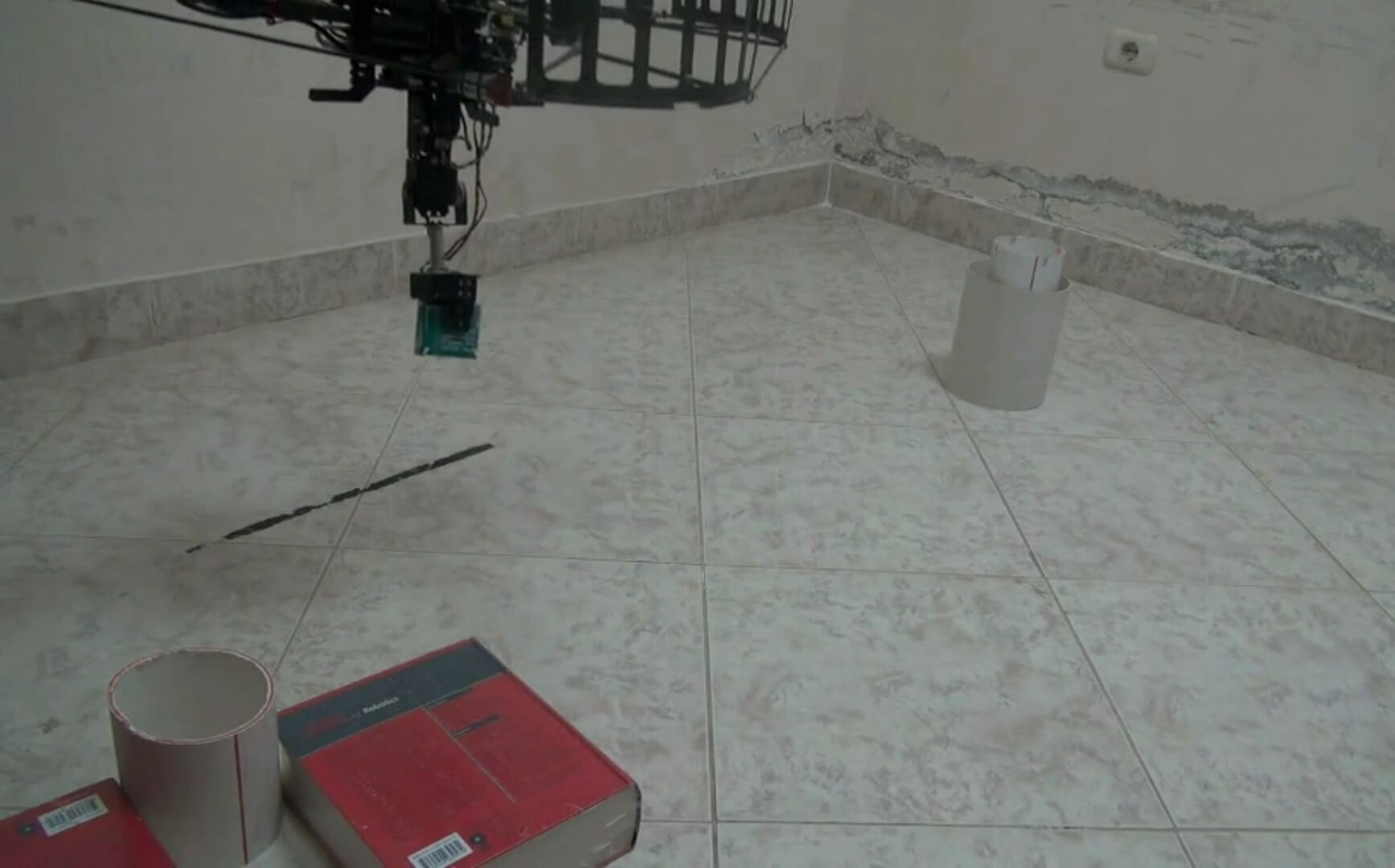}}\\
		\subfloat[Reaching at the target position]{\includegraphics[width=0.5\columnwidth]{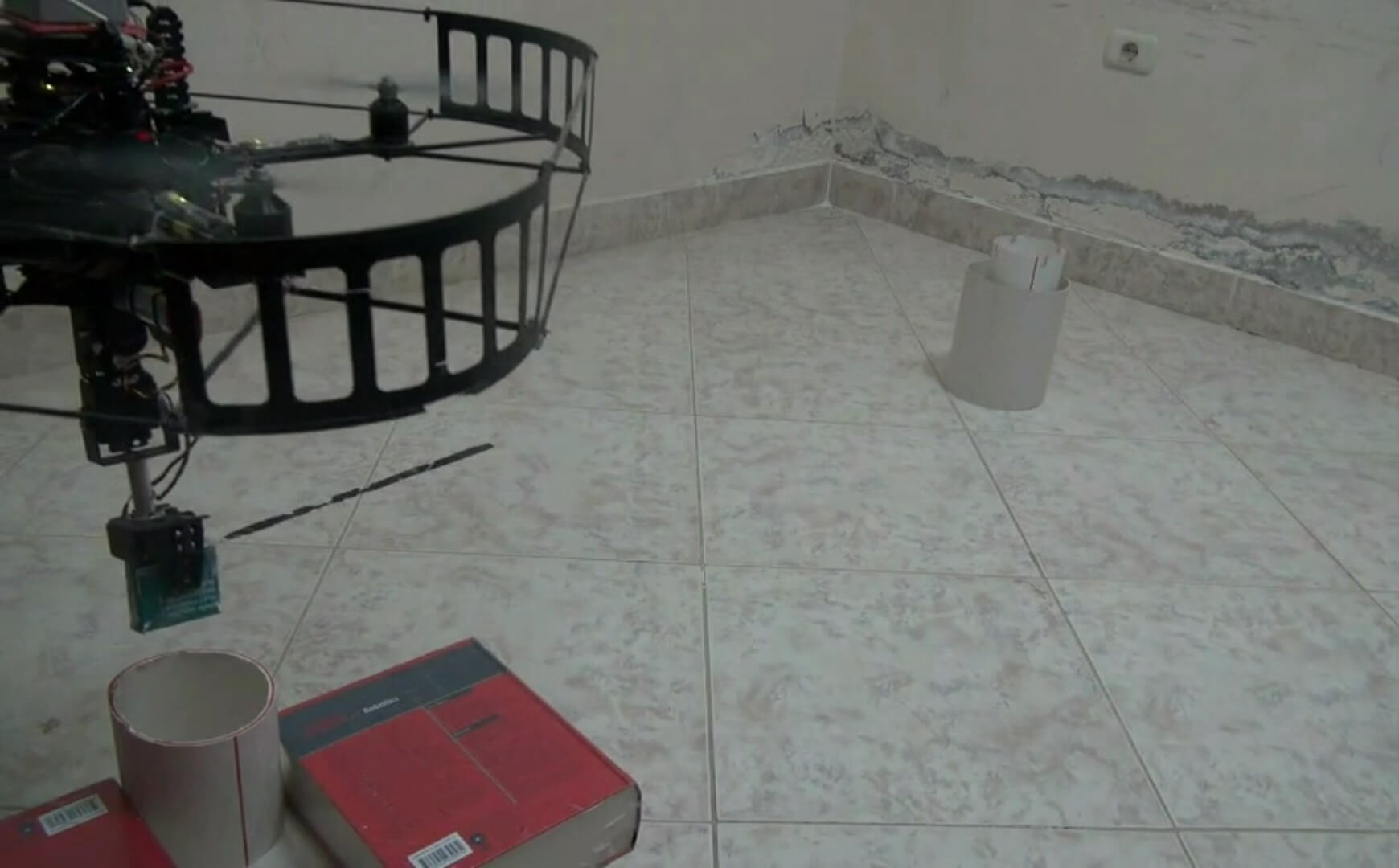}}&
		\subfloat[Releasing the object]{\includegraphics [width=0.5\columnwidth]{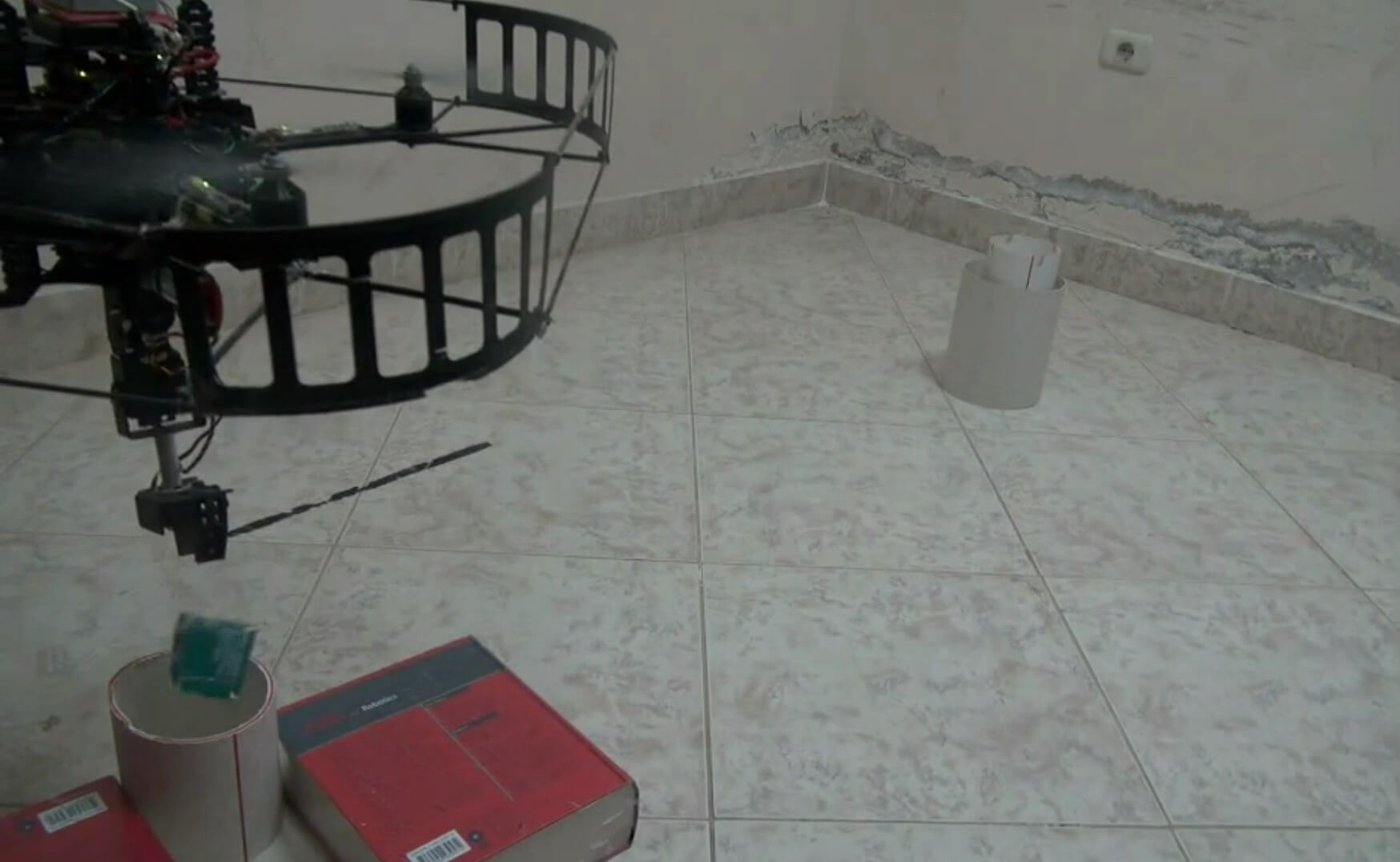}}\\
		\subfloat[Returning back to the home  position]{\includegraphics [width=0.5\columnwidth]{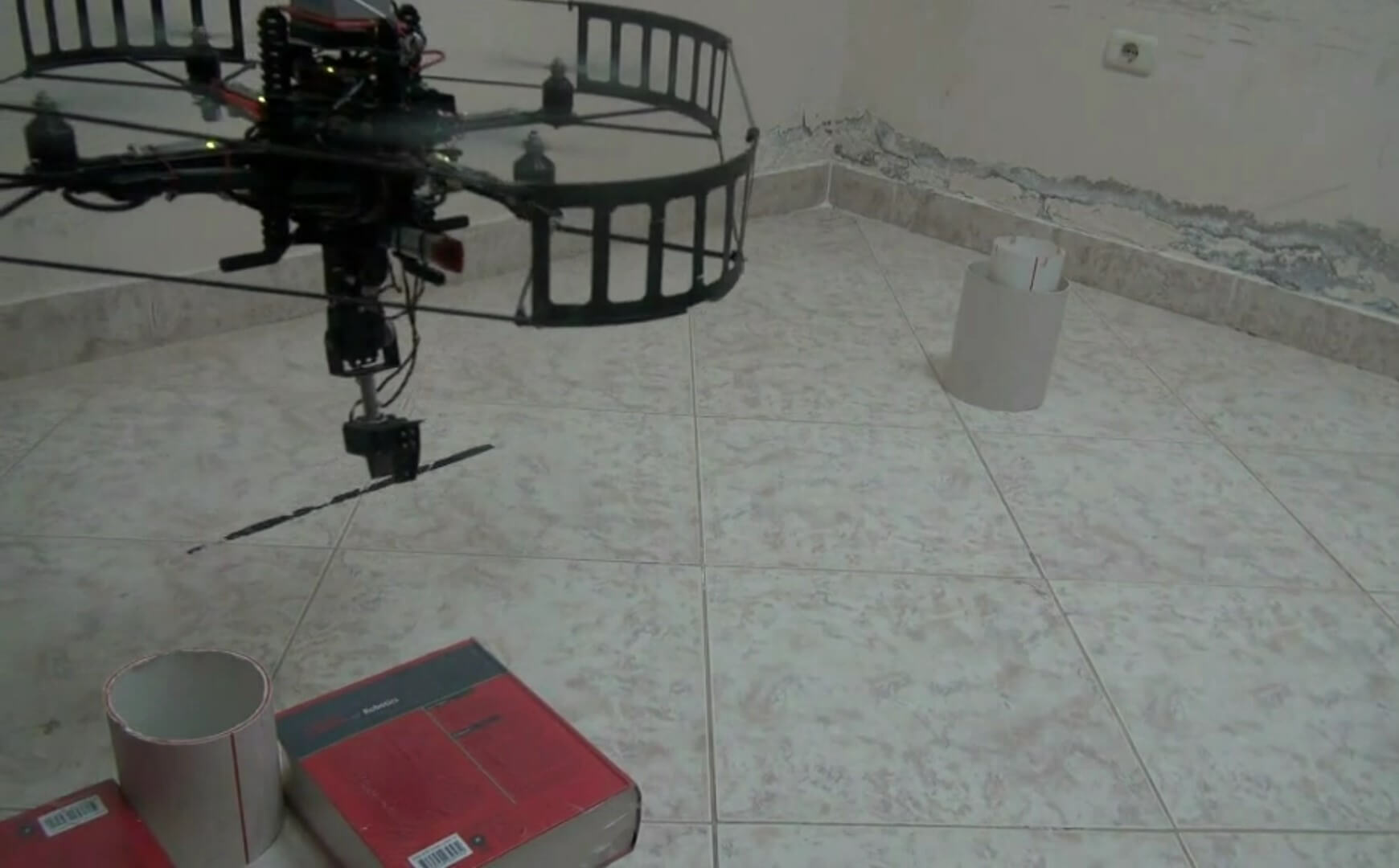}}&
		\subfloat[Reaching at the home position]{\includegraphics [width=0.5\columnwidth]{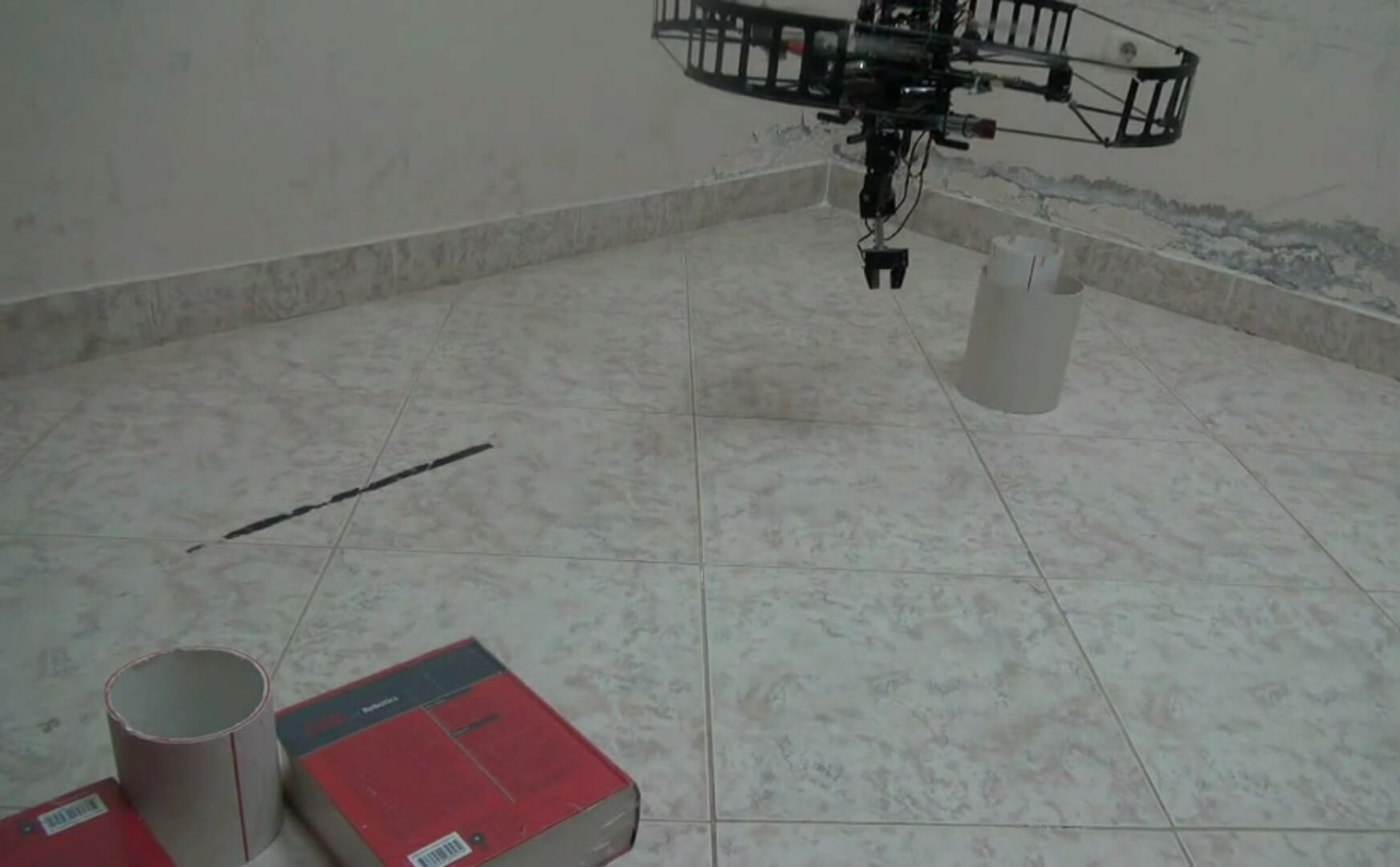}}\\
	\end{tabular}
	\caption{Pictures taken during the experiment}
	\label{fig:exp_opert}
\end{figure}

\begin{figure}
	\centering
	\begin{tabular}{cc}
		\subfloat[Motion in $x$ axis]{\includegraphics[width=0.4\columnwidth]{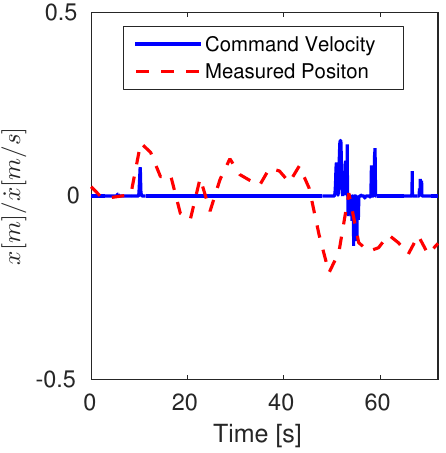}\label{fig:exp_results_x}}&
		\subfloat[Motion in $y$ axis]{\includegraphics [width=0.4\columnwidth]{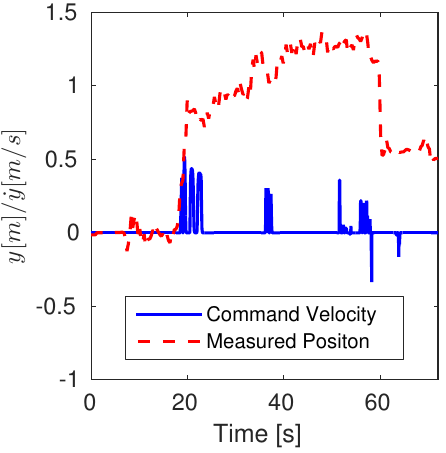}\label{fig:exp_results_y}}\\
		\subfloat[Motion in $z$ axis]{\includegraphics[width=0.4\columnwidth]{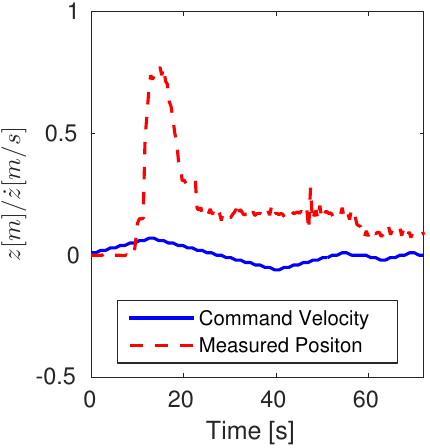}\label{fig:exp_results_z}}&
		\subfloat[Motion around $z$ axis]{\includegraphics [width=0.4\columnwidth]{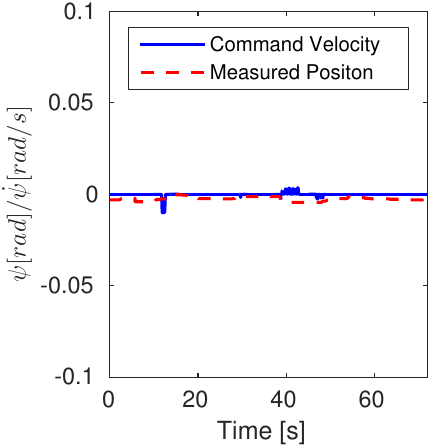}\label{fig:exp_results_ep}}\\
		\subfloat[Motion of joint 1]{\includegraphics[width=0.4\columnwidth]{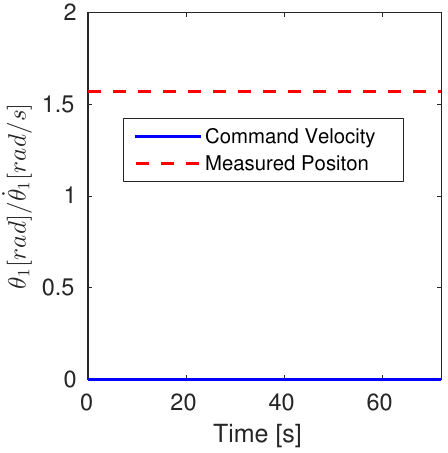}\label{fig:exp_results_th1}}&
		\subfloat[Motion of joint 2]{\includegraphics [width=0.4\columnwidth]{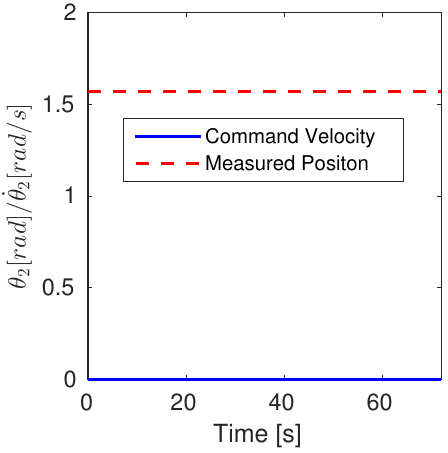}\label{fig:exp_results_th2}}\\
	\end{tabular}
	\caption{Experimental results: measured position/command velocity history of the system}
	\label{fig:exp_results}
\end{figure}

In this section, experimental results of the proposed system are presented and discussed. The objective is to transfer an object in a teleoperation mode. For the teleoperation purpose, it is assumed that one cannot know the accurate coordinates of the target position before operation. Thus, velocity commands will be sent to move the system to that position. The same scenario that done in the simulation study will be implemented except that in some situations a velocity command in the $x$-direction will be sent because it is difficult to determine the target place accurately along $y$-axis as assumed in the simulation test. Then, if the target position is reached, the position holding is activated to place the object in the target position. The target object, to be transfered to another place, has a weight of $50$ g which will be attached to the gripper at the start of system operation. Experimental controller parameters are given in Table \ref{tab:pid_par_exp}. Fig. \ref{fig:exp_opert} presents photos from the experiment that is divided into six phases; starts from taking off (by given velocity command in the $z$ direction) with the object attached to the gripper, moving to the destination by giving velocity command in the $y$ direction and fine adjustment in the $x$ direction, then reaching at the target position, releasing the object by achieving position holding by giving zero velocity commands, moving towards the home position, and finally, reaching the home position. The real experimental data is recored in the ground station through the ROS framework and the WiFi connection. Fig. \ref{fig:exp_results} presents the experimental results. In this figure, the velocity commands as well as the actual  position for the coordinates $x$, $y$, $z$, $\psi$, $\theta_1$, $\theta_2$ are presented. The manipulator joints commands are sent using WPS2 controller assuming that the built-in controller is good enough to track accurately the sent commands. To start the mission, see Fig. \ref{fig:exp_results_z}, a positive $z$ velocity is sent to take off, then position holding is activated for small period by sending a zero $z$ velocity, after that the platform is commanded to go to a suitable height at lower level by sending negative $z$ velocity until the target height is reached, then one can activate the position holding again. During the vehicle taking off, a small drift occurs in the $x$, $y$, $z$, and $\psi$ directions which is quickly recovered due to the position holding at the starting point. Fig. \ref{fig:exp_results_x} shows that one have to send velocity commands in $x$ direction, as it is claimed before, to fine tune the vehicle position towards the target place accurately. After taking off, the vehicle moves in $y$ direction towards the target place, see Fig. \ref{fig:exp_results_y}, till reaching the target point at which the position holding is activated. At this point, a command is sent to the gripper to open and release the object. After releasing the object, the vehicle returns back to the starting point, then a negative $z$ velocity is sent until the platform becomes at height suitable to the base station. In Fig. \ref{fig:exp_results_ep}, as it is planed, there is no any command in $\psi$ direction, and consequently there is almost no rotation around $z$. Since, the required task is simple, and there is no need to move the manipulator joints (the manipulator needs to be stretched, i.e., $\theta_1$ and $\theta_2$ = $pi/2$), zero velocity commands are sent to the manipulator joints, see Figs. \ref{fig:exp_results_th1} and \ref{fig:exp_results_th2}. However, for sophisticated tasks, one can command the joints to do the $6$-DOF motion based on the specified task. These results prove the feasibility and a satisfactory efficiency of the proposed system, state estimation scheme, and the control algorithm. However, they show that the control algorithm for position control needs improvements in the future work by using robust estimation control techniques. Moreover, the sonar can be replaced by more accurate sensor.

\begin{table}
\caption{Experimental Controller Parameters}
\label{tab:pid_par_exp}
\begin{center}
	\begin{tabular}{|c|c|c|c|}
			\hline
			Parameter & Value & Parameter & Value\\
			\hline
			$K_{p_{x,y}}$	& $5$ & $K_{d_{x,y}}$	& $12$\\
			\hline
			$K_{i_{x,y}}$	& $2$ & $K_{p_z}$	& $15$\\
			\hline
			$K_{d_z}$& $15$ & $K_{i_z}$	& $4$\\
			\hline
			$L_{{obs}_{x,y}}$ & $[18.01 \ 45.18 \ 0.45]^T$ &$L_{{obs}_{z}}$ & $[18.01 \ 45.18 \ 0.45]^T$ \\
			\hline
		\end{tabular}
	\end{center}
\end{table}



\chapter{\uppercase{Inverse Kinematics Solution}} 
\label{ch:InverseKinematics}


\ifpdf
    \graphicspath{{4_InverseKinematics/figures/}}
\else
    \graphicspath{{4_InverseKinematics/figures/}}
\fi

All the previous research in the aerial manipulation area do not provide solution to the system inverse kinematics. However, in the motion planning of such system, the task is specified in terms of a desired trajectory for the end-effector. To achieve this task, the system inverse kinematics must be available to get the reference quadrotor/joint space trajectories from desired 6-DOF task space trajectories. Then, a controller is designed to track the generated trajectories from the inverse kinematics. Therefore, in this chapter a novel inverse kinematics analysis is presented to get the reference quadrotor/joint space trajectories from desired 6-DOF task space trajectories, in addition to prove that the end-effector of the proposed robot can achieve arbitrary 6-DOF trajectory tracking. The trajectory inverse kinematics of our proposed non-redundant system is difficult because of the existence of high order nonholonomic constraints. Eight complex algebraic differential equations are needed to be solved. However, obtaining exact solution of these equations is impractical. Thus, an algorithm is developed to get approximate efficient solution of these equations. The effectiveness of the proposed algorithm is validated and enlightened via numerical results.

\section{Novel 6-DOF Trajectory Inverse Kinematics} \label{se:inv_kin_pos}

A task for the quadrotor manipulation system is usually specified in terms of a desired trajectory for the end-effector pose, $\chi_{e,r}(t)$ ( position, $p_{e,r}(t)$, and orientation, $\Phi_{e,r}(t)$). In this section, six algebraic kinematic equations relating $\chi_e$ with $q$ are derived.

For the desired task space trajectories, one can assume:
\begin{assumption}
	The desired trajectories for the end-effector are bounded.
\end{assumption}

To find the eight variables of $q$ from the given six variables of $\chi_e$, two additional equations are required. These two equations are the nonholonomic constraints (\ref{sin_ph} and \ref{tan_th}), which are differential equations. 

\begin{figure}[h]
	\centering
	\includegraphics[width=1\columnwidth]{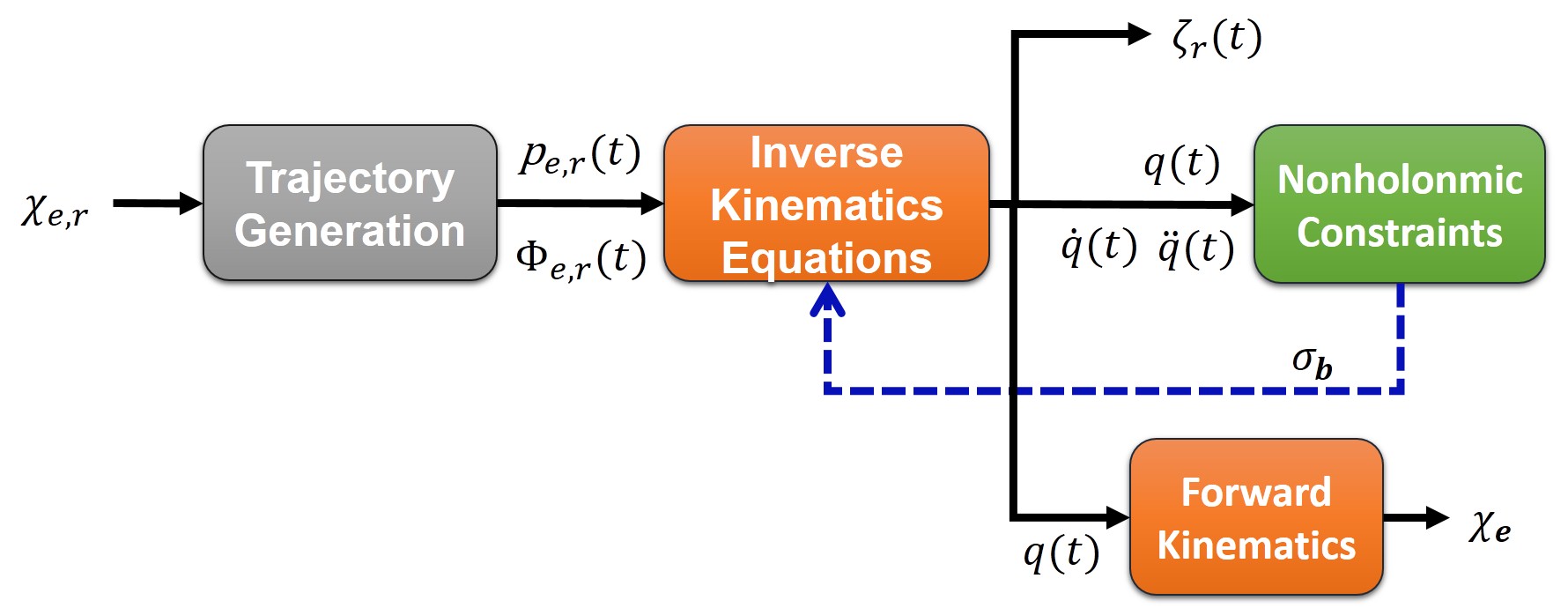}
	\caption{Block diagram of the inverse kinematic algorithm}
	\label{fig:inv_non_blk}
\end{figure}

We propose algorithm \ref{alg:inv_kin} to get a solution to these eight algebraic/differential equations, see Fig. \ref{fig:inv_non_blk}.

\begin{algorithm}[h]
	\caption{Algorithm to solve the inverse kinematics problem}
	\label{alg:inv_kin}
	\begin{algorithmic}[1]
		\State $Initailization:$ Specify the desired 6-DOF trajectory in the task space, $\chi_{e,r}(t)$.
		Put $i=1$, $t_1=0$, and $\Delta t$ = very small positive number ($1$ ms).
		Put $\sigma_b(t_0)$ = $\dot{\sigma}_b(t_0)$ = $\ddot{\sigma}_b(t_0)$ = $[0 \ 0]^T$
	
		\While{($t \leq t_f$)} \Comment{$t_f$ is the final time}

			\State At time $t_i$, obtain the values of $\chi_{e,r}(t_i)$ and put $\sigma_b(t_i)$ = $\sigma_b(t_{i-1})$, $\dot{\sigma}_b(t_i)$ = $\dot{\sigma}_b(t_{i-1})$, and $\ddot{\sigma}_b(t_i)$ = $\ddot{\sigma}_b(t_{i-1})$.
	
			\State Use the six inverse kinematics algebraic equations and get the unknowns $\zeta_r(t_i)$, $\dot{\zeta}_r(t_i)$, $\ddot{\zeta}_r(t_i)$ which with $\sigma_b(t_i)$, $\dot{\sigma}_b(t_i)$, and $\ddot{\sigma}_b(t_i)$ constitute $q(t_i)$, $\dot{q}(t_i)$, and $\ddot{q}(t_i)$.
		
			\State one can substitute by $q(t_i)$, $\dot{q}(t_i)$, $\ddot{q}(t_i)$ values in the right hand side of the nonholonomic constraints, and obtain the new $\sigma_b(t_i)$ from left hand side and by numerical differentiation, one can get the new $\dot{\sigma}_b(t_i)$ and $\ddot{\sigma}_b(t_i)$.
			
			\State Put $i=i+1$, $t_i=t_{i-1}+\Delta t$, and repeat to obtain $\zeta_r(t_i)$.

		\EndWhile
	\end{algorithmic}
\end{algorithm} 
The validation of this algorithm will be checked at the end of this section. The six algebraic inverse kinematic equations are derived next.

Define the general form for the rotation matrix $R_e$ as a function of the end-effector variables $\chi_e$, as
\begin{equation}
R_e= R_b R^b_e = \begin{bmatrix}
R_{11} & R_{12} & R_{13} \\
R_{21} & R_{22}& R_{23} \\
R_{31} & R_{32} & R_{33}
\end{bmatrix},
\label{eq:Re}
\end{equation}

where the elements of $R_e$ are given as

\begin{equation}
R_{{11}} =  C_{\theta_2}( C_{\theta_1}( C_{\phi} S_{\psi} -  C_{\psi} S_{\phi} S_{\theta}) -  S_{\theta_1}( S_{\psi} S_{\phi} +  C_{\psi} C_{\phi} S_{\theta})) +  C_{\psi} C_{\theta} S_{\theta_2}, 
\end{equation}
\begin{equation}
R_{{12}} =  C_{\psi} C_{\theta} C_{\theta_2} -  S_{\theta_2}( C_{\theta_1}( C_{\phi} S_{\psi} -  C_{\psi} S_{\phi} S_{\theta}) -  S_{\theta_1}( S_{\psi} S_{\phi} +  C_{\psi} C_{\phi} S_{\theta})), 
\end{equation}
\begin{equation}
R_{{13}} =   C_{\theta_1}( S_{\psi} S_{\phi} +  C_{\psi} C_{\phi} S_{\theta}) +  S_{\theta_1}( C_{\phi} S_{\psi} -  C_{\psi} S_{\phi} S_{\theta}), 
\end{equation}
\begin{equation}
R_{{21}} =     C_{\theta} S_{\psi} S_{\theta_2} -  C_{\theta_2}(C_{\theta_1}( C_{\psi} C_{\phi} +  S_{\psi} S_{\phi} S_{\theta}) -  S_{\theta_1}( C_{\psi} S_{\phi} -  C_{\phi} S_{\psi} S_{\theta})), 
\end{equation}
\begin{equation}
R_{{22}} =    S_{\theta_2}( C_{\theta_1}( C_{\psi} C_{\phi} +  S_{\psi} S_{\phi} S_{\theta}) -  S_{\theta_1}( C_{\psi} S_{\phi} -  C_{\phi} S_{\psi} S_{\theta})) +  C_{\theta} C_{\theta_2} S_{\psi}, 
\end{equation}
\begin{equation}
R_{{23}} = -  C_{\theta_1}( C_{\psi} S_{\phi} -  C_{\phi} S_{\psi} S_{\theta}) -  S_{\theta_1}( C_{\psi} C_{\phi} +  S_{\psi} S_{\phi} S_{\theta}), 
\end{equation} 
\begin{equation}
R_{{31}} =  -  S_{\theta} S_{\theta_2} -  C_{\theta_2}( C_{\phi} C_{\theta} S_{\theta_1} +  C_{\theta} C_{\theta_1} S_{\phi}), 
\end{equation}
\begin{equation}
R_{{32}} =   S_{\theta_2}( C_{\phi} C_{\theta} S_{\theta_1} +  C_{\theta} C_{\theta_1} S_{\phi}) -  C_{\theta_2} S_{\theta}, 
\end{equation}
\begin{equation}  
R_{{33}}  =   C_{\phi} C_{\theta} C_{\theta_1} -  C_{\theta} S_{\phi} S_{\theta_1}.   
\end{equation}

According to the structure of $R_e$ from (\ref{eq:Re_cpct}), the inverse orientation is carried out first followed by inverse position. 

Let us define the end-effector orientation, $R_{e,\Phi_e}$, as function of the Euler angles, $\Phi_e = [\psi_e \ \theta_e \ \phi_e]^T$, as

\begin{equation}
R_e= \begin{bmatrix}
  C_{\psi_e} C_{\theta_e} &  C_{\psi_e} S_{\phi_e} S_{\theta_e} - C_{\phi_e} S_{\psi_e} &  S_{\psi_e} S_{\phi_e} + C_{\psi_e} C_{\phi_e} S_{\theta_e} \\
  C_{\theta_e} S_{\psi_e} &  C_{\psi_e} C_{\phi_e} + S_{\psi_e} S_{\phi_e} S_{\theta_e} &  C_{\phi_e} S_{\psi_e} S_{\theta_e} - C_{\psi_e} S_{\phi_e} \\
  -S_{\theta_e} &                               C_{\theta_e} S_{\phi_e} &                               C_{\phi_e} C_{\theta_e} \\  
\end{bmatrix} = \begin{bmatrix}
r_{11} & r_{12} & r_{13} \\
r_{21} & r_{22}& r_{23} \\
r_{31} & r_{32} & r_{33}
\end{bmatrix},
\label{eq:Red}
\end{equation}

Equating (\ref{eq:Re}) and (\ref{eq:Red}) provides the inverse orientation as following.
The inverse orientation has three cases based on the value of $\theta_1$ which can be calculated from the element $r_{33}$ as
\begin{equation}
C_{\phi} C_{\theta} C_{\theta_1} - S_{\phi} C_{\theta} S_{\theta_1} = r_{33}.
\label{case1_th1_1}
\end{equation}
By rearranging (\ref{case1_th1_1}) and solving the resulting equation for $\theta_1$, then
\begin{equation}
\theta_1 = 2 \ {\rm atan2} (-2b_1 \pm \sqrt{(2b_1)^2 - 4(-a_1 - r_{33}) (a_1 - r_{33})}, 2(-a_1 - r_{33})),
\label{case1_th1_2}
\end{equation}
where $a_1= C_{\phi} C_{\theta}$ and $b_1= - S_{\phi} C_{\theta}$.

Based on the value of $\theta_1$, three cases for the rotational inverse kinematics solution exist.

\uppercase{\textbf{Case 1}}: $\theta_1$ $\neq$ $0$ and $\theta_1$ $\neq$ $\pi$

Inspecting $r_{13}$ and $r_{23}$, one can find value of $\psi$ as
\begin{equation}
r_{13} = a_2 S_{\psi} + b_2 C_{\psi},
\label{case1_ep_1}
\end{equation}
\begin{equation}
r_{23} = c_2 S_{\psi} + d_2 C_{\psi},
\label{case1_ep_2}
\end{equation}

where $a_2= C_{\phi} S_{\theta_1} + S_{\phi} C_{\theta_1}$, $b_2= C_{\phi} S_{\theta} C_{\theta_1} - S_{\phi} S_{\theta} S_{\theta_1}$, $c_2= b_2 $, and $d_2= -a_2$.

Solving (\ref{case1_ep_1}) and (\ref{case1_ep_2}), will give
\begin{equation}
S_{\psi} = \frac{r_{23} b_2 - r_{13} d_2}{-d_2 a_2 + b_2 c_2},
\label{case1_ep_3}
\end{equation}

\begin{equation}
C_{\psi} = \frac{r_{23} a_2 - r_{13} c_2}{d_2 a_2 - b_2 c_2},
\label{case1_ep_4}
\end{equation}

\begin{equation}
\psi = {\rm atan2}(S_{\psi},C_{\psi}).
\label{case1_ep_5}
\end{equation}

By inspecting $r_{32}$ and $r_{31}$, $\theta_2$ can be found as

\begin{equation}
r_{32} = a_3 S_{\theta_2} + b_3 C_{\theta_2},
\label{case1_th2_1}
\end{equation}
\begin{equation}
r_{31} = c_3 S_{\theta_2} + d_3 C_{\theta_2},
\label{case1_th2_2}
\end{equation}

where $a_3= C_{\phi} C_{\theta} S_{\theta_1} + C_{\theta} S_{\phi} C_{\theta_1}$, $b_3= -S_{\theta} $, $c_3= b_3 $, and $d_3= - a_3$.

Solving (\ref{case1_th2_1}) and (\ref{case1_th2_2}), then
\begin{equation}
S_{\theta_2} = \frac{r_{31} b_3 - r_{32} d_3}{-d_3 a_3 + b_3 c_3},
\label{case1_th2_3}
\end{equation}

\begin{equation}
C_{\theta_2} = \frac{r_{31} a_3 - r_{32} c_3}{d_3 a_3 - b_3 c_3},
\label{case1_th2_4}
\end{equation}

\begin{equation}
\theta_2 = {\rm atan2}(S_{\theta_2},C_{\theta_2}).
\label{case1_th2_5}
\end{equation}

\uppercase{\textbf{Case 2}}: $\theta_1 = 0 $

If $\theta_1 = 0$, then  the sum $\theta_2+\psi$ can be determined. One can assume any value for $\psi$ and get $\theta_2$. Therefore, there are infinity of solutions. By putting $\psi$ = $0$,  the value of $\theta_2$ can be determined as follows

Inspecting $r_{11}$ and $r_{12}$, $\theta_2$ can be found as

\begin{equation}
r_{11} = a_4 S_{\theta_2} + b_4 C_{\theta_2},
\label{case2_th2_1}
\end{equation}
\begin{equation}
r_{12} = c_4 S_{\theta_2} + d_4 C_{\theta_2},
\label{case2_th2_2}
\end{equation}

where $a_4= C_{\theta}$, $b_4= - S_{\phi} S_{\theta}$, $c_4= - b_4 $, and $d_4= a_4$.

Solving (\ref{case2_th2_1}) and (\ref{case2_th2_2}), then
\begin{equation}
S_{\theta_2} = \frac{r_{12} b_4 - r_{11} d_4}{-d_4 a_4 + b_4 c_4},
\label{case2_th2_3}
\end{equation}

\begin{equation}
C_{\theta_2} = \frac{r_{12} a_4 - r_{11} c_4}{d_4 a_4 - b_4 c_4},
\label{case2_th2_4}
\end{equation}

\begin{equation}
\theta_2 = {\rm atan2}(S_{\theta_2},C_{\theta_2}).
\label{case2_th2_5}
\end{equation}

\uppercase{\textbf{Case 3}}: $\theta_1 = \pi$ 

Since $S_{\theta_1}$ = $0$, this is similar to case $2$. However, $\theta_2-\psi$ can be determined and by choosing $\psi$ = $0$, an expression for $\theta_2$ can be determined as follows:  

Inspecting $r_{11}$ and $r_{12}$, $\theta_2$ can be found as

\begin{equation}
r_{11} = a_5 S_{\theta_2} + b_5 C_{\theta_2},
\label{case3_th2_1}
\end{equation}
\begin{equation}
r_{12} = c_5 S_{\theta_2} + d_5 C_{\theta_2},
\label{case3_th2_2}
\end{equation}

where $a_5= a_4$, $b_5= -b_4 $, $c_5= - b_5 $, and $d_5= a_5$.

Solving (\ref{case3_th2_1}) and (\ref{case3_th2_2}), then
\begin{equation}
S_{\theta_2} = \frac{r_{12} b_5 - r_{11} d_5}{-d_5 a_5 + b_5 c_5},
\label{case3_th2_3}
\end{equation}

\begin{equation}
C_{\theta_2} = \frac{r_{12} a_5 - r_{11} c_5}{d_5 a_5 - b_5 c_5},
\label{case3_th2_4}
\end{equation}

\begin{equation}
\theta_2 = {\rm atan2}(S_{\theta_2},C_{\theta_2}).
\label{case3_th2_5}
\end{equation}

As shown above, there are two possible solutions for the rotational inverse kinematics problem provided that one puts $\psi$ = $0$ in cases $2$ and $3$. When one starts applying the above described algorithm in real time, one can select the solution of $\theta_1$ which coincide with the given configuration of the quadrotor manipulation system. After that, one can continue with the one of the two solutions which produces this $\theta_1$.

Finally, the inverse position is determined from (\ref{eq:pe}) as
\begin{equation}
p_b = p_e - R_b p^b_{eb}.
\label{eq:inv_pb}
\end{equation}
There is one solution for this inverse position problem.
\section{Verification}

To verify the proposed solution of the inverse kinematics, the desired task space trajectories are chosen to make a circular helix in position and quintic polynomial \cite{spong2006robot} for the orientation.

After obtaining the joint space variables from the proposed algorithm, as shown in Fig. \ref{fig:inv_non_blk}, the forward kinematics is applied to find the actual task space trajectory.

\begin{figure}[h]
	\centering
	\begin{tabular}{cc}
		\subfloat[$x_{e}$]{\includegraphics[width=0.4\columnwidth]{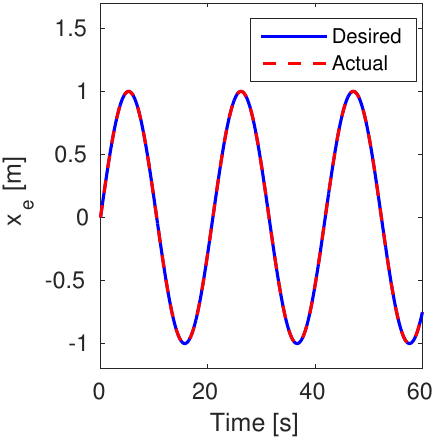}}&
		\subfloat[$y_{e}$]{\includegraphics [width=0.4\columnwidth]{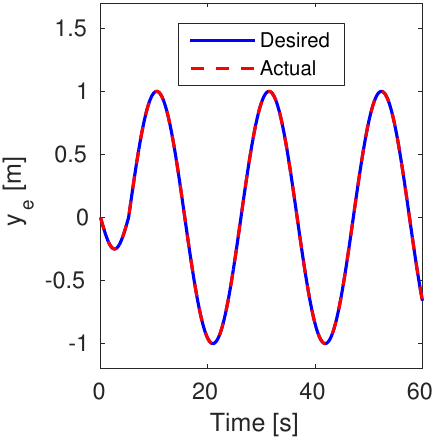}}\\
		\subfloat[$z_{e}$]{\includegraphics [width=0.4\columnwidth]{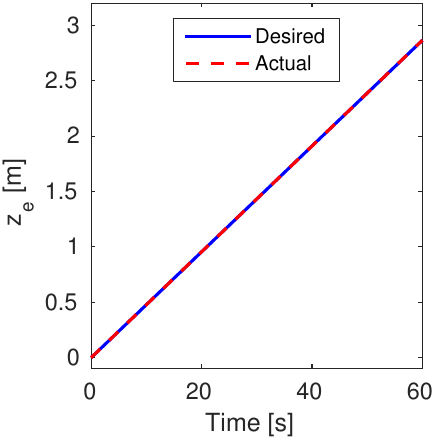}}&
		\subfloat[$\phi_{e}$]{\includegraphics [width=0.4\columnwidth]{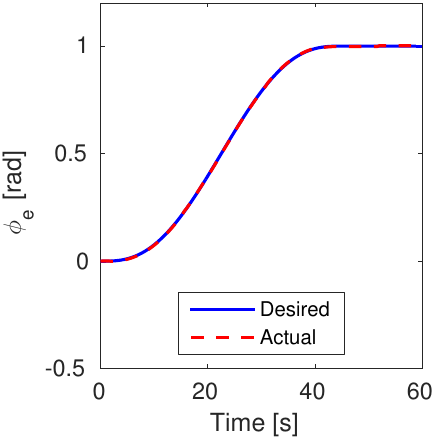}}\\
		\subfloat[$\theta_{e}$]{\includegraphics [width=0.4\columnwidth]{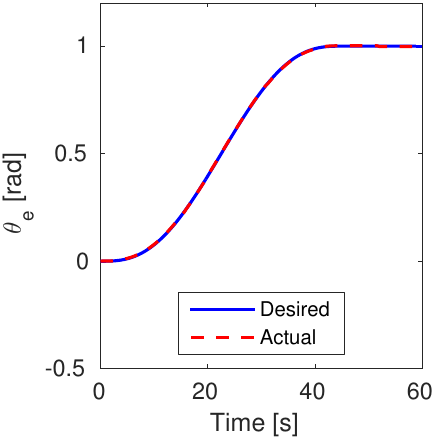}}&
		\subfloat[$\psi_{e}$]{\includegraphics [width=0.4\columnwidth]{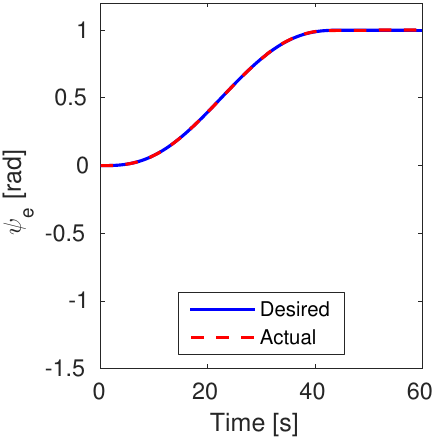}}
	\end{tabular}
	\caption{The actual response of the end-effector pose in the case of the slower trajectories: a) $x_{e}$, b) $y_{e}$, c) $z_{e}$, d) $\phi_{e}$, e) $\theta_{e}$, and f) $\psi_{e}$}
	\label{fig:inv_kin_ee_nonhol_results}
\end{figure}

\begin{figure}[h]
	\centering
     \includegraphics[width=0.5\columnwidth]{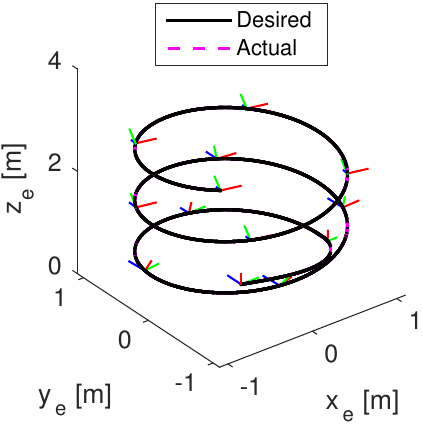}
	\caption{End-effector 3D trajectory in the case of the slower trajectories}
	      \label{fig:inv_non_rslt}
\end{figure}

\begin{figure}[h]
	\centering
	\begin{tabular}{cc}
		\subfloat[$x_{e}$]{\includegraphics[width=0.4\columnwidth]{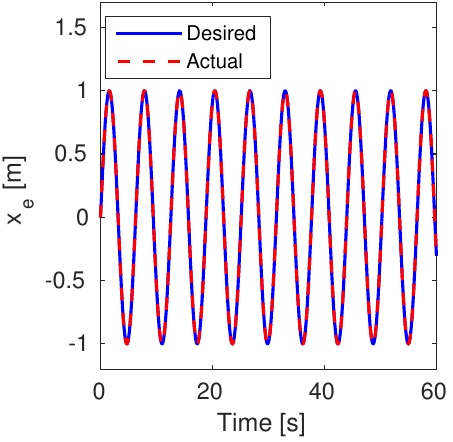}}&
		\subfloat[$y_{e}$]{\includegraphics [width=0.4\columnwidth]{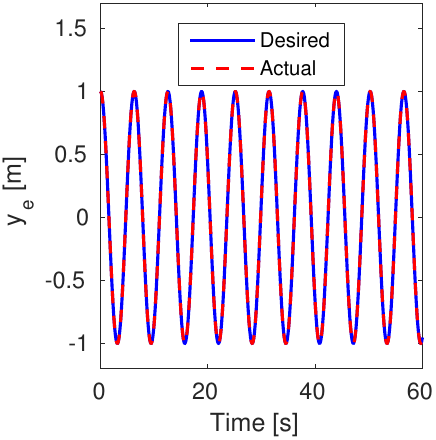}}\\
		\subfloat[$z_{e}$]{\includegraphics [width=0.4\columnwidth]{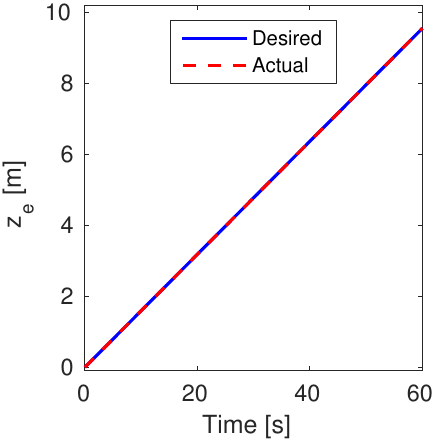}}&
		\subfloat[$\phi_{e}$]{\includegraphics [width=0.4\columnwidth]{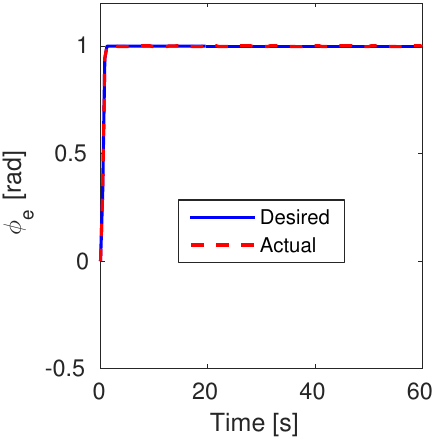}}\\
		\subfloat[$\theta_{e}$]{\includegraphics [width=0.4\columnwidth]{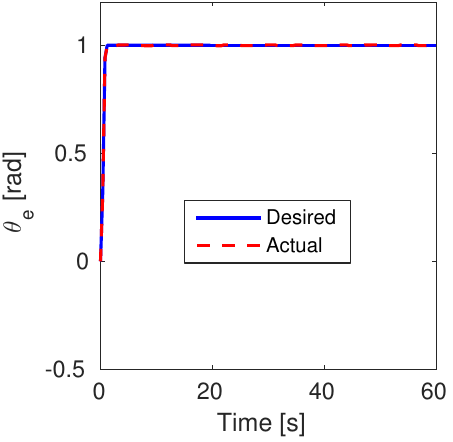}}&
		\subfloat[$\psi_{e}$]{\includegraphics [width=0.4\columnwidth]{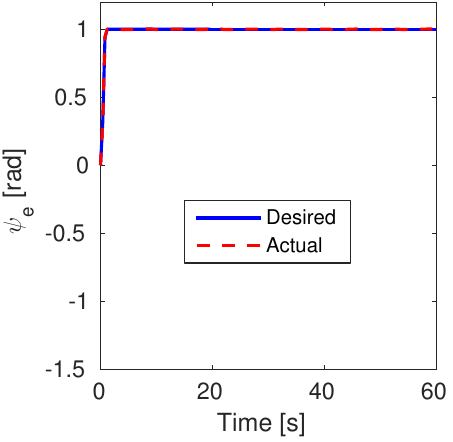}}
	\end{tabular}
	\caption{The actual response of the end-effector pose in the case of the faster trajectories: a) $x_{e}$, b) $y_{e}$, c) $z_{e}$, d) $\phi_{e}$, e) $\theta_{e}$, and f) $\psi_{e}$}
	\label{fig:inv_kin_ee_nonhol_results_fast}
\end{figure}

\begin{figure}[h]
	\centering
      \includegraphics [width=0.5\columnwidth]{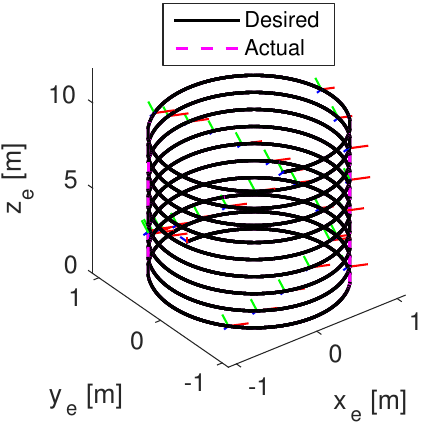}
       \caption{End-effector 3D trajectory in the case of the faster trajectories}
       \label{fig:inv_non_rslt_fast}
\end{figure}

The comparison between the actual and the desired task space trajectories is shown in Figs. \ref{fig:inv_kin_ee_nonhol_results} and \ref{fig:inv_non_rslt}from which one can recognize that the actual and desired trajectories are coincided. In Figs. \ref{fig:inv_kin_ee_nonhol_results_fast} and \ref{fig:inv_non_rslt_fast}, another case study is investigated to show the capability of the proposed algorithm to deal with higher speed desired trajectories. In this study, the desired trajectories are chosen to be faster than that in Fig. \ref{fig:inv_non_rslt} by ten times. Note that the maximum speed for the trajectories are 5 m/s for the translation motion and 6 rad/s for the rotational motion .These figures show that the actual and desired trajectories are coincided under both slower or faster trajectories, which ensures the validity of the proposed inverse kinematic algorithm, and proves that the proposed system has the ability to track arbitrary 6-DOF task space trajectory. 

The obtained accurate inverse kinematics solution enables one to design the controller in the quadrotor/joint space to track the desired trajectories in the task space. This control design is the subject of the next chapter where certain control objectives are achieved for such complex dynamic system.
%
%
%
%
%


\chapter*{\uppercase{List of Publications}} 
\addcontentsline{toc}{chapter}{LIST OF PUBLICATIONS}
\markboth{LIST OF PUBLICATIONS}{LIST OF PUBLICATIONS}
\phantomsection
	\footnotesize
\section*{Conference Papers} 
\begin{enumerate}

\item \underline{Ahmed Khalifa}, Mohamed Fanni, Ahmed Ramadan, and Ahmed Abo-Ismail, "Controller Design of a New Quadrotor Manipulation System Based on Robust Internal-loop Compensator", in \emph{the IEEE International Conference on Autonomous Robot Systems and Competitions (ICARSC), Vila Real, Portugal}, April 8-10, 2015, pp. 97-102. 

\item \underline{Ahmed Khalifa}, Mohamed Fanni, and Toru Namerikawa, "MPC and DOb-based Robust Optimal Control of a New Quadrotor Manipulation System", in \emph{the European Control Conference (ECC), Aalborg, Denmark}, June 29 - July 1, pp. 483-488.

\item \underline{Ahmed Khalifa}, Mohamed Fanni, and Toru Namerikawa, "On the Fault Tolerant Control of a Quadrotor Manipulation System via MPC and DOb Approaches", in \emph{the $55^{th}$ Annual Conference of the Society of Instrument and Control Engineers of Japan (SICE), Tsukuba, Japan}, September 20-23, 2016.

\item \underline{Ahmed Khalifa}, Mohamed Fanni, and Toru Namerikawa, "Hybrid\\ 
Acceleration/Velocity-based Disturbance Observer for a Quadrotor Manipulation System", in \emph{ the 2016 IEEE Multi-Conference on Systems and Control (MSC), International Conference on Control Applications (CCA), Buenos Aires, Argentina}, September 19-22, 2016.

\end{enumerate}

\section*{Journal Papers}

\begin{enumerate}
	\item \underline{Ahmed  Khalifa} and Mohamed Fanni, "Position Inverse Kinematics and Robust Internal-loop Compensator-based Control of a New Quadrotor Manipulation System", in \emph{the International Journal of Imaging and Robotics}, 16.1 (2016): 94-113.
	
	\item \underline{Ahmed  Khalifa} and Mohamed Fanni, "POSITION ANALYSIS AND CONTROL OF A NEW QUADROTOR MANIPULATION SYSTEM", in \emph{the International Journal of Robotics and Automation}, (Accepted for publication).
	
	\item \underline{Ahmed  Khalifa} and Mohamed Fanni, "A New Quadrotor Manipulation System: Modeling and Point-to-Point Task Space Control", in \emph{the International Journal of Control, Automation and Systems}, (Accepted for publication).
\end{enumerate}	

%
%
	
\normalsize





\renewcommand{\bibname}{REFERENCES}
\bibliography{11_backmatter/references} 
\bibliographystyle{Latex/Classes/ieeetr}


\renewcommand{\appendixtocname}{APPENDICES}
\renewcommand{\appendixpagename}{APPENDICES}

\end{spacing}
%
\ifdefined\phantomsection
\phantomsection  
\else
\fi

\end{document}